\documentclass[10pt,journal,compsoc]{IEEEtran}
\usepackage[utf8]{inputenc} 
\usepackage[T1]{fontenc}    
\usepackage{url}            
\usepackage{booktabs}       
\usepackage{amsfonts}       
\usepackage{nicefrac}       
\usepackage{microtype}      
\usepackage{times}      
\usepackage[dvipsnames]{xcolor}
\usepackage{sidecap} 
\usepackage{wrapfig}

\usepackage{setspace}

\usepackage{graphicx}
\usepackage{amsmath}
\usepackage{amssymb}
\usepackage{booktabs}
\usepackage{bm}
\usepackage{float}
\usepackage{multicol}
\usepackage{multirow}
\usepackage{caption}
\usepackage{graphicx}
\usepackage{adjustbox}
\usepackage[pagebackref=false,breaklinks=true,colorlinks,bookmarks=false]{hyperref}

\usepackage{array}
\newcolumntype{P}[1]{>{\centering\arraybackslash}p{#1}}
\usepackage{dblfloatfix}
\usepackage{enumitem}
\usepackage{mathabx}

\usepackage{algpseudocode}
\usepackage{placeins}
\usepackage[linesnumbered,ruled,vlined]{algorithm2e}
\SetKwInput{KwInput}{Input}  
\SetKwInput{KwOutput}{Output}
\SetKwInput{KwRequire}{Require}
\SetKwInput{KwIP}{\em Importance Probing}
\SetKwInput{KwMA}{\em Main Adaptation}

\SetCommentSty{mycommfont}

\newcommand{\nocontentsline}[3]{}
\newcommand{\tocless}[2]{\bgroup\let\addcontentsline=\nocontentsline#1{#2}\egroup}

\newcommand{\eq}{\hspace{-0.15em}=\hspace{-0.15em}}

\usepackage{amssymb}
\usepackage{pifont}
\newcommand{\cmark}{\ding{51}}%
\newcommand{\xmark}{\ding{55}}%

\ifCLASSOPTIONcompsoc
  \usepackage[nocompress]{cite}
\else
  \usepackage{cite}
\fi
\ifCLASSINFOpdf

\else

\fi

\begin{document}
%
\title{AdAM: Few-Shot Image Generation via Adaptation-Aware Kernel Modulation
}
%
%
%
%

\author{
Yunqing Zhao$^{*}$,
Keshigeyan Chandrasegaran$^{*}$,
Milad Abdollahzadeh$^{*}$,\\
Chao Du, Tianyu Pang,
Ruoteng Li,
Henghui Ding,
Ngai-Man Cheung
\IEEEcompsocitemizethanks{
\IEEEcompsocthanksitem 
Yunqing Zhao, Keshigeyan Chandrasegaran, Milad Abdollahzadeh and Ngai-Man Cheung are with ISTD Pillar, Singapore University of Technology and Design, 487372, Singapore ({\footnotesize $^{*}$} indicates Equal Contribution).
\IEEEcompsocthanksitem Chao Du and Tianyu Pang are with Sea AI Lab (SAIL), Singapore.
\IEEEcompsocthanksitem Ruoteng Li was with ByteDance. Henghui Ding is with NTU, Singapore.
\IEEEcompsocthanksitem Corresponding to Ngai-Man Cheung. email: ngaiman\_cheung@sutd.edu.sg 
\IEEEcompsocthanksitem{Project Page \& Code:
\href{https://yunqing-me.github.io/AdAM/}{{\color{RubineRed}{https://yunqing-me.github.io/AdAM/}}}}
}
}

%
%

\markboth{Journal of \LaTeX\ Class Files,~Vol.~14, No.~8, June~2023}%
{Shell \MakeLowercase{\textit{et al.}}: Bare Demo of IEEEtran.cls for Computer Society Journals}
%



\IEEEtitleabstractindextext{%

\begin{abstract}

Few-shot image generation (FSIG) aims to learn to generate new and diverse images given few (e.g., 10) training samples. 
Recent work has addressed FSIG by leveraging a GAN pre-trained on a large-scale source domain and adapting it to the target domain with few target samples. 
Central to recent FSIG methods are knowledge preservation criteria, which select and preserve a subset of source knowledge to the adapted model. However, a major limitation of existing methods is that their knowledge preserving criteria consider only source domain/task and fail to consider target domain/adaptation in selecting source knowledge, casting doubt on their suitability for setups of different proximity between source and target domain.
Our work makes two contributions. Firstly, we revisit recent FSIG works and their experiments. We reveal that under setups which assumption of close proximity between source and target domains is relaxed, many existing state-of-the-art (SOTA) methods which consider only source domain in knowledge preserving perform no better than a baseline method. As our second contribution, we propose Adaptation-Aware kernel Modulation (AdAM) for general FSIG of different source-target domain proximity.
Extensive experiments show that AdAM consistently achieves SOTA performance in FSIG, including challenging setups where source and target domains are more apart.

\end{abstract}

}

\maketitle

\IEEEdisplaynontitleabstractindextext

\IEEEpeerreviewmaketitle


\tocless
\IEEEraisesectionheading{\section{Introduction}}
\label{section1}


\IEEEPARstart{G}{enerative} Adversarial Networks (GANs) \cite{goodfellow2014GAN, brock2018bigGAN, karras2020styleganv2} have been applied to a range of important applications including image generation \cite{karras2018styleGANv1, karras2020styleganv2, choi2020starganv2}, image-to-image translation \cite{zhu2017cycleGAN, liu2019FUNit}, image editing \cite{lin2021anycostGANs, liu2021denet_imageediting}, 
anomaly detection \cite{lim2018doping}, and data augmentation \cite{tran2021data_aug_gan, chai2021ensembling}. 
However, a critical issue is that these GANs often require large-scale datasets and computationally expensive resources to achieve good performance.
For example,  StyleGAN \cite{karras2018styleGANv1} is trained on Flickr-Faces-HQ (FFHQ) \cite{karras2018styleGANv1} that contains 70K images,
, and BigGAN \cite{brock2018bigGAN} is trained on ImageNet-1K \cite{deng2009imagenet}.
However, in many practical applications only a few samples are available (e.g., 
photos of rare animal species / skin diseases, or oil paintings by artists \cite{yaniv2019face_Modigliani}).
Training a generative model is problematic in this low-data regime, where the generator often suffers from mode collapse or blurred generated images \cite{feng2021WhenGansReplicate, ojha2021fig_cdc, noguchi2019BSA}.
To address this, {\em few-shot image generation} (FSIG) studies the possibility of generating sufficiently diverse and high-quality images, given very limited training data (e.g., 10 samples). FSIG also attracts an increasing interest for some downstream tasks, e.g., few-shot classification \cite{chai2021ensembling}.

\begin{table*}[!t]
    \centering
    \footnotesize
    \small
    \caption{Transfer learning for few-shot image generation: Various criteria are proposed 
    in baseline and state-of-the-art methods 
    to augment baseline transfer learning to preserve subset of source model’s knowledge into the adapted model.
    }
    \vspace{-2.5 mm}
        \begin{tabular}{p{1.7cm}|P{10.2cm}|P{2.0cm}|P{2.5cm}}
        \toprule[1.2pt]
        
        \multirow{2}{*}{\textbf{Method}} &
        \multirow{2}{*}{\textbf{Knowledge preserving criteria for FSIG}} & 
        {\textbf{Source domain /task aware}} &
        {\textbf{Target domain /adaptation aware}}
        \\\hline
        TGAN \cite{wang2018transferringGAN} &
        Not available.
        & --
        & --
        \\ \hline
        FreezeD \cite{mo2020freezeD} & 
                Preservation of lower layers of the discriminator pre-trained on the {\em source} domain.
        & {\color{black}\cmark}
        & {\color{black}\xmark}
        \\ \hline
        EWC \cite{li2020fig_EWC} & 
                Preservation of weights important to the {\em source} generative model pre-trained on the {\em source} domain.
        & {\color{black}\cmark}
        & {\color{black}\xmark}
        \\ \hline
        CDC \cite{ojha2021fig_cdc} & 
                Preservation of pairwise distances of generated images by the {\em source} generator pre-trained on the {\em source} domain.
        & {\color{black}\cmark}
        & {\color{black}\xmark}
        \\ \hline
        DCL \cite{zhao2022dcl} & 
                Preservation of multilevel semantic diversity of the generated images by the {\em source} generator pre-trained on the {\em source} domain.
        & {\color{black}\cmark}
        & {\color{black}\xmark}
        \\ \hline
        RSSA \cite{xiao2022rssa} & 
                Preservation of the spatial structural knowledge of the {\em source} model pre-trained on the {\em source} domain via cross-domain consistency losses.
        & {\color{black}\cmark}
        & {\color{black}\xmark}
        \\ \hline
        LLN \cite{mondal2023ill} & 
                Preservation of the {\em entire source} generator pre-trained on the {\em source} domain by freezing the generator and optimize the latent code during adaptation.
        & {\color{black}\cmark}
        & {\color{black}\xmark}
        \\ \hline
        {\bf AdAM  (Ours)} & 
        Preservation of kernels that are identified important in  {\textbf{adaptation}} of the source model to the {\em target}. 
        & {\color{black}\cmark}
        & {\color{black}\cmark}
         \\
        \bottomrule[1.2pt]
        \end{tabular}
    \label{table:criteria}
    \vspace{-3mm}
\end{table*}

\subsection{Transfer Learning for FSIG} 
Recent works in FSIG are based on transfer learning approach \cite{pan2009yang-qiang-transfer} i.e., leveraging the prior knowledge of a GAN pretrained on a large-scale, diverse source dataset (e.g., FFHQ \cite{karras2018styleGANv1} or ImageNet-1K \cite{deng2009imagenet}) and adapting it to a target domain with very limited samples (e.g., face paintings \cite{yaniv2019faceofart}).
As only very limited samples are provided to define the underlying distribution, standard fine-tuning of a pre-trained GAN suffers from mode collapse: the adapted model can only generate samples closely resembling the given few shot target samples \cite{wang2018transferringGAN,ojha2021fig_cdc}.
Therefore, recent works \cite{li2020fig_EWC, ojha2021fig_cdc, zhao2022dcl, xiao2022rssa, mondal2023ill}
have proposed to {\em augment} standard fine-tuning with different criteria to carefully preserve subset of source model's knowledge into the adapted model.
Various criteria has been proposed (Table~\ref{table:criteria}), and these 
{\em knowledge preserving criteria} have been central in recent FSIG research.
In general, these criteria aim to preserve subset of 
source model's knowledge which is deemed to be useful for target-domain sample generation, e.g., improving the diversity of target sample generation.

\subsection{Research Gaps to Prior Works} 
One major limitation of existing methods is that 
they consider {\em only} source domain in preserving subset of source model's knowledge into the adapted model.
In particular, these methods {\em fail to consider} target domain/adaptation task in selection of source model's knowledge (Table \ref{table:criteria}).
For example, EWC \cite{li2020fig_EWC} applies Fisher Information \cite{ly2017tutorial} to select important weights entirely based on the pretrained {\em source} model, and it aims to preserve these selected weights regardless of the target domain in adaptation.
Similar to EWC \cite{li2020fig_EWC}, CDC \cite{ojha2021fig_cdc} proposes an additional constraint to preserve pairwise distances of generated images by the {\em source} model, and there is no consideration of target domain/adaptation.
These {\em target/adaptation-agnostic} knowledge preserving criteria in recent works raise question regarding their suitability in different source/target domain setups.
It should be noted that  
existing FSIG works
(under very limited target samples) focus largely on setups where source and target domains are in {\em close proximity} (semantically) e.g., Human faces $\rightarrow$Baby faces \cite{ojha2021fig_cdc,zhao2022dcl}, or Cars$\rightarrow$Abandoned Cars \cite{ojha2021fig_cdc,zhao2022dcl}.
It is unclear about their performance when source/target domains are more apart
(e.g., FFHQ (Human faces) \cite{karras2018styleGANv1} $\rightarrow$ AFHQ (Animal faces) \cite{choi2020starganv2}).

\subsection{Our Contributions} 
In this paper, we take an important step to address these research gaps for FSIG. 
Specifically,
our work makes two contributions. {\bf As our first contribution}, we
revisit state-of-the-art (SOTA) algorithms and their experiments.
Importantly, we observe that when the close proximity assumption is relaxed in experiment setups and source/target domains are more apart,
SOTA
methods perform {\em no better} than a baseline fine-tuning method.
Our observation suggests that recent methods
considering only source domain/source task in knowledge preserving 
may not be suitable for {\em general} 
FSIG
when source and target domains are more apart.
To validate our claims,
we introduce additional experiments with 
different source/target domains,
analyze their proximity qualitatively and quantitatively, and examine existing methods under a unified framework.

Informed by our analysis, {\bf as our second contribution}, we propose Adaptation-Aware kernel Modulation (AdAM) to address general FSIG of different source/target domain proximity.
In marked contrast to existing works which preserve
knowledge important to {\em source} task,
AdAM aims to preserve a subset of source model's knowledge that are important to the {\em target} domain and the {\em adaptation} task.
Specifically, we propose an {Importance Probing} (IP) algorithm to identify kernels that encode important knowledge for adaptation to the target domain. 
Then, we preserve the knowledge of these kernels using a parameter-efficient {\em rank-constrained Kernel ModuLation} (KML) during adaptation. We conduct extensive experiments to show that our proposed method consistently achieves SOTA performance across source/target domains of different proximity, including challenging setups when source/target domains are more apart. 
\textbf{Our main contributions are summarized as follows:}
\begin{itemize}[leftmargin=10pt]
    \item 
    {We revisit existing FSIG methods and experiment setups. Our study uncovers  issues with existing methods when applied to source/target domains of different proximity.}
    \item 
    {We propose Adaptation-Aware kernel Modulation (AdAM) for FSIG. 
    Our method consistently achieves state-of-the-art
    performance both visually and quantitatively across source/target domains with different proximity. 
    }
\end{itemize}

\begin{figure*}[!t]
    \centering
    \includegraphics[width=\textwidth]{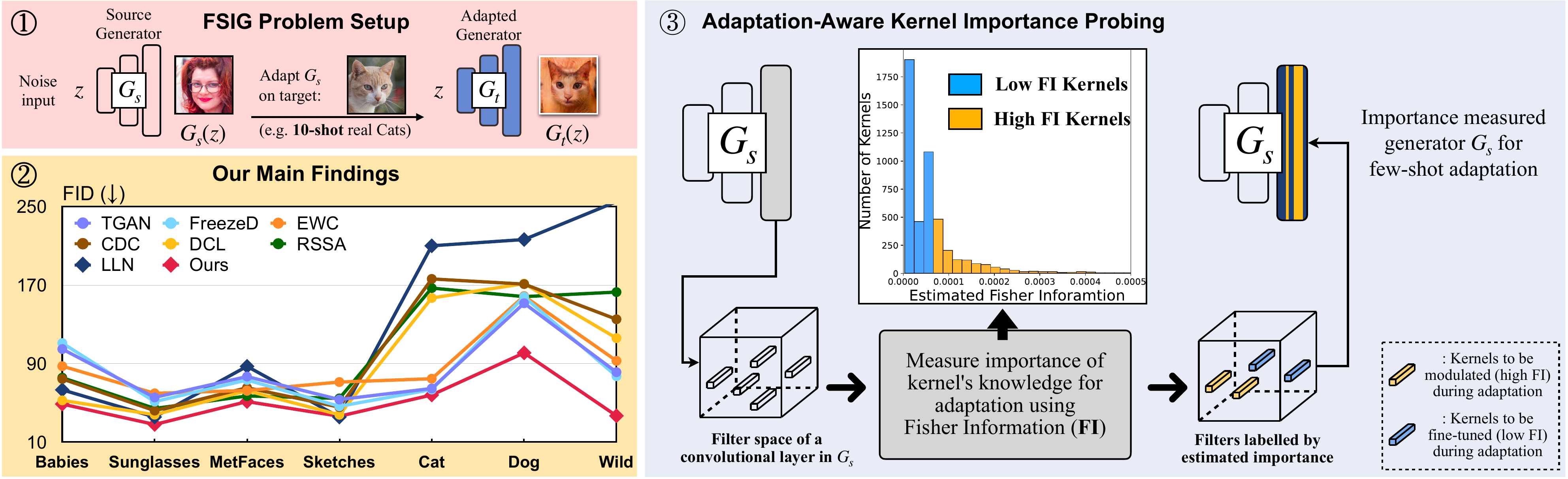}
    \vspace{-6mm}
    \caption{
    \textbf{Overview and our contributions.}
    \textcircled{\raisebox{-0.9pt}{1}}: 
    We consider the problem of FSIG with transfer learning using very limited target samples (e.g., 10-shot).
    \textcircled{\raisebox{-0.9pt}{2}}:
    $\bullet$ We discover that when the close proximity assumption between source-target domain is relaxed, SOTA methods (e.g. EWC \cite{li2020fig_EWC}, CDC \cite{ojha2021fig_cdc}, DCL \cite{zhao2022dcl}, RSSA \cite{xiao2022rssa}, LLN \cite{mondal2023ill})
    which consider only source domain/task in knowledge preserving perform \textit{no better} than a baseline fine-tuning method (TGAN \cite{wang2018transferringGAN}) (Sec.~\ref{sec:proximity-transferability}).
    $\bullet$ We propose a novel Adaptation-Aware kernel Modulation (AdAM) for FSIG that achieves SOTA performance across source / target domains with different proximity (Sec.~\ref{section4}). 
    \textcircled{\raisebox{-0.9pt}{3}} Schematic diagram of our proposed Importance Probing mechanism: 
    We measure the importance of each kernel for the target domain after probing and 
    preserve source domain knowledge that is important for target domain adaptation (Sec.~\ref{section4}). 
    The same operations are applied to the discriminator.
    }
    \label{fig:overview}
    \vspace{-2mm}
\end{figure*}

\tocless
\section{Related Works}
\label{section2}

\subsection{Few-shot Image Generation}

 Conventional few-shot learning \cite{fei2006one, snell2017prototypical} aims at learning a discriminative classifier for classification \cite{guo2020awgim, sun2021explanation, milad2021revisit, zhao2023fs}, segmentation \cite{liu2020crnet_fs_seg, boudiaf2021few} or detection \cite{zhang2021pnpdet, fan2021generalized} tasks.
 Differently, FSIG
 \cite{ojha2021fig_cdc, li2020fig_EWC, zhao2022dcl, zhao2022few, zhao2023exploring}
 aims at generating new and diverse samples given extremely limited samples (e.g., 10 shots) in training. Transfer learning has been applied to FSIG. 
 For example, \textbf{TGAN} \cite{wang2018transferringGAN} applies the simple GAN loss \cite{goodfellow2014GAN} to fine-tune all layers of both the generator and the discriminator. 
 \textbf{FreezeD} \cite{mo2020freezeD} fixes a few high-resolution discriminator layers during fine-tuning.
 To augment and improve simple fine-tuning, more recent works focus on preserving specific knowledge from the source models.  
 Elastic weight consolidation 
 (\textbf{EWC}) \cite{li2020fig_EWC} identifies important weights for the \textit{source} model and tries to preserve these weights. 
 Cross-domain Correspondence (\textbf{CDC}) \cite{ojha2021fig_cdc} preserves pair-wise distance of generated images from the source model to alleviate mode collapse.  
 Dual Contrastive Learning (\textbf{DCL}) \cite{zhao2022dcl} applies mutual information maximization to preserve multi-level diversity of the generated images by the source model.
 \textbf{RSSA} \cite{xiao2022rssa} aims to preserve the spatial structural knowledge of the generated image by the {\em source model}.
 Latent-code Learning Network (\textbf{LLN}) \cite{mondal2023ill} {\em freezes the entire source generator} for the {\em source} knowledge preservation.
 \textbf{However}, in this work, we observe that these 
 SOTA methods perform poorly when the source and target domains are more apart. Therefore, their proposed source knowledge preservation criteria {\em may not} be generalizable. 
 Based on our analysis, we propose a target/adaptation-aware knowledge selection which is more generalizable for 
 source/target domains with different proximity.
 
\subsection{Image Generation with Limited Data}
There is also a fair amount of work to focus on training GANs with less (but not few-shot) data, with efforts on introducing additional data augmentation methods \cite{karras2020ADA, tran2018distGAN}, regularization terms \cite{tseng2021regularizingGAN, zheng2023my}, modifying the GAN architectures  \cite{zhao2020leveraging_icml_adafm, zhao2020leveraging_icml_adafm,cong2020gan_memory, varshney2021cam-gan}.
Many of these works focus on setups with 
thousands of training images, e.g.: Flowers dataset \cite{nilsback2008oxford_flower} with 8,189 images used in \cite{cong2020gan_memory},  10\% of ImageNet-1K used in \cite{tseng2021regularizingGAN} and MetFaces introduced in \cite{karras2020ADA}.
On the other hand, FSIG with extremely few-shot data (e.g., 10) where we focus on in this work poses unique challenges. 
In particular, as pointed out in \cite{li2020fig_EWC, ojha2021fig_cdc,zhao2022dcl}, severe mode collapse and loss in diversity are critical challenges that require special attention. 

\subsection{Parameter Efficient Training}
 Kernel ModuLation (KML) was originally proposed in \cite{milad2021revisit} for adapting the model between different modes for few-shot classification (FSC).
 However, due to some differences between the multimodal meta-learner in \cite{milad2021revisit} and our transfer-learning-based scheme, there are different design choices when applying KML to our problem.
 First, in contrast to FSC work \cite{milad2021revisit} which follows a {\em discriminative learning} setup, we aim to address a problem in a {\em generative learning} setup. 
 Second, in the FSC setup, the modulation parameters are generated during adaptation to the target task with a pretrained modulation network trained on tens of thousands of few-shot tasks. 
 So the modulation parameters are not directly learned for a target few-shot task.
 In contrast, in our setup, the base kernel is frozen during the adaptation, and we directly learn the modulation parameters for a target domain/task using a very limited number of samples (e.g., 10-shot). 
 Finally, in FSC, usually, source and target tasks follow the same task distribution. In fact, in implementation, even though the classes are disjoint between source and target tasks, all of them are constructed using the data from the same domain (e.g., miniImageNet~\cite{vinyals2016matching}). However, in our setup, the source and target tasks/domains distributions could be very different (e.g., Human Faces (FFHQ) $\rightarrow$ Cats).
 We remark that in \cite{cong2020gan_memory}, a 
technique called AdaFM is introduced to update kernels. However, the underlying ideas and mechanisms of AdaFM and our KML are quite different. 
AdaFM is inspired by style-transfer literature \cite{adain}, and introduces independent scale and shift (scalar)  parameters to update individual channels of kernels to manipulate their styles. 
On the other hand, as will be discussed in Sec.~\ref{sec-kernel_modulation}, KML updates multiple kernels in a coordinated and parameter-efficient manner. 
In our experiment, we also test AdaFM in few-shot setups and compare its performance with KML.

\begin{figure*}

    \includegraphics[width=0.99\linewidth]{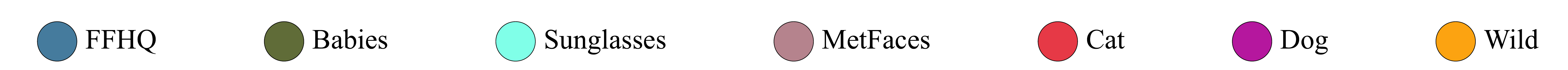} \\
    \begin{tabular}{ccc}
    \begin{minipage}{0.3\textwidth}
    \hspace*{-5 mm}
     \includegraphics[width=0.9\linewidth]{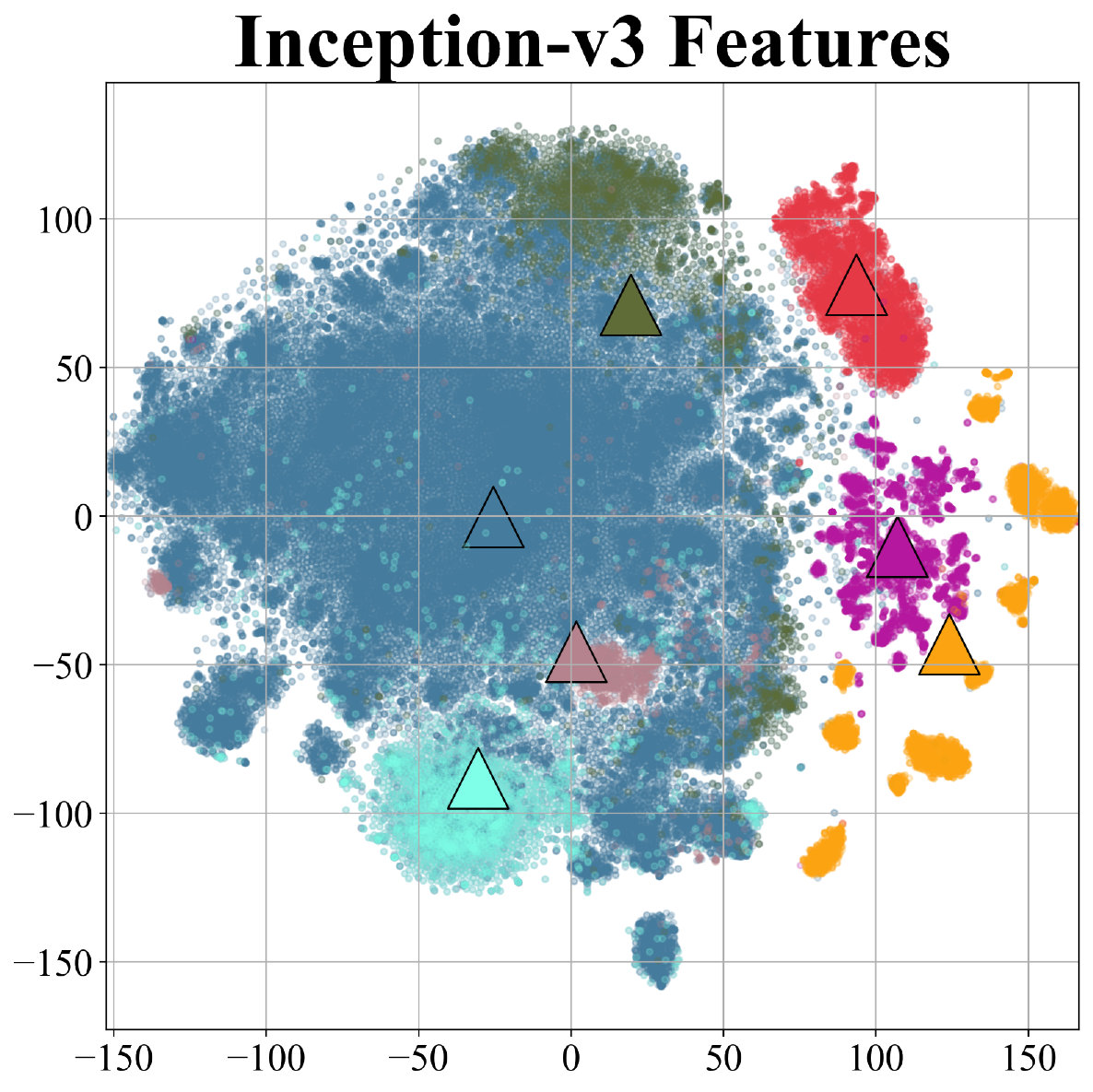}
     \end{minipage}
     
     & 

    \begin{minipage}{0.3\textwidth}
    \hspace*{-5 mm}
     \includegraphics[width=0.9\linewidth]{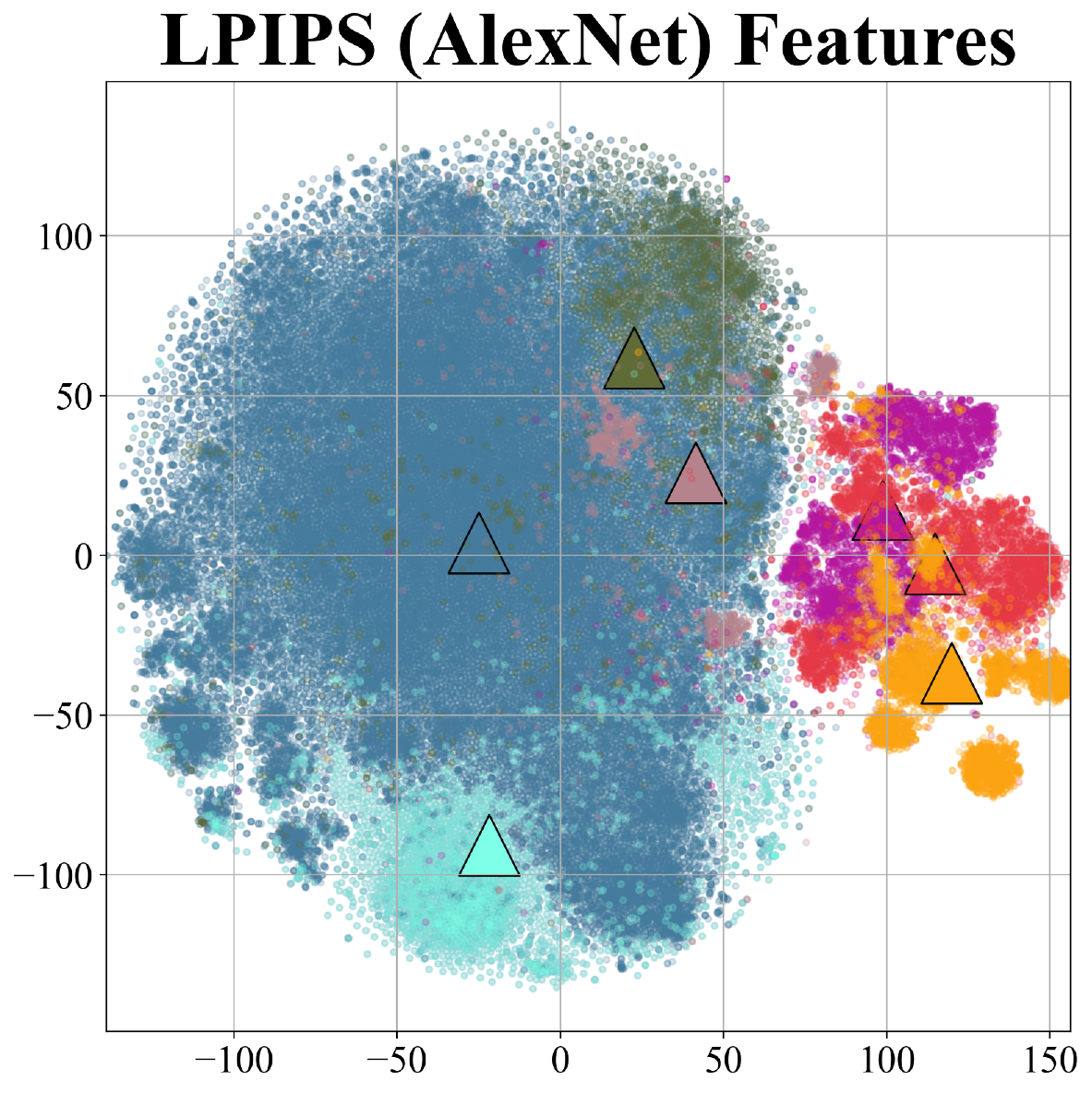}
     \end{minipage}
     
     & 
    
    \hspace*{-5 mm}
    \begin{minipage}{0.35\textwidth}
    \begin{adjustbox}{width=1.0\textwidth,center}
       \begin{tabular}{lccc}\toprule
        \textbf{Target Domain} &\textbf{Size} &\textbf{FID $\downarrow$} &\textbf{LPIPS $\downarrow$} \\ \toprule
        
        FFHQ \cite{karras2018styleGANv1} &70.0K &- &- \\ \midrule
        Babies \cite{ojha2021fig_cdc} &2.49K &147 &0.274 \\ 
        \midrule
        Sunglasses \cite{ojha2021fig_cdc} &2.68K &108 &0.347 \\ 
        \midrule
        MetFaces \cite{karras2020ADA} &1.33K &107 &0.358 \\ \midrule
        Cat \cite{choi2020starganv2} &5.15K &227 &0.479 \\ \midrule
        Dog \cite{choi2020starganv2} &4.74K &210 &0.442 \\ \midrule
        Wild \cite{choi2020starganv2} &4.74K &272 &0.484 \\ 
        
        \bottomrule
        \end{tabular}
    
    \end{adjustbox}
    \label{table:eta_set1}
    \end{minipage}
     \\
    \end{tabular}
  \caption{
\textbf{Qualitative / Quantitative analysis of source-target domain proximity:}
We use FFHQ \cite{karras2020styleganv2} as the source domain.
We show source-target domain proximity qualitatively by visualizing Inception-v3 \textbf{(Left)} \cite{szegedy2016rethinking} and LPIPS \textbf{(Middle)} \cite{zhang2018lpips} -- using AlexNet \cite{krizhevsky2012alexnet} backbone -- features, and quantitatively using FID / LPIPS metrics \textbf{(Right)}. 
For feature visualization, we use t-SNE \cite{JMLR:v9:vandermaaten08a_tsne} and show centroids ($\bigtriangleup$) for all domains. 
FID / LPIPS is measured with respect to FFHQ. 
There are two important observations: 
\textcircled{\raisebox{-0.9pt}{1}} Common target domains used in existing FSIG works (Babies, Sunglasses, MetFaces) are notably proximal to the source domain (FFHQ). This can be observed from the feature visualization and verified by FID / LPIPS measurements.
\textcircled{\raisebox{-0.9pt}{2}} 
We clearly show using feature visualizations and FID / LPIPS measurements that additional setups -- Cat \cite{choi2020starganv2}, Dog \cite{choi2020starganv2} and Wild \cite{choi2020starganv2} -- represent target domains that are distant from the source domain (FFHQ).
We remark that large FID values in this analysis are reasonable due to the distance between the source (FFHQ) and different target domains as observed from centroid distance/feature variance.
The effect of limited sample size (target domains) for FID / LPIPS measurements is minimal and we include rich supportive studies in our Supplement.}
\label{fig:proximity-visualization-measurements}
\vspace{-2mm}
\end{figure*}

\tocless

\section{Revisiting FSIG through the Lens of Source--Target Domain Proximity}
\label{sec:proximity-transferability}

In this section, we revisit existing FSIG methods (10-shot) \cite{wang2018transferringGAN, mo2020freezeD, li2020fig_EWC, ojha2021fig_cdc, zhao2022dcl} through the lens of source-target domain proximity.
Specifically, we scrutinize the experimental setups of existing FSIG methods and observe that SOTA \cite{li2020fig_EWC, ojha2021fig_cdc, zhao2022dcl} largely focus on adapting to target domains that are (semantically) proximal to the source domain: 
FFHQ $\rightarrow$ Baby Faces;
FFHQ $\rightarrow$ Sunglasses;
Cars $\rightarrow$ Wrecked Cars; 
Church $\rightarrow$ Haunted Houses.
This raises the question as to whether existing source-target domain setups sufficiently represent general FSIG scenarios.
Particularly, real-world FSIG applications may not contain target domains that are always proximal to the source domain (e.g.: FFHQ $\rightarrow$ Animal Faces).
Motivated by this, we conduct an in-depth qualitative and quantitative analysis of source-target
domain proximity where we introduce target domains that are distant from the source domain (Sec.~\ref{sub-sec:proximity-analysis}).
\textbf{Our analysis uncovers an important finding:}
Under our additional setups where the assumption of close proximity between source and target domain is relaxed, existing SOTA methods \cite{li2020fig_EWC, ojha2021fig_cdc, zhao2022dcl,xiao2022rssa,mondal2023ill}
which consider only the source domain in knowledge-preserving
perform \textit{no better} than a baseline fine-tuning method \cite{wang2018transferringGAN}.
We show this is due to their strong focus on preserving source domain/task knowledge, thereby not being able to adapt well to distant target domains (Sec.~\ref{sub-sec:proximity-sota-methods}).

\subsection{Source--Target Domain Proximity Analysis}
\label{sub-sec:proximity-analysis}

\textbf{Introducing target domains with varying degrees of proximity to the source domain.}
In this section, we formally introduce
source-target
domain proximity with in-depth analysis to scrutinize existing FSIG methods under different degrees of 
source-target domain proximity.
Following prior FSIG works \cite{ ojha2021fig_cdc, zhao2022dcl,xiao2022rssa, mondal2023ill}, 
we use FFHQ \cite{karras2020styleganv2} as the source domain in this analysis. We remark that existing works largely consider different types of human faces as target domains (i.e.: Babies \cite{ojha2021fig_cdc}, Sunglasses \cite{ojha2021fig_cdc}, MetFaces \cite{karras2020ADA}). 
To relax the close proximity assumption and study \textit{general} FSIG problems, we introduce more distant target domains namely Cat, Dog, and Wild (from AFHQ \cite{choi2020starganv2}, consisting of $\sim$15,000 high-quality animal face images) for our analysis. 

\textbf{Characterizing source-target domain proximity.}
Given the wide success of deep neural network features in representing meaningful semantic concepts \cite{talebi2018learned, talebi2018nima, morozov2021on_ssl_gan_evaluation}, we visualize Inception-v3 \cite{szegedy2016rethinking} and LPIPS \cite{zhang2018lpips} features for source and target domains to qualitatively characterize domain proximity. 
Further, we use FID \cite{heusel2017FID} and LPIPS  distance to quantitatively characterize source-target domain proximity.
We remark that FID involves distribution estimation (first, second order moments) \cite{heusel2017FID} and LPIPS computes pairwise distances (learned embeddings) \cite{zhang2018lpips} between source and target domains. 

\textbf{Results and analysis.}
Feature visualization and FID/ LPIPS measurement results are shown in Figure~\ref{fig:proximity-visualization-measurements}.
Our results both qualitatively (columns 1, 2) and quantitatively (column 3) show that target domains used in existing works (Babies \cite{karras2020styleganv2}, Sunglasses \cite{karras2020styleganv2}, MetFaces \cite{karras2020ADA}) are notably proximal to the source domain (FFHQ), and our additionally introduced target domains (Dog, Cat and Wild \cite{choi2020starganv2}) are distant from the source domain thereby relaxing the close proximity assumption in existing FSIG works.

\subsection{FSIG Methods under Relaxation of Close Domain Proximity Assumption}
\label{sub-sec:proximity-sota-methods}

Motivated by our analysis in Sec.~\ref{sub-sec:proximity-analysis}, we study the performance of existing FSIG methods \cite{wang2018transferringGAN, mo2020freezeD, li2020fig_EWC, ojha2021fig_cdc, zhao2022dcl,xiao2022rssa,mondal2023ill}
by relaxing the close proximity assumption between source and target domains: 
we investigate the performance of these FSIG methods across target domains of different proximity to the source domain,
which includes our additionally introduced target domains: Dog, Cat, and Wild.
The FID ($\downarrow$) results for FFHQ $\rightarrow$ Cat are: TGAN
(simple fine-tuning) 
\cite{wang2018transferringGAN}: \textbf{64.68}, EWC \cite{li2020fig_EWC}: 74.61, CDC \cite{ojha2021fig_cdc}: 176.21, DCL \cite{zhao2022dcl}: 156.82. The complete results are in Table~\ref{table:fid_scores}.

We emphasize that our investigation uncovers an important finding: 
under setups in which the assumption of close proximity between source and target domain is relaxed, existing SOTA FSIG methods \cite{li2020fig_EWC, ojha2021fig_cdc, zhao2022dcl, xiao2022rssa, mondal2023ill}
perform \textit{no better} than a baseline method \cite{wang2018transferringGAN}. 
This can be consistently observed in Table~\ref{table:fid_scores}.

\begin{figure}
\vspace{-0.205cm}
    \includegraphics[width=0.49\textwidth]{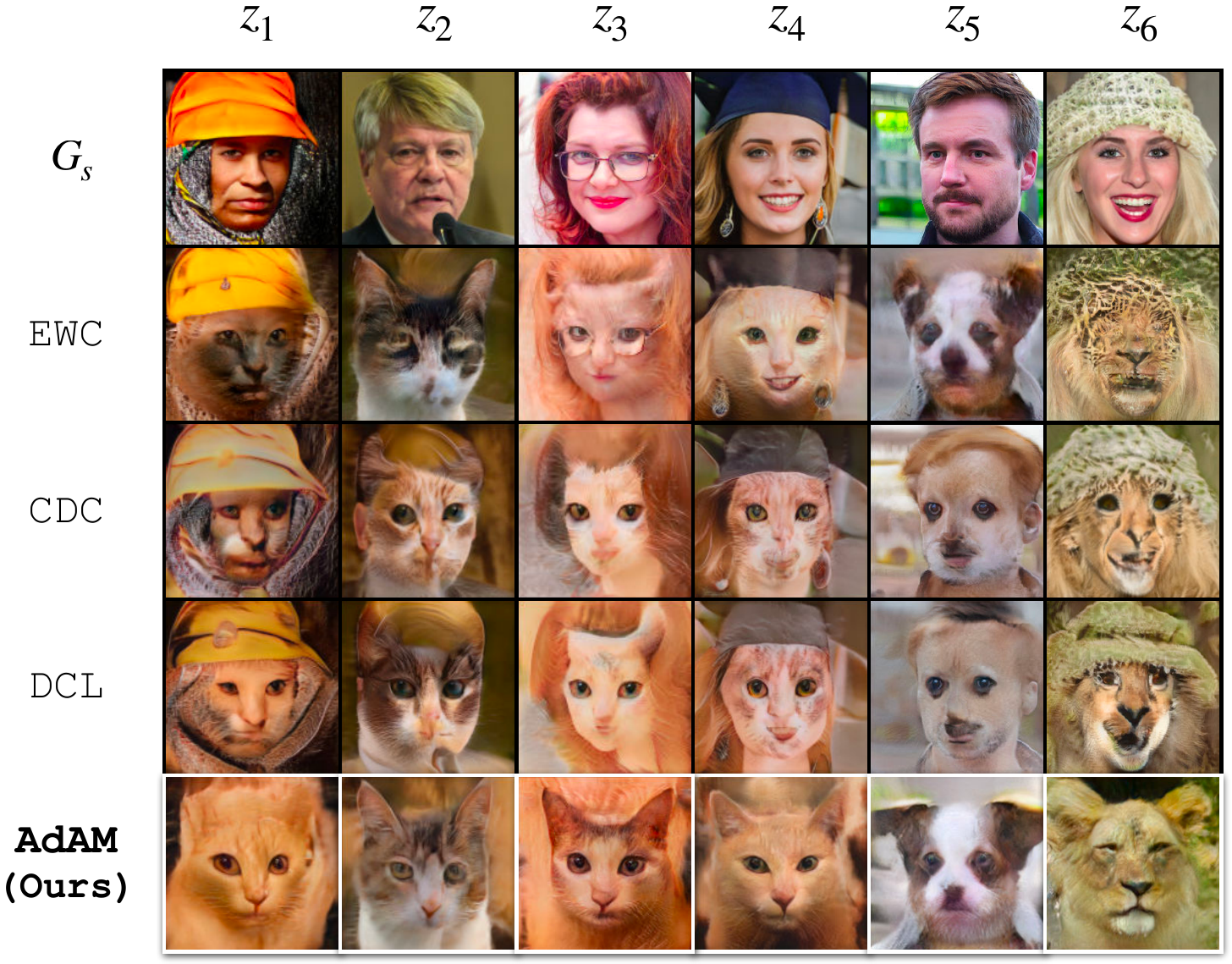}
\caption{
$G_s$ is the source generator (FFHQ). Adapting from the source domain (FFHQ) to distant target domains (Cat/Dog/Wild) using SOTA FSIG methods EWC \cite{li2020fig_EWC}, CDC \cite{ojha2021fig_cdc}, DCL \cite{zhao2022dcl} results in observable knowledge transfer that is incompatible to the target domain, e.g.: source task knowledge such as \textit{Caps ($z_1, z_4, z_6$), Hair styles/color -- brown ($z_2,z_5, z_6$), red-hair ($z_3$), Eyeglasses ($z_3$)} from FFHQ are transferred to Cat/Dog/Wild during adaptation which is not appropriate. 
Our method can alleviate these issues.
}
\label{fig:failure-cases-sota}
\vspace{-4mm}
\end{figure}

This finding is critical as it exposes a serious drawback of SOTA FSIG methods \cite{li2020fig_EWC, ojha2021fig_cdc, zhao2022dcl, xiao2022rssa, mondal2023ill} when close domain proximity (between source and target) assumption is relaxed.
We further analyse generated images from these methods and observe that they are unable to adapt well to distant target domains due to \textit{only considering source domain / task in knowledge preservation.}
This can be clearly observed from Figure~\ref{fig:failure-cases-sota}.
We remark that TGAN (simple baseline) \cite{wang2018transferringGAN} also suffers from severe mode collapse. 
Given that our investigation uncovers an important problem in SOTA FSIG methods, we tackle this problem in Sec.~\ref{section4}.
Figure~\ref{fig:failure-cases-sota} (last row) shows a glimpse of our proposed method.

\tocless
\section{Adaptation-Aware Kernel Modulation}
\label{section4}

We focus on this question: 
{\em ``Given a pretrained GAN on a source domain $\mathcal{S}$, and a few samples from a target domain $\mathcal{T}$, which part of the source model's knowledge should be preserved, and which part should be updated, during the adaptation from $\mathcal{S}$ to $\mathcal{T}$?''}
In contrast to SOTA FSIG methods \cite{li2020fig_EWC, ojha2021fig_cdc, zhao2022dcl, xiao2022rssa, mondal2023ill}, we propose an adaptation-aware FSIG that also considers the target domain/adaptation task in deciding which part of the source model's knowledge to be preserved.
In a CNN, each {\em kernel} is responsible for a specific part of knowledge (e.g., pattern or texture). Similar behavior is also observed for both generator \cite{bau2019gan} and discriminator \cite{bau2017network} in GANs.
Therefore, in this work, we make this knowledge preservation decision at the kernel level, i.e., \emph{\bfseries\boldmath casting the knowledge preservation in FSIG to a decision problem of whether a kernel is important when adapting from $\mathcal{S}$ to $\mathcal{T}$}.

Our FSIG algorithm has two
main steps: (i)  a lightweight {\em importance probing} step, and (ii)  {\em main adaptation} step.
In the first step, i.e.,  importance probing,  we adapt the model using a parameter-efficient design to the target domain for a limited number of iterations, 
and during this adaptation, we measure the importance of each individual kernel for the {\em target domain}.
The outputs of importance probing are decisions of importance/unimportance of individual kernels.
Then, in the main adaptation step, we preserve the knowledge of 
important kernels and update the knowledge of 
unimportant kernels.
The overview of the proposed system is shown in Figure~\ref{fig:overview}
and the pseudo-code is shown in Algorithm~\ref{alg:main}.

\subsection{Importance Probing for FSIG}
Our intuition for the proposed importance probing is: {\em ``The source GAN kernels have different levels of importance for each target domain.''}
For example, different subsets of kernels could be important when adapting a pretrained GAN on FFHQ to Babies~\cite{ojha2021fig_cdc} compared to adapting the same pretrained GAN to Cat~\cite{choi2020starganv2}. 
Therefore, we aim for a knowledge preservation criterion that is target domain/adaptation-aware (a comparison is in Table~\ref{table:criteria}).
In order to achieve adaptation awareness, we propose a lightweight importance probing algorithm that considers adaptation from the source to the target domain.
There are two important design considerations: probing under (i) extremely limited number of target data and (ii) low computation overhead.

As discussed, in this  {\em importance probing} step, we adapt the source model to the target domain for a limited number of iterations and with a few available target samples.
During this short adaptation step, we measure the importance of the kernel for the adaptation task.
To measure the importance, we use Fisher information (FI) which gives the {\em informative knowledge} of that kernel in handling adaptation task ~\cite{achille2019task2vec}.
Then, based on FI measurement, we classify kernels into important/unimportant.  
These kernel-level importance decisions are then used in the next step, i.e., main adaptation.

In the main adaptation step, we propose to apply {\em kernel modulation}
to achieve restrained update for the important kernels, and {\em simple fine-tuning} for the unimportant kernels. As will be discussed in Sec.~\ref{sec:low-rank}, the modulation is rank-constrained and has a restricted degree of freedom, therefore, it is capable of preserving the knowledge of the important kernels (also see results in our Supplement).
On the other hand, simple fine-tuning has a large degree of freedom for updating knowledge of the unimportant kernels.
Furthermore, the rank-constrained kernel modulation is parameter-efficient, therefore, we also apply this rank-constrained kernel modulation in the importance probing step to determine the importance of each kernel.

\subsection{Kernel Modulation}
\label{sec-kernel_modulation}
The Kernel ModuLation (KML) is used in the main adaptation step to preserve knowledge of important kernels in the adapted model. Furthermore, it is also used in the probing step as a parameter-efficient technique to determine importance of kernels.
Specifically, we apply KML which is proposed very recently 
\cite{milad2021revisit}.
In \cite{milad2021revisit}, KML is proposed for multi-modal few-shot classification (FSC).
In particular, in \cite{milad2021revisit},  KML is found to be effective for knowledge transfer between different {\em classification} tasks of different modes under few-shot constraints.
Therefore, in our work, we apply KML for knowledge transfer between different {\em generation} tasks of different domains under limited target domain samples.

Specifically, in each convolutional layer of a CNN, the \emph{i\textsuperscript{th}} kernel of that layer $\mathbf{W}_i \in \mathbb{R}^{c_{in} \times k\times k}$ is convolved with the input feature $\mathbf{X} \in \mathbb{R}^{c_{in} \times h \times w}$ to the layer to produce the \emph{i\textsuperscript{th}} output channel (feature map) $\mathbf{Y}_i \in \mathbb{R}^{ h' \times w'}$, i.e., $\mathbf{Y}_i = \mathbf{W}_i*\mathbf{X} + b_i$, where $b_i \in \mathbb{R}$ denotes the bias term. 
Then, KML modulates $\mathbf{W}_i$ by multiplying it with the modulation matrix $\mathbf{M}_i \in \mathbb{R}^{c_{in} \times k\times k}$ plus an all-ones matrix $\mathbf{J} \in \mathbb{R}^{c_{in} \times k\times k}$:
\begin{equation}
    \label{eq:kml}
    \hat{\mathbf{W}}_i = \mathbf{W}_{i} \odot (\mathbf{J} + \mathbf{M}_{i}),
\end{equation}
where $\odot$ denotes the Hadamard multiplication. 
In Eqn.~\ref{eq:kml}, using $\mathbf{J}$ allows learning the modulation matrix in a residual format.
Therefore, the modulation weights are learned as perturbations around the pretrained (and fixed) kernel which help to preserve the source knowledge. 
The exact pretrained kernel can also be transferred to the target model if it is optimal.

\subsection{
\small Parameter-Efficient KML via Low-rank Approximation}
\label{sec:low-rank}
The baseline KML in Sec.~\ref{sec-kernel_modulation} learns an individual modulation parameter for each coefficient of the kernel. Therefore, it could suffer from {\em parameter explosion} when using some recent GAN architectures (e.g., more than 58M parameters in StyleGAN-V2 \cite{karras2020styleganv2}
To address this issue, instead of learning the modulation matrix, we learn a {\em  low-rank} version of it
\cite{milad2021revisit,simon2020modulating}.
More specifically, for a Conv layer within CNN, with a total number of $d_{out}$ kernels to be modulated, instead of learning $\mathbf{M}=\{\mathbf{M}_i\}_{i=1}^{d_{out}}$, we learn two proxy vectors $\mathbf{m}_1 \in \mathbb{R}^{d_{out}}$, and $\mathbf{m}_2 \in \mathbb{R}^{(c_{in} \times k \times k)}$, and construct the modulation matrix using the outer product of these vectors, i.e., $\mathbf{M}=\mathbf{m}_1 \otimes \mathbf{m}_2 $. 
Furthermore, 
as we are using KML for adaptable knowledge preservation, we {\em freeze} the base kernel $\mathbf{W}_i$ during adaptation. Therefore, trainable parameters are $\mathbf{m}_1, \mathbf{m}_2$.

\begin{figure}[t]
    \centering
        \includegraphics[width=0.48\textwidth]{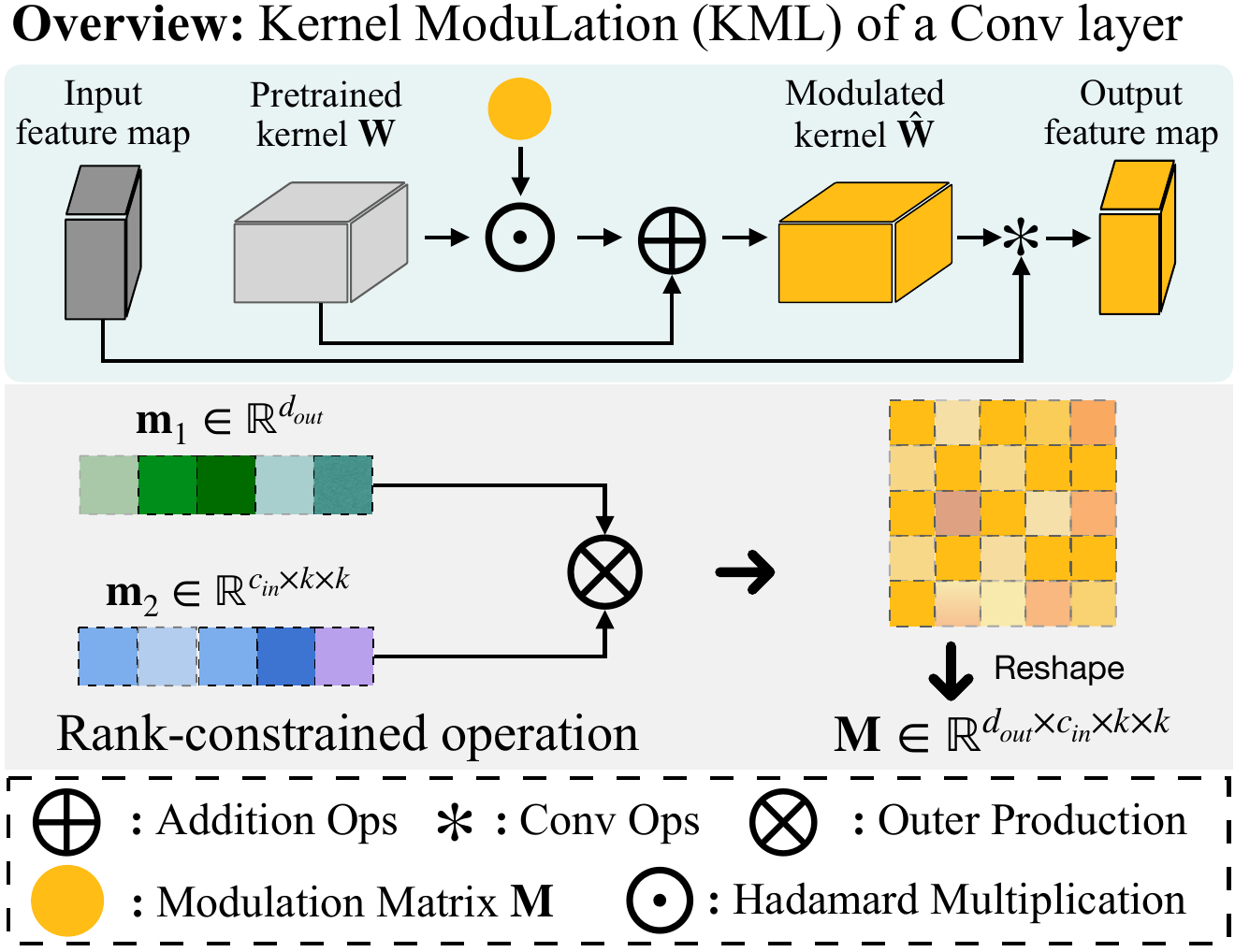}
    \caption{
    \textbf{Left:} Overview of KML on Conv layer. 
    \textbf{Right:} Parameter-efficient KML via rank-constraint operation. 
    For a Conv kernel with size $\mathbb{R}^{c^{out}\times c^{in} \times k \times k}$, the two learnable proxy vectors $\mathbf{m}_1 \in \mathbb{R}^{d_{out}}$ and $\mathbf{m}_2 \in \mathbb{R}^{(c_{in} \times k \times k)}$ where $d_{out} \leq c_{out}$. 
    We remark that the KML operation is parameter-efficient:
    For example, if the Conv kernel $\in \mathbb{R}^{512\times 512 \times 3 \times 3}$ (which is standard in StyleGAN-V2), the number of parameters learned via KML is up to $512+512\times 3 \times 3$ (when $d_{out} \eq c_{out}$), which is only $1/460$ of the entire Conv layer.
    }
    \label{fig:kml_ops}
    \vspace{-3mm}
\end{figure}

KML reduces the number of trainable parameters significantly and has better performance in restraining the update of important kernels  (results are in Supplement).
An illustration of our parameter-efficient KML operations is in Figure~\ref{fig:kml_ops}. 

As will be discussed in Sec.~\ref{sec-importance-measure}, the value of $d_{out}$ equals to the total number of kernels in a layer ($c_{out}$) during the importance probing process, and for the main adaptation, it is determined by the output of our probing method (i.e., $d_{out} \leq c_{out}$).

\begin{algorithm}[t]
\footnotesize
	\DontPrintSemicolon
	\SetAlgoLined
	

\KwRequire{Pre-trained GAN: $G_s$ and $D_s$, $iter_{probe}$, $iter_{adapt}$, 
threshold quantile $\boldsymbol{t}_\text{h}$,
learning rate $\alpha$
}

\KwIP{}

Freeze all kernels $\{ \mathbf{W}_i \}_{i=1}^{N}$ in pre-trained networks $G_s$, and $D_s$\\

Randomly initialize a modulation matrix $\mathbf{M}_i$ for each kernel $\mathbf{W}_i$ \\

\For{$k = 0$, $k{+}{+}$, \text{\bf while} $k < iter_{probe}$} 
{
Perform kernel modulation for all kernels using Eqn. \ref{eq:kml} to obtain modulated weights $\hat{\mathbf{W}}$\\
Update $\mathbf{M} \leftarrow \mathbf{M} - \alpha \nabla_{\mathbf{M}} \mathcal{L}(G(z);\hat{\mathbf{W}})$
\tcc{\small lightweight, i.e., $iter_{probe} << iter_{adapt}$}

}

Measure the importance of each kernel $\mathbf{W}_i$ by computing FI for the corresponding $\mathbf{M}_i$ using Eqn. \ref{eq:FI_vectors}\\

Compute the index set $\mathcal{A}$ of important kernels using quantile $\boldsymbol{t}_\text{h}$ of FI values as threshold \\ 

\KwMA{}

\eIf{$j \in \mathcal{A}$}
{Initialize the kernel by $\mathbf{W}_j$ and freeze the kernel, randomly initialize $\mathbf{M}_j$}
{Initialize the kernel by $\mathbf{W}_j$}

\For{$k = 0$, $k{+}{+}$, \text{\bf while} $k < iter_{adapt}$}
{
\eIf{$j \in \mathcal{A}$}
{Modulate kernel using Eqn. \ref{eq:kml} to obtain modulated weights $\hat{\mathbf{W}_j}$\\
Update $\mathbf{M}_j \leftarrow \mathbf{M}_j - \alpha \nabla_{\mathbf{M}_j} \mathcal{L}(G(z);\hat{\mathbf{W}})$}
{Update $\mathbf{W}_j \leftarrow \mathbf{W}_j - \alpha \nabla_{\mathbf{W}_j} \mathcal{L}(G(z);\hat{\mathbf{W}})$}
}

\caption{Few-Shot Image Generation via Adaptation-Aware Kernel Modulation (\textbf{AdAM})}
\label{alg:main}
\end{algorithm} 

\setlength{\tabcolsep}{0.75 mm}
\renewcommand{\arraystretch}{1.2}
\begin{table*}[t]
    \centering
    \small
    \caption{
    {FSIG (10-shot) results: We report FID scores ($\downarrow$) of AdAM (ours) and compare with existing methods for FSIG.
    We emphasize that Cat, Dog, and Wild target domains are 
    additional experiments included in this work
    (Sec \ref{sub-sec:proximity-analysis}).
    Our experiment results show two important findings: 
    \textbf{1)} Under setups where the assumption of close proximity between source and target domains is relaxed (Cat, Dog, Wild), SOTA FSIG methods -- EWC, CDC, DCL, RSSA, LLN --
    which consider only the source domain in knowledge preserving
    perform \textit{no better} than a baseline fine-tuning method (TGAN).
    \textbf{2)} Our proposed adaptation-aware AdAM achieves SOTA performance in all target domains due to preserving source domain knowledge that is important for few-shot target domain adaptation.
    We generate 5,000 images using the adapted generator to evaluate FID on the whole target domain.
    Additional results (e.g., KID) are in the Supplement.
    } 
    }\vspace{-3mm}
    \begin{adjustbox}{width=\textwidth,center}
        \setlength{\tabcolsep}{3mm}{\begin{tabular}{ l| c c c c c c c}
        \toprule[1.25pt]
        \textbf{Method}
        & \textbf{Babies} \cite{karras2018styleGANv1}
        & \textbf{Sunglasses} \cite{karras2018styleGANv1}
        & \textbf{MetFaces} \cite{karras2020ADA}
        & \textbf{Sketches} \cite{wang2008cuhk_sketches}
        & \textbf{AFHQ-Cat} \cite{choi2020starganv2}
        & \textbf{AFHQ-Dog} \cite{choi2020starganv2}
        & \textbf{AFHQ-Wild} \cite{choi2020starganv2}
        \\ 
        \hline
        TGAN \cite{wang2018transferringGAN} & $104.79{\footnotesize \pm 0.03}$ & $55.61{\footnotesize \pm 0.04}$ & $76.81{\footnotesize \pm 0.04}$ & $53.41{\footnotesize \pm 0.02}$ & $64.68{\footnotesize \pm 0.03}$ & $151.46{\footnotesize \pm 0.05}$ & $81.30{\footnotesize \pm 0.02}$ \\ 
        TGAN+ADA \cite{karras2020ADA} & $102.58{\footnotesize \pm 0.12}$ & $53.64{\footnotesize \pm 0.08}$ & $75.82{\footnotesize \pm 0.06}$ & $66.99{\footnotesize \pm 0.01}$ & $80.16{\footnotesize \pm 0.20}$ & $162.63{\footnotesize \pm 0.31}$ & $81.55{\footnotesize \pm 0.17}$ \\ 
        BSA \cite{noguchi2019BSA} & $140.34{\footnotesize \pm 0.01}$ & $76.12{\footnotesize \pm 0.01}$ & $93.42{\footnotesize \pm 0.03}$ & $69.32{\footnotesize \pm 0.02}$ &
        $154.62{\footnotesize \pm 0.05}$ &
        $158.32{\footnotesize \pm 0.04}$ &
        $168.12{\footnotesize \pm 0.07}$ \\
        FreezeD \cite{mo2020freezeD} & $110.92{\footnotesize \pm 0.02}$ & $51.29{\footnotesize \pm 0.05}$ & $73.33{\footnotesize \pm 0.07}$ & $46.54{\footnotesize \pm 0.01}$ & $63.60{\footnotesize \pm 0.08}$ & $157.98{\footnotesize \pm 0.28}$ & $77.18{\footnotesize \pm 0.13}$ \\
        MineGAN \cite{wang2020minegan} & $98.23{\footnotesize \pm 0.03}$ & $68.91{\footnotesize \pm 0.03}$ & $81.70{\footnotesize \pm 0.07}$ & $64.34{\footnotesize \pm 0.02}$ & 
        $70.98{\footnotesize \pm 0.04}$ & 
        $148.51{\footnotesize \pm 0.03}$ &
        $59.53{\footnotesize \pm 0.05}$ \\
        EWC \cite{li2020fig_EWC} & $87.41{\footnotesize \pm 0.02}$ & $59.73{\footnotesize \pm 0.04}$ & $62.67{\footnotesize \pm 0.09}$ & $71.25{\footnotesize \pm 0.01}$ & $74.61{\footnotesize \pm 0.17}$ & $158.78{\footnotesize \pm 0.14}$ & $92.83{\footnotesize \pm 0.15}$ \\
        CDC \cite{ojha2021fig_cdc} & $74.39{\footnotesize \pm 0.03}$ & $42.13{\footnotesize \pm 0.04}$ & $65.45{\footnotesize \pm 0.08}$ & $45.67{\footnotesize \pm 0.02}$ & $176.21{\footnotesize \pm 0.09}$ & $170.95{\footnotesize \pm 0.11}$ & $135.13{\footnotesize \pm 0.10}$  \\ 
        DCL \cite{zhao2022dcl} & $52.56{\footnotesize \pm 0.02}$ & $38.01{\footnotesize \pm 0.01}$ & $62.35{\footnotesize \pm 0.07}$ & $37.90{\footnotesize \pm 0.02}$ & $156.82{\footnotesize \pm 0.04}$ & $171.42{\footnotesize \pm 0.14}$ & $115.93{\footnotesize \pm 0.09}$ \\ 
        RSSA \cite{xiao2022rssa} & $75.67{\footnotesize \pm 0.39}$ & $44.35{\footnotesize \pm 0.06}$ & $57.06{\footnotesize \pm 0.07}$ & $54.58{\footnotesize \pm 0.51}$ & $166.89{\footnotesize \pm 0.06}$ & $158.20{\footnotesize \pm 0.34}$ & $162.80{\footnotesize \pm 0.48}$ \\
        LLN \cite{mondal2023ill} & $63.31{\footnotesize \pm 0.05}$ & $35.64{\footnotesize \pm 0.15}$ & $87.21{\footnotesize \pm 0.05}$ & \bm{$35.59{\footnotesize \pm 0.13}$} & $209.95{\footnotesize \pm 0.12}$ & $216.31{\footnotesize \pm 0.77}$ & $254.89{\footnotesize \pm 0.31}$ \\
        \hline
        \textbf{AdAM (Ours)} & \bm{$48.83{\footnotesize \pm 0.03}$} & \bm{$28.03{\footnotesize \pm 0.07}$} & \bm{$51.34{\footnotesize \pm 0.06}$} & \bm{$36.44{\footnotesize \pm 0.29}$} & \bm{$58.07{\footnotesize \pm 0.13}$} & \bm{$100.91{\footnotesize \pm 1.01}$} & \bm{$36.87{\footnotesize \pm 0.23}$} \\
        \bottomrule[1.25pt]
        \end{tabular}}
    \end{adjustbox}
\label{table:fid_scores}
\end{table*}

\subsection{Importance Measurement}
\label{sec-importance-measure}
Recall our FSIG has two main steps: (i)  importance probing step (Lines 1-8 in Algorithm~\ref{alg:main}), and (ii) main
adaptation step (Lines 9-21 in Algorithm~\ref{alg:main}).
In probing, we also apply KML as a parameter-efficient technique to determine the importance of individual kernels.
In particular, for probing, 
we propose to apply KML to all kernels (in both generator and discriminator) to identify which of the {\em modulated} kernels are important for the adaptation task. To measure the importance of the modulated kernels, we apply Fisher information  (FI) to the modulation parameters.
In our FSIG setup, for a modulated GAN with parameters $\Theta$, 
Fisher information $\mathcal{F}$
can be computed as:
\begin{equation}
    \label{eq:fisher_information}
    \mathcal{F}(\Theta) = \mathbb{E} \big[- \frac{\partial^2}{\partial\mathbf{\Theta}^2} \mathcal{L}(x|\Theta) \big],
\end{equation}
where $\mathcal{L}(x|\Theta)$ is the binary cross-entropy loss  computed using the output of the discriminator, and $x$ includes few-shot target samples, and fake samples generated by GAN.
Then, FI for a modulation matrix $\mathcal{F}(\mathbf{M}_i)$ can be computed by averaging over FI values of parameters within that matrix. 
As we are using the low-rank estimation to construct the modulation matrix, we can estimate $\mathcal{F}(\mathbf{M}_i)$ by FI values of the proxy vectors (i.e., $\mathbf{m}_1$ and $\mathbf{m}_2$).
In particular, considering the outer product in low-rank approximation, we have $\mathbf{M}_i = ([\mathbf{m}_1^i\mathbf{m}_2^{1}, \dots, \mathbf{m}_1^i\mathbf{m}_2^{(c_{in}\times k \times k)}])\texttt{.reshape()}$, where $|\mathbf{m}_2| = c_{in}\times k \times k$.
Then we use the unweighted average of FI for parameters of $\mathbf{m}_1$ and $\mathbf{m}_2$, proportional to their occurrence frequency in the calculation of $\mathbf{M}_i$, as an estimate of $\mathcal{F}(\mathbf{M}_i)$ (we discuss more details in Supplement):
\begin{equation}
\label{eq:FI_vectors}
    \hat{\mathcal{F}}(\mathbf{M}_i) =
    \mathcal{F}(\mathbf{m}_1^i) + \frac{1}{|\mathbf{m}_2|} \sum_{j=1}^{|\mathbf{m}_2|}\mathcal{F}(\mathbf{m}_2^j).
\end{equation}
After calculating $\hat{\mathcal{F}}(\mathbf{M}_i)$ for all modulation matrices in both generator and discriminator, we use the  $\boldsymbol{t}_\text{h}$(\%) quantile of these values as a threshold 
to decide whether modulation of a kernel is important or unimportant for adaptation to the target domain.
If the modulation of a kernel is determined to  be important (during probing), the kernel is modulated using KML during the main adaptation step; otherwise, the kernel is updated using simple fine-tuning during the main adaptation.
In all setups, we perform probing for 500 iterations.
We remark that in probing only modulation parameters  $\mathbf{m}_1, \mathbf{m}_2$  are trainable, and FI is only computed on them, therefore the probing is a very lightweight step and can be performed with minimal overhead (details are in Supplement).
The output of the probing step is the decision to apply kernel modulation or simple fine-tuning on individual kernels.
Then, based on these decisions, the main adaptation is performed.
Overall, our proposed method, AdAM, is summarized in Algorithm~\ref{alg:main}.

\tocless
\section{Empirical Studies}
\label{section5}

\subsection{Implementation Details} 
For a fair comparison, we strictly follow prior works  \cite{wang2018transferringGAN, mo2020freezeD, li2020fig_EWC, ojha2021fig_cdc, zhao2022dcl, xiao2022rssa, mondal2023ill} in the choice of GAN architecture, source-target adaptation setups, and hyper-parameters. 
We use StyleGAN-V2 \cite{karras2020styleganv2} as the GAN architecture and FFHQ as the source domain. 
Our experiments include setups with different source-target proximity: Babies/Sunglasses \cite{ojha2021fig_cdc}, MetFaces \cite{karras2020ADA} and Cat/Dog/Wild (AFHQ) \cite{choi2020starganv2} (See Sec.~\ref{sec:proximity-transferability}).
Adaptation is performed with 256x256 resolution, batch size 4 with initial learning rate 0.002 on a single Tesla V100 GPU. We apply importance probing and modulation on the base kernels of both the generator and the discriminator. 
We focus on 10-shot target adaptation setup in the main experiments, with additional setups in our analysis and ablation studies, see Sec.~\ref{section6} and also Supplement.

\subsection{Qualitative Results}
 We show generated images with our proposed AdAM along with baseline \cite{wang2018transferringGAN, mo2020freezeD} and SOTA methods \cite{li2020fig_EWC, ojha2021fig_cdc, zhao2022dcl, xiao2022rssa, mondal2023ill} for two target domains, Babies and AFHQ-Cat with different degrees of proximity to FFHQ, before and after adaptation. 
 The results are shown in Figure~\ref{fig4} top and bottom, respectively.
 By preserving source domain knowledge that is important for the target domain, our proposed adaptation-aware FSIG method can generate substantially high-quality images with high diversity for both the Babies and Cat domains. We also include FID ($\downarrow$) \cite{heusel2017FID} and Intra-LPIPS ($\uparrow$) \cite{ojha2021fig_cdc} (for measuring diversity) in Figure~\ref{fig4} to quantitatively show that our proposed method outperforms existing SOTA FSIG methods \cite{li2020fig_EWC, ojha2021fig_cdc, zhao2022dcl, xiao2022rssa, mondal2023ill}.  

\subsection{Quantitative Results}
\begin{figure*}[!h]
     \centering
     \hspace{-1.9mm}
     \includegraphics[width=0.935\textwidth]{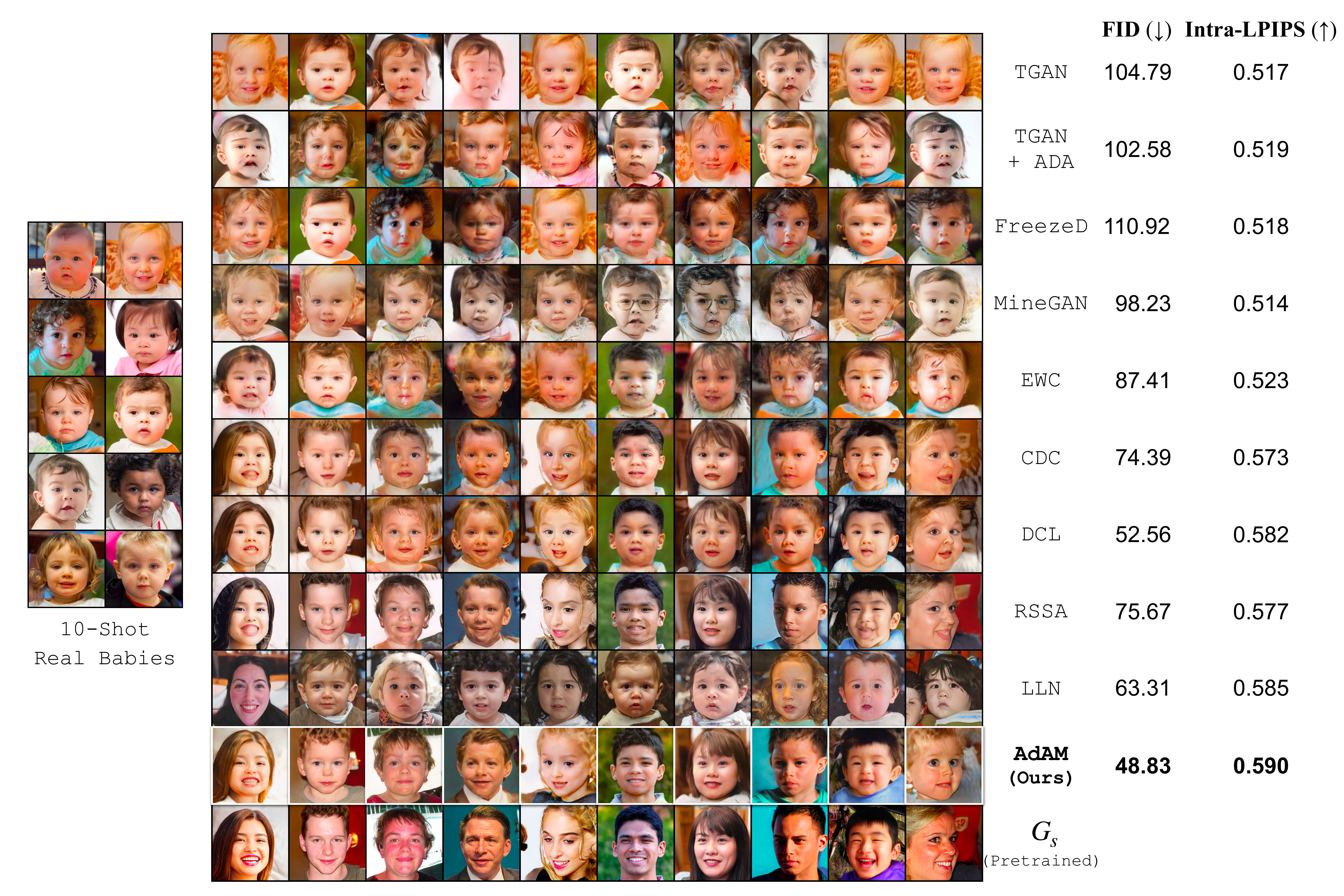}
     \includegraphics[width=0.935\textwidth]{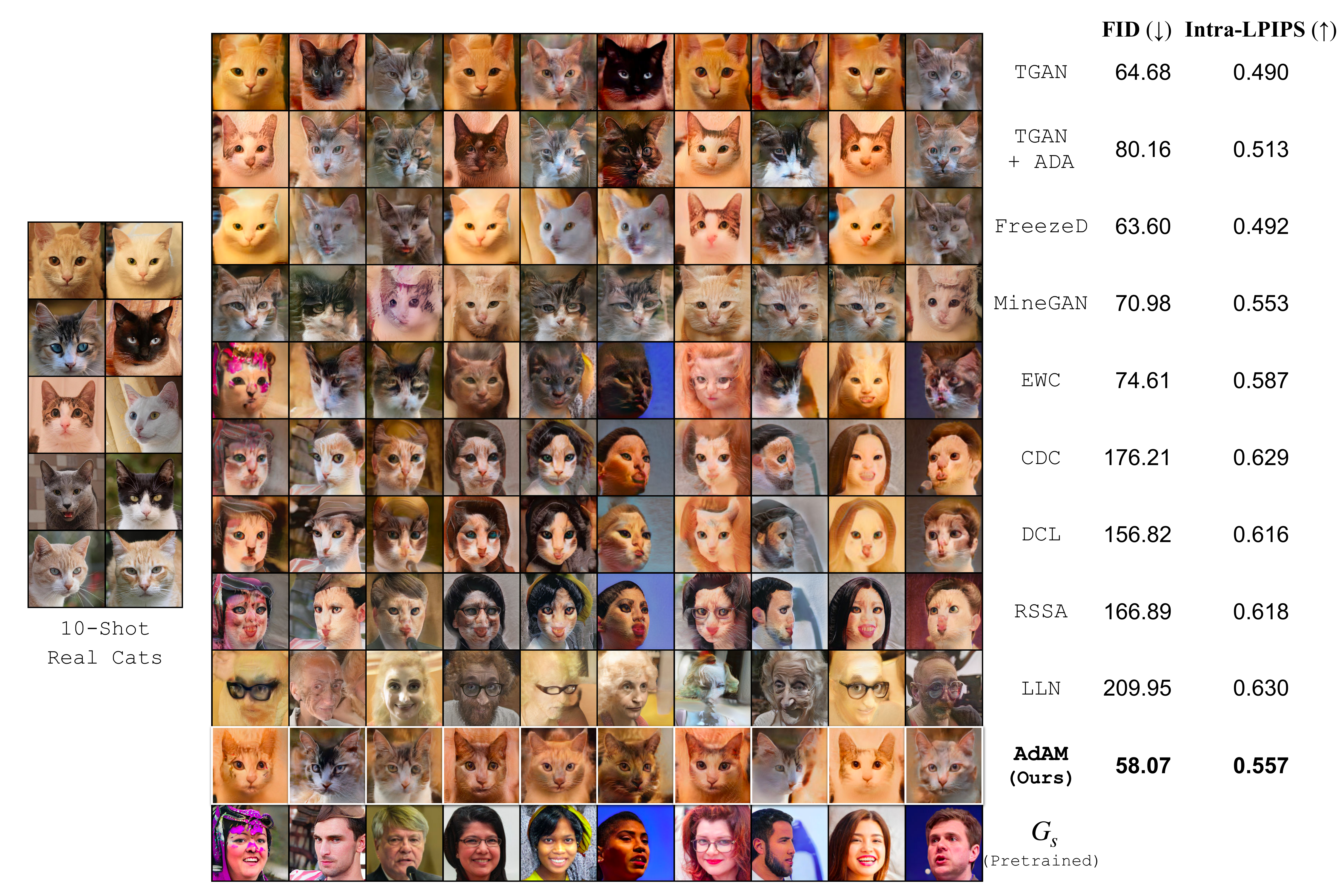}
     \vspace{-2mm}
     \caption{
     Qualitative and quantitative comparison of 10-shot image generation with different FSIG methods.
     Images of each column are from the same noise input (except for LLN \cite{mondal2023ill} where the images are randomly sampled, due to LLN directly learning a latent code in $\mathcal{W}^{+}$ space). 
     For target domains with close proximity (e.g. Babies, \textbf{top}), our method can generate high-quality images with more refined details and diverse knowledge, achieving the best FID and Intra-LPIPS score.
     For target domain that is distant (e.g., Cat, \textbf{bottom}), TGAN/FreezeD overfit to the 10-shot samples and others fail. In contrast, our method preserves meaningful semantic features at different levels (e.g., posture and color) from the source, achieving a good trade-off between the quality and diversity of the generated images. 
     }
     \label{fig4}
     \vspace{-3mm}
 \end{figure*}

We show complete FID ($\downarrow$) scores (with standard deviation computed by running three times) in Table~\ref{table:fid_scores}.
Our proposed AdAM for FSIG achieves SOTA results across all target domains of varying proximity to the source (FFHQ).
We emphasize that it is achieved by preserving source domain knowledge that is important for target domain adaptation (Sec.~\ref{section4}). We also report Intra-LPIPS ($\uparrow$) as an indicator of diversity, as Figure~\ref{fig4}.

\begin{figure*}[h]
    \centering
    \small
    \includegraphics[width=\textwidth]{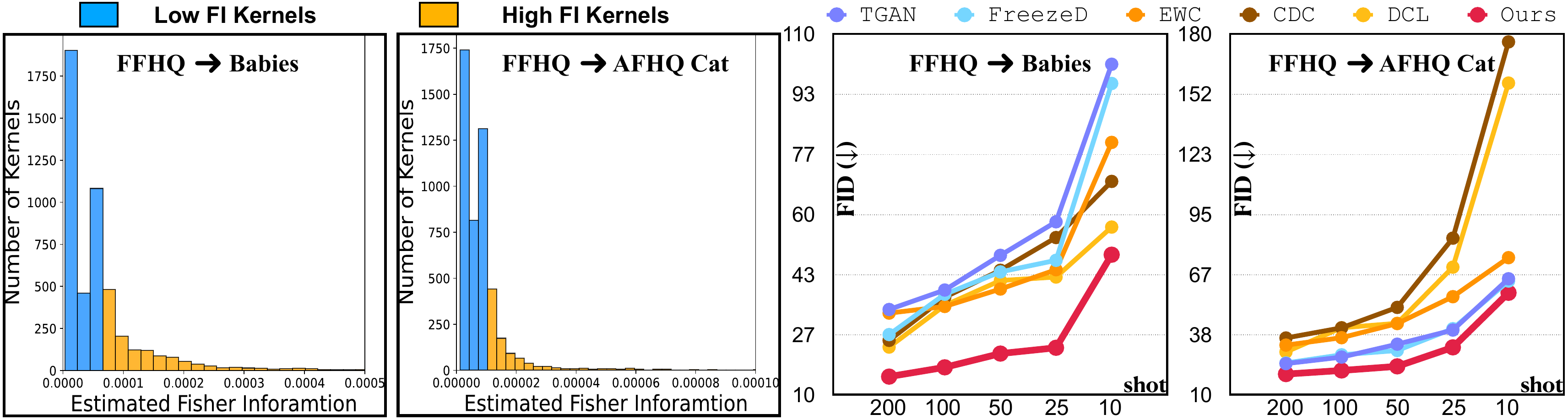}\\
    \vspace{2mm}
    \setlength{\tabcolsep}{3.6mm}{
        \begin{tabular}{l| c c c c c c c c }
        \toprule
        \textbf{Target Domain FID ($\downarrow$)}
         & \textbf{Babies}
         & \textbf{Sunglasses}
         & \textbf{MetFaces}
        & \textbf{AFHQ-Cat}
        & \textbf{AFHQ-Dog}
        & \textbf{AFHQ-Wild}
         \\ 
        \hline
        \textbf{AdAM} (w/o IP) & 54.46 & 33.66  & 60.43 & 82.41 & 160.87 & 81.24    \\
        \textbf{AdAM} (Ours) & {\bf 48.83} & {\bf 28.03} & {\bf 51.34} & {\bf 58.07} & {\bf 100.91} & {\bf 36.87}    \\
        \bottomrule
        \end{tabular}
    }
    \caption{
    \textbf{(Top Left)} Our proposed IP (in AdAM) identifies and preserves source kernels important (high FI) for target adaptation. \textbf{(Bottom)} FID score on different datasets. We validate the effectiveness of IP by modulating all kernels without IP. On the other hand, if we fine-tune all parameters without IP and modulation (TGAN), it suffers mode collapse (Table~\ref{table:fid_scores} and Figure~\ref{fig4}).     
    \textbf{(Top Right)} We evaluate the performance with different number of shots (10, 25, 50, 100, 200) on Babies and AFHQ-Cat. We show that our method consistently outperforms other FSIG methods in all setups. The complete comparison is in Table~\ref{table:number-of-shots}. 
    In our Supplement, we also show the comparison of generated images given more target domain images (e.g. 100-shot) during adaptation.
    }
    \label{table:ablation}
    \vspace{-2mm}
\end{figure*}

We remark that the goal of importance probing (denoted as ``IP'') is to identify kernels that are 
important for \textit{few-shot target adaptation} as shown in Figure~\ref{table:ablation} (Top). 
To justify the effectiveness of our design choice, we perform an ablation study that discards the IP stage and regards all kernels as \textit{equally important} for target adaptation. 
Therefore, we simply modulate all kernels \textit{without any knowledge selection}. 
As one can observe from Figure~\ref{table:ablation} (Bottom), knowledge selection plays a vital role in adaptation performance. Specifically, the significance of knowledge preservation is more evident when the target domains are distant from the source domain.

\renewcommand{\arraystretch}{1}


\tocless
\begin{figure*}[t]
    \centering
        \includegraphics[width=0.9\textwidth]{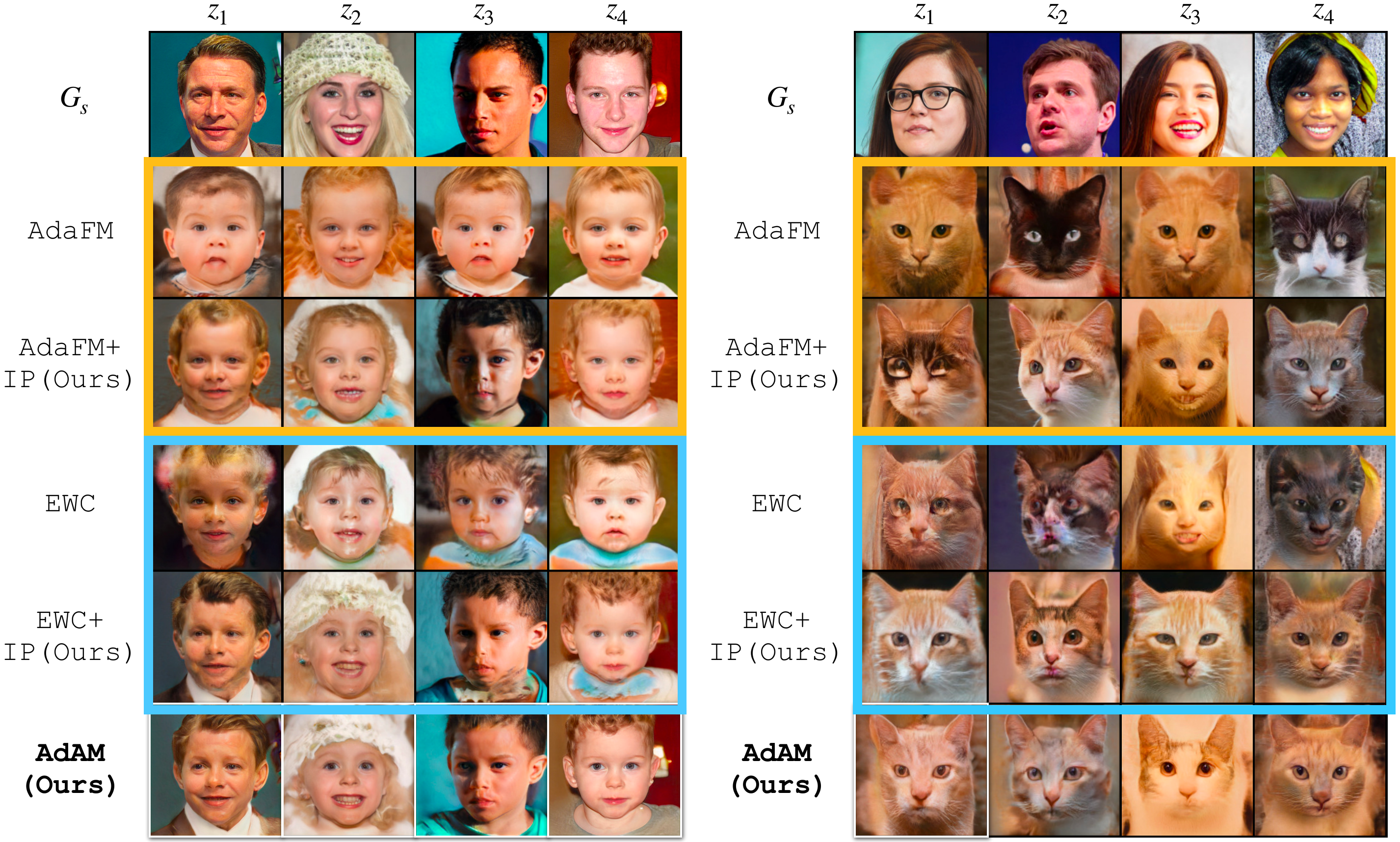}
    \caption{
    \textbf{Qualitative results of the effectiveness of Importance Probing (IP)}.
    $G_s$ is the source generator (FFHQ).
    We show results for our major setups, FFHQ $\rightarrow$ Babies (left) and FFHQ $\rightarrow$ Cat (right).
    Applying IP to EWC \cite{li2020fig_EWC}, AdaFM \cite{cong2020gan_memory}, we observe better quality in FSIG. 
    This shows that our proposed idea of probing the importance of kernels for FSIG is principally a suitable approach to improve FSIG on various methods.
    One can also observe that images generated by our proposed method with KML, e.g. the last row, have good quality compared to other methods. Our observation is quantitatively confirmed in Table~\ref{table-supp:ablation_results}.
    }
    \label{fig:ablation_ip}
    \vspace{-2mm}
\end{figure*}

\section{Analysis and Ablation Studies}
\label{section6}

\subsection{Effectiveness of Importance Probing}

In this section, we conduct extensive and comprehensive ablation studies to show the significance of our proposed method for FSIG. Similar to the main experiments in Sec.~\ref{section5}, we use FFHQ \cite{karras2020styleganv2} as the source domain, and use Babies and Cat \cite{choi2020starganv2} as target domains. 
The different approaches in this study are as follows:

  \noindent\textbf{TGAN \cite{wang2018transferringGAN}:} The source GAN models pretrained on FFHQ are updated using {\em simple fine-tuning} of all parameters with the 10-shot target samples.

  \noindent\textbf{EWC \cite{li2020fig_EWC}:} 
  Following \cite{li2020fig_EWC}, an L2 regularization is applied to all model weights to augment simple fine-tuning. The regularization is scaled by the importance of individual model weights as determined by the 
  FI of the model weights based on the {\em source} models.

  \noindent\textbf{EWC + IP:} We apply our probing idea on top of EWC. In the probing step, original EWC as discussed above is used but with a small number of iterations. At the end of probing, the FI of model weights based on the {\em updated} models is computed.
  Then, during main adaptation, this {\em target-aware} FI is used to scale the L2 regularization. In other words, EWC + IP is a target-aware version of EWC in \cite{li2020fig_EWC} using our probing idea.

  \noindent\textbf{AdaFM \cite{cong2020gan_memory}:} AdaFM (see Sec. \ref{section2}) is applied to all kernels. 
  
  \noindent\textbf{AdaFM + IP:} We apply our probing idea on top of AdaFM. In the probing step, the original AdaFM as discussed above is used but with a small number of iterations. 
  At the end of probing, the FI of AdaFM parameters is computed, and kernels are classified as important/unimportant using the same quantile threshold $\boldsymbol{t}_\text{h}$\% as in our work. 
  Then, during the main adaptation, the important kernels are updated via AdaFM, and the unimportant kernels are updated via simple fine-tuning. In other words, AdaFM + IP is a target-aware version of AdaFM using our probing idea.

  \noindent\textbf{Ours w/o IP (i.e. main adaptation only):} KML modulation is applied to all kernels.

  \noindent\textbf{Ours w/ Freeze:} We apply our probing idea as discussed in Sec. \ref{section4}, i.e., with KML applied to all kernels but adaptation with a small number of iterations.
  At the end of probing, FI of KML parameters is computed, and kernels are classified as important/unimportant using the same quantile threshold $\boldsymbol{t}_\text{h}$\% as in our work. 
  Then, during the main adaptation, the important kernels are {\em frozen}, and the unimportant kernels are updated via simple fine-tuning. In other words, this is similar to our proposed method except that kernel freezing is used in the main adaptation instead of KML for important kernels.

  \noindent\textbf{Ours w/ KML (i.e. AdAM):} 
  We apply our probing idea as discussed in Sec. \ref{section4}, i.e., with KML applied to all kernels but adaptation with a small number of iterations.
  At the end of probing, FI of KML parameters is computed, and kernels are classified as important/unimportant using quantile threshold $\boldsymbol{t}_\text{h}$\% (to be discussed in Sec. \ref{sec:kml_threshold}). 
  Then, during the main adaptation, the important kernels are modulated by KML, and the unimportant kernels are updated via simple fine-tuning. 

\textbf{Qualitative Results.}
We show generated images corresponding to approaches discussed above in Figure~\ref{fig:ablation_ip}. These results show that our proposed idea on importance probing is principally a suitable approach to improve FSIG by identifying kernels important for target domain adaptation. Figure
\ref{fig:ablation_ip} also shows that our proposed method can generate images with better quality.

\renewcommand{\arraystretch}{1}
\begin{table}[t]  
    \caption{
    Quantitative results for IP: 
    For each method, the best FID and LPIPS results are shown in \textbf{bold}.
    IP is performed for 500 iterations (where relevant).
    These results show that our proposed IP is principally a suitable approach for FSIG. This can be clearly observed when applying IP to EWC \cite{li2020fig_EWC} (EWC+IP) and AdaFM \cite{cong2020gan_memory} (AdaFM+IP).
    We also observe that methods performing IP at kernel level (Ours w/ KML, AdaFM + IP) perform better than the method performing IP at parameter level (EWC + IP).
    Overall, we quantitatively show that our proposed method outperforms all existing FSIG methods with IP, thereby generating images with high quality (FID $\downarrow$) and good diversity (Intra-LPIPS $\uparrow$).
    }
    \begin{adjustbox}{width=0.48\textwidth}
    \begin{tabular}{l| c c | cc }
    \toprule
    \multirow{2}{*}{\textbf{Method}}
     & \multicolumn{2}{c|}{\textbf{FFHQ $\rightarrow$ Babies}}
     & \multicolumn{2}{c}{\textbf{FFHQ $\rightarrow$ Cat}}
     \\ 
     & {FID ($\downarrow$)} & {Intra-LPIPS ($\uparrow$)} & {FID ($\downarrow$)} & {Intra-LPIPS ($\uparrow$)}  \\
            \hline
    TGAN \cite{wang2018transferringGAN} & 104.79  & 0.517 & 64.68 & 0.490 \\\hline
    EWC \cite{li2020fig_EWC} & 87.41 & 0.521 & 74.61 & \textbf{0.587} \\
    EWC + [IP (Ours)] & \textbf{70.80} & \textbf{0.625} & \textbf{66.35} & 0.540 \\\hline
    AdaFM \cite{cong2020gan_memory} & 62.90 & 0.568 & 64.44 &  0.525 \\
    AdaFM + [IP (Ours)] & \textbf{55.64} & \textbf{0.577} & \textbf{60.04} & \textbf{0.540}  \\\hline
    Ours w/o IP & 54.46 & \textbf{0.613} & 82.41 &  0.522 \\
    Ours w/ Freeze [w/ IP] & 50.81 & 0.581 & 61.60 & 0.559  \\
    \textbf{AdAM} (w/ KML [w/ IP]) & \textbf{48.83} & 0.590 & \textbf{58.07} & 0.557 \\
    \bottomrule
    \end{tabular}
    \end{adjustbox}
    \label{table-supp:ablation_results}
\end{table}

\begin{figure}[h]
    \centering
    \includegraphics[width=0.49\textwidth]{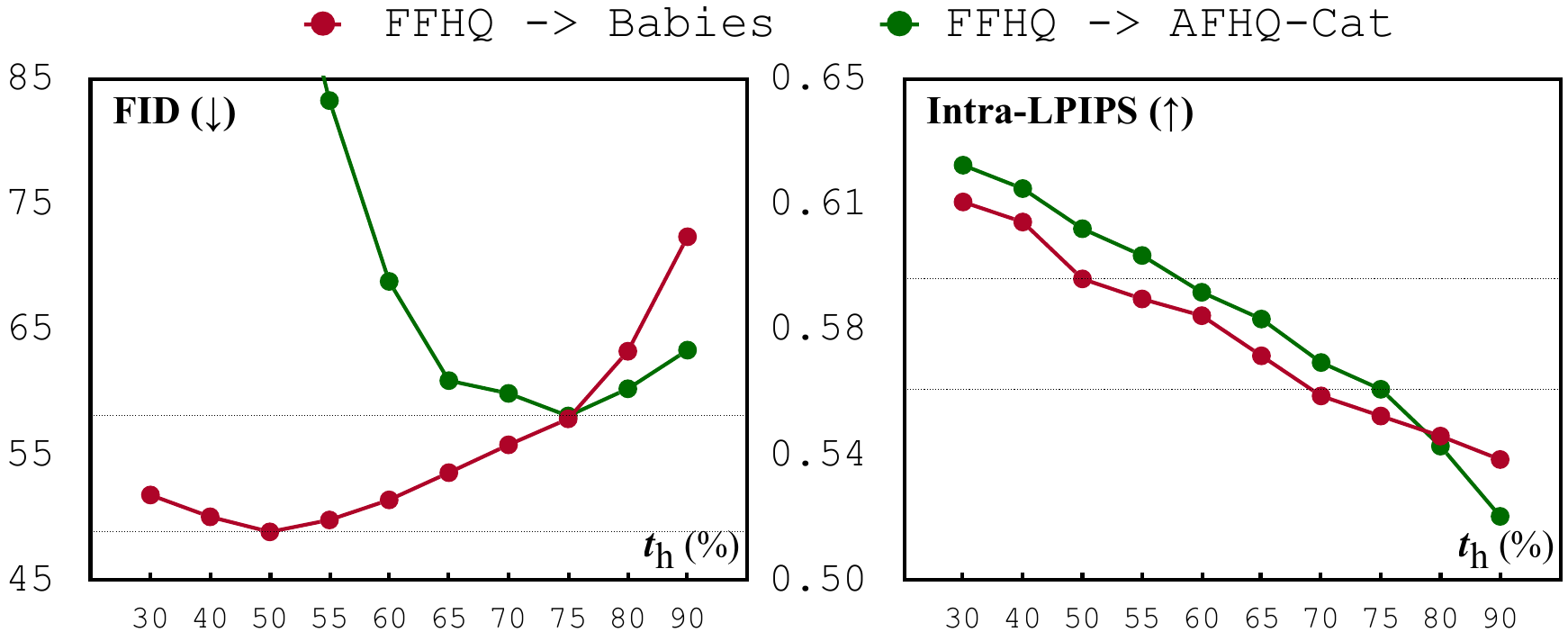}
    \caption{
    Evaluation of the performance by varying the thresholds for preserving important kernels. 
    In practice, we choose $\boldsymbol{t}_\text{h}$=50\% for close target domains (e.g., Babies) and $\boldsymbol{t}_\text{h}$=75\% for remote target domains (e.g., Cats) which empirically obtain the best performance.
    }
    \label{fig:kml_threshold}
    \vspace{-2mm}
\end{figure}

\textbf{Quantitative results.}
We show FID ($\downarrow$) / LPIPS ($\uparrow$) results in Table~\ref{table-supp:ablation_results}, which demonstrates our proposed IP is principally a suitable approach for FSIG. 
This can be clearly observed when applying IP to EWC \cite{li2020fig_EWC} and AdaFM \cite{cong2020gan_memory}.
We remark that probing with KML (AdAM) is computationally more efficient compared to probing with EWC and AdaFM due to less trainable parameters (see Supplement).
Overall, we show that our proposed method outperforms existing FSIG methods with IP, thereby generating images with a good balance between quality (FID $\downarrow$) and diversity (Intra-LPIPS $\uparrow$).
We also empirically observe that methods performing IP at kernel level (AdAM, AdaFM + IP) perform better than methods performing IP at parameter level (EWC + IP).

\subsection{Threshold of Preserving Important Kernels}
\label{sec:kml_threshold}
Our proposed AdAM for FSIG aims to preserve different amount of source knowledge that is useful for target adaptation. 
Specifically, we preserve filters that are deemed important for target adaptation by modulating them via parameter-efficient KML (see Figure~\ref{fig:kml_ops}). 
We select the high-importance filters by using a quantile $\boldsymbol{t}_\text{h}$(\%) as a threshold to determine the importance of a kernel. 
In this section, we conduct a study to show the effectiveness and impact of preserving different amount of filters that are deemed to be most relevant for target adaptation.

As shown in Figure~\ref{fig:kml_threshold}, varying the amount of filters for preservation improves the performance in different ways.
In practice, we select $\boldsymbol{t}_\text{h}$=50\% for FFHQ $\mapsto$ Babies and $\boldsymbol{t}_\text{h}$=75\% for FFHQ $\mapsto$ AFHQ-Cat, and this choice is also intuitive: for target domains that are semantically closer to the source, preserving more source knowledge might be useful for few-shot adaptation.

\subsection{Number of Target Samples (shots)}
\label{subsec-number-of-shots}
The number of target domain training samples is an important factor that can impact the FSIG performance. In general, more target domain samples can allow better estimation of the target distribution.
We empirically study the efficacy of our proposed method under different number of target domain samples.
A direct comparison is shown in Figure~\ref{table:ablation} and the complete quantitative results are in Table~\ref{table:number-of-shots}. 
We show that our proposed adaptation-aware FSIG method consistently outperforms existing SOTA methods in different source $\xrightarrow{}$ target setups with different proximity, even when adapting the model to the entire target domain samples.

\subsection{Alternative Importance Measure}
\label{subsec-supp:class_saliency}
In literature, Class Salience \cite{simonyan2013deep} (CS) is used as a property to explain which area/pixels of an input image stand out for a specific classification decision. Similar to the estimated Fisher Information (FI) used in our work, the complexity of CS is based on the first-order derivatives. Therefore, conceptually CS could have a connection with FI as they both use the knowledge encoded in the gradients.

We perform an experiment to replace FI with CS in importance probing and compare it with our original approach. Note that, in \cite{simonyan2013deep}, CS is computed w.r.t. input image pixels. To make CS suitable for our problem, we modify it and compute CS w.r.t. modulation parameters. Similar to our approach in Sec.~\ref{section4}, we average the importance of all parameters within a kernel to calculate the importance of that kernel. Then we use these values during our importance probing to determine the important kernels for adapting from source to target domain (as Sec.~\ref{section4}). The results in Table~\ref{table:cs-fi} are obtained with our proposed method using FI and CS during importance probing.

Our results suggest that importance probing using FI (approximated by first-order derivatives) can perform better in the selection of important kernels, leading to better performance (FID, intra-LPIPS) in the adapted models as shown in Table~\ref{table:cs-fi}.





\begin{table}[t]  
    \caption{
    {
    We conduct comprehensive experiments to evaluate the performance, i.e., FID ($\downarrow$) of different few-shot generation methods, given a different number of shots for adaptation: from one shot to the entire dataset. The images for adaptation are randomly sampled and kept the same for all methods for fair comparison. The results are as below.
    \textbf{Top:} FFHQ $\rightarrow$ Babies. 
    \textbf{Bottom:} FFHQ $\rightarrow$ Cat.}
    }
   \centering
        \begin{adjustbox}{width=0.49\textwidth}
        \begin{tabular}{c| c c c c |c }
        \toprule
        \textbf{Number of Shots}
         & TGAN & TGAN+ADA & EWC & CDC & AdAM \\        \hline
         1    & 172.49 & 188.84 & 104.50 & 88.86 & \textbf{77.71}       \\ 
         5    & 108.65 & 105.19 & 88.51 & 78.29 & \textbf{52.85}        \\
         10   & 104.79 & 102.58 & 87.41 & 74.39 & \textbf{48.83}        \\
         25   & 57.94 & 58.86 & 44.67 & 53.58 & \textbf{23.05}          \\
         50   & 48.65 & 52.18 & 39.32 & 44.52 & \textbf{21.44}          \\
         100  & 39.04 & 45.71 & 34.49 & 37.04 & \textbf{17.63}          \\
         200  & 33.65 & 38.84 & 32.65 & 25.1 & \textbf{15.06}           \\
         500  & 27.21 & 26.31 & 28.11 & 22.53 & \textbf{14.71}          \\
         1000 & 25.68 & 25.30 & 25.55 & 21.79 & \textbf{14.12}          \\
         All Samples ($\sim$ 2700) & 25.03 & 25.47 & 24.57 & 21.91 & \textbf{13.59}   \\
        \end{tabular}
        \end{adjustbox}
        \begin{adjustbox}{width=0.49\textwidth}
        \begin{tabular}{c| c c c c |c }
        \toprule
        \toprule
        \textbf{Number of Shots}
         & TGAN & TGAN+ADA & EWC & CDC & AdAM \\        \hline
         1 & 125.52 & 125.81 & 139.11 & 197.79 & \textbf{118.25} \\ 
         5 & 90.24 & 86.94 & 136.65 & 180.34 & \textbf{79.53} \\
         10 & 64.68 & 80.16 & 74.61 & 176.21 & \textbf{58.07} \\
         25 & 40.52 & 48.61 & 56.23 & 83.80 & \textbf{32.38} \\
         50 & 33.87 & 35.76 & 43.58 & 51.20 & \textbf{26.43} \\
         100 & 27.78 & 28.16 & 36.93 & 41.58 & \textbf{21.50} \\
         200 & 24.73 & 26.78 & 33.43 & 36.79 & \textbf{19.79} \\
         500 & 20.25 & 19.01 & 32.73 & 30.81 & \textbf{14.80} \\
         1000 & 17.20 & 16.46 & 31.50 & 28.50 & \textbf{16.80} \\
         All Samples ($\sim$ 5000) & 10.52 & 9.56 & 18.76 & 20.53 & \textbf{6.52} \\
        \bottomrule
        \end{tabular}
        \end{adjustbox}
    \label{table:number-of-shots}
\end{table}

\subsection{Discussion: How much can the proximity between source and target domain be relaxed?}
\label{sec-supp:rebuttal_proximity_relaxation}
In this section, we discuss the proximity limitation between source domain $\mathcal{S}$ and target domain $\mathcal{T}$ in our experiment setups.
First, we remark that the upper bound on proximity between $\mathcal{S}$ and $\mathcal{T}$ could be conditioning on \textbf{(a):} the number of available samples (shots) from the target domain, and \textbf{(b):} the method used for knowledge transfer.

\textbf{(a):} Proximity bound conditioning on the number of target domain samples. In this paper, we focus on few-shot setups, e.g. 10 shots. However, with more target domain samples available, proximity between $\mathcal{S}$ and $\mathcal{T}$ can be further relaxed, and the proximity bound would increase, i.e. for a given generative model on $\mathcal{S}$, we could learn an adapted model for $\mathcal{T}$ which is more distant. Intuitively, increasing the number of target domain samples can provide more diverse knowledge for $\mathcal{T}$, and as a result, there is less reliance on the knowledge of $\mathcal{S}$ that is generalizable for $\mathcal{T}$ (which would decrease as $\mathcal{S}$ and $\mathcal{T}$ are more apart). In the limiting cases when abundant target domain samples are available, knowledge of $\mathcal{S}$ would not be critical, and proximity constraints between $\mathcal{S}$ and $\mathcal{T}$ may be totally relaxed (ignored).

\begin{table}[t]  
    \caption{
    {
    Results of different importance measurements.
    In this experiment, we replace Fisher Information (FI) with Class Saliency (CS) \cite{simonyan2013deep} in importance probing and compare with our original approach. We evaluate the performance under different source $\rightarrow$ target adaptation setups.}
    }
   \centering
        \begin{adjustbox}{width=0.48\textwidth}
        \begin{tabular}{l| c c | cc }
        \toprule
        \multirow{2}{*}{\textbf{Importance Measure}}
         & \multicolumn{2}{c|}{\textbf{FFHQ $\rightarrow$ Babies}}
         & \multicolumn{2}{c}{\textbf{FFHQ $\rightarrow$ Cat}}
         \\ 
         & {FID ($\downarrow$)} & {Intra-LPIPS ($\uparrow$)} & {FID ($\downarrow$)} & {Intra-LPIPS ($\uparrow$)}  \\
                \hline
        Class Salience \cite{simonyan2013deep}  & 52.46 & 0.582 & 61.68 & 0.556 \\
        Fisher Information (Ours) & \textbf{48.83} & \textbf{0.590} & \textbf{58.07} & \textbf{0.557} \\
        \bottomrule
        \end{tabular}
        \end{adjustbox}
    \label{table:cs-fi}
\end{table}

\begin{table}[t]
\small
    \caption{
    We conduct experiments for {FFHQ ($\mathcal{S}$) $\rightarrow$ Cars} (Remote domain $\mathcal{T}$) adaptation and evaluate the performance in such a challenging setup. We show that our method can achieve similar diversity as ADA \cite{karras2020ADA} and the overall performance (FID) is better than other baseline and SOTA methods.
    }
   \centering
        \begin{tabular}{l| c c}
        \toprule
        \multirow{2}{*}{\textbf{Method}}
         & \multicolumn{2}{c}{\textbf{FFHQ $\rightarrow$ Cars} (Remote domain)}
         \\ 
         & {FID ($\downarrow$)} & {Intra-LPIPS ($\uparrow$)} \\
                \hline
        Training from Scratch & 201.34 & 0.300 \\ 
        TGAN+ADA \cite{karras2020ADA} & 171.98 & 0.438 \\
        EWC \cite{li2020fig_EWC} & 276.19 & \textbf{0.620} \\
        CDC \cite{ojha2021fig_cdc} & 109.53 & 0.484 \\
        DCL \cite{zhao2022dcl} & 125.96 & 0.464 \\
        AdAM (Ours) & \textbf{80.55} & 0.425
        \\
        \bottomrule
        \end{tabular}
    \label{table:ffhq_cars}
\end{table}

\textbf{(b):} Proximity bound conditioning on the knowledge transfer method. Given a generative model pretrained on $\mathcal{S}$ and a certain number of available samples from $\mathcal{T}$, the method used for knowledge transfer plays a critical role. If the method is superior in identifying suitable transferable knowledge from $\mathcal{S}$ to $\mathcal{T}$, the proximity between $\mathcal{S}$ and $\mathcal{T}$ can be relaxed, and the proximity bound would increase. In our work, our first contribution is to reveal that existing SOTA approaches (which are based on target-agnostic ideas) are inadequate in identifying transferable knowledge from $\mathcal{S}$ to $\mathcal{T}$. As a result, when proximity between $\mathcal{S}$ and $\mathcal{T}$ is relaxed, the performance of the adapted models is miserably poor, as discussed in Sec.~\ref{sec:proximity-transferability}, Sec.~\ref{section5} and Supplement. 
Therefore, our second contribution is to propose a target-aware approach that could identify more meaningful transferable knowledge from $\mathcal{S}$ to $\mathcal{T}$, allowing relaxation of the proximity constraint.

\begin{figure*}[!h]
\vspace{-2mm}
    \centering
    \includegraphics[width=0.96\textwidth]{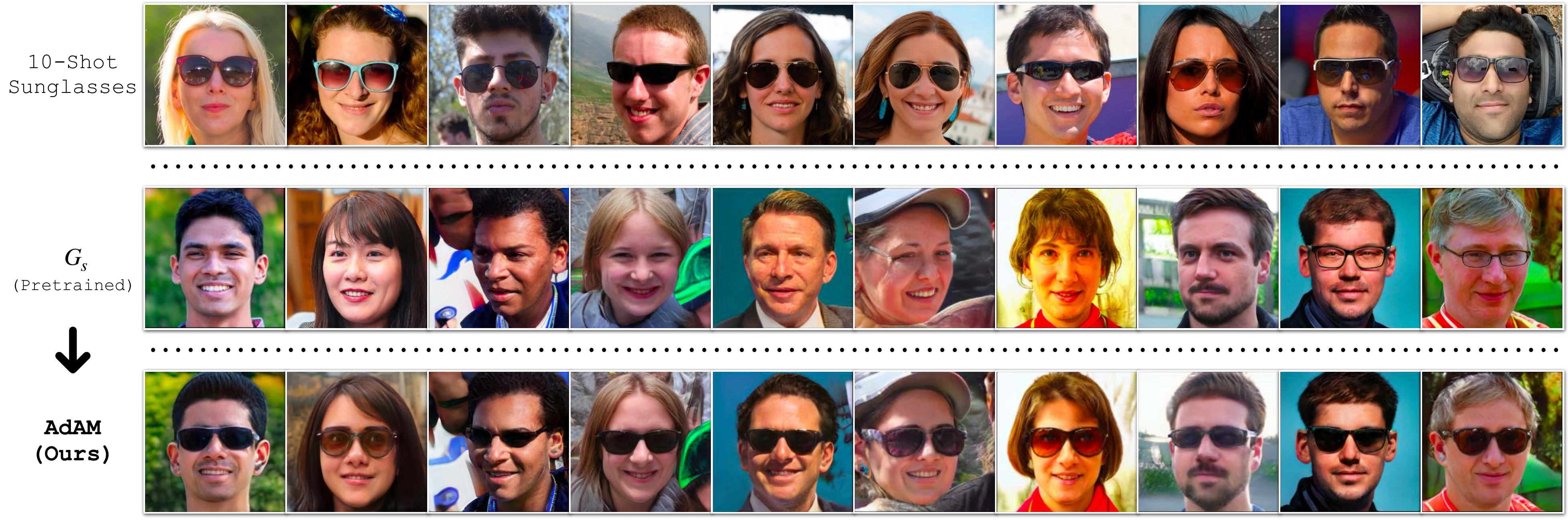}
    \vspace{1mm}
    \includegraphics[width=0.96\textwidth]{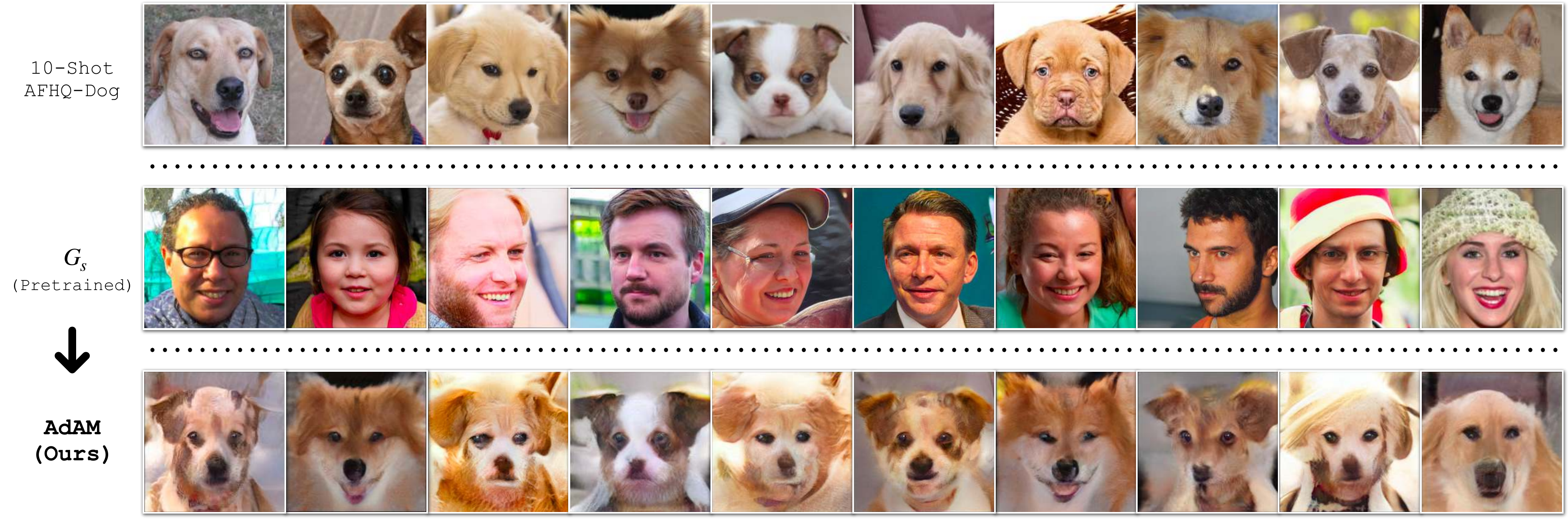}
    \vspace{1mm}
    \includegraphics[width=0.96\textwidth]{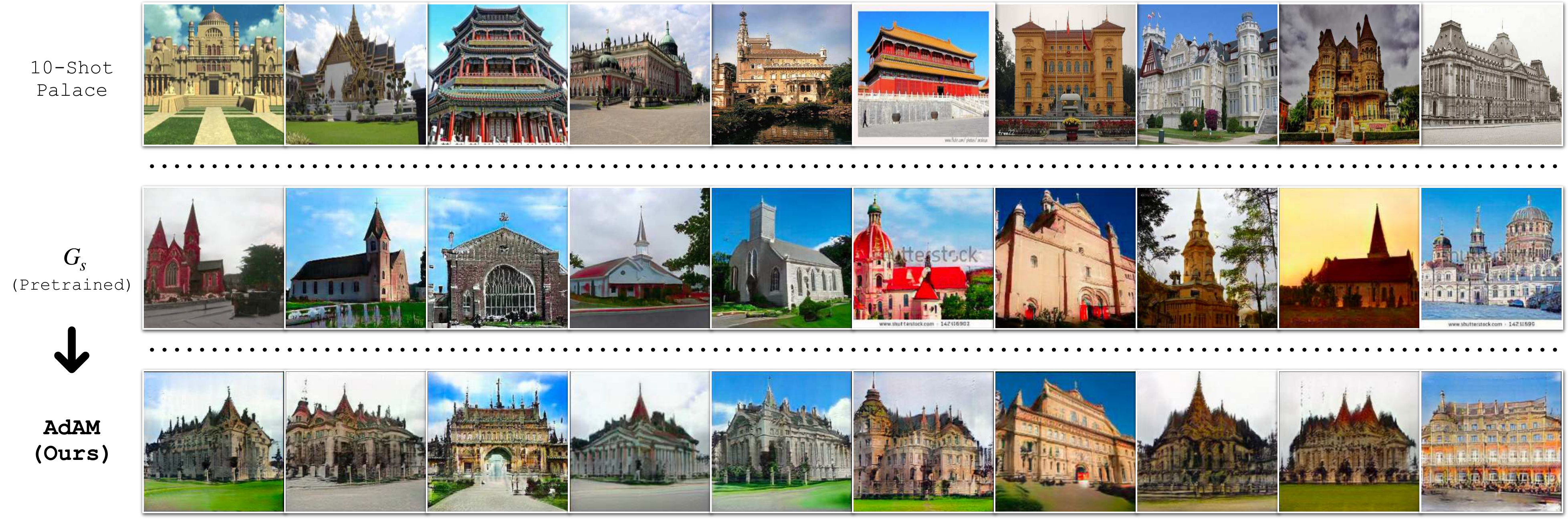}
    \vspace{1mm}
    \includegraphics[width=0.96\textwidth]{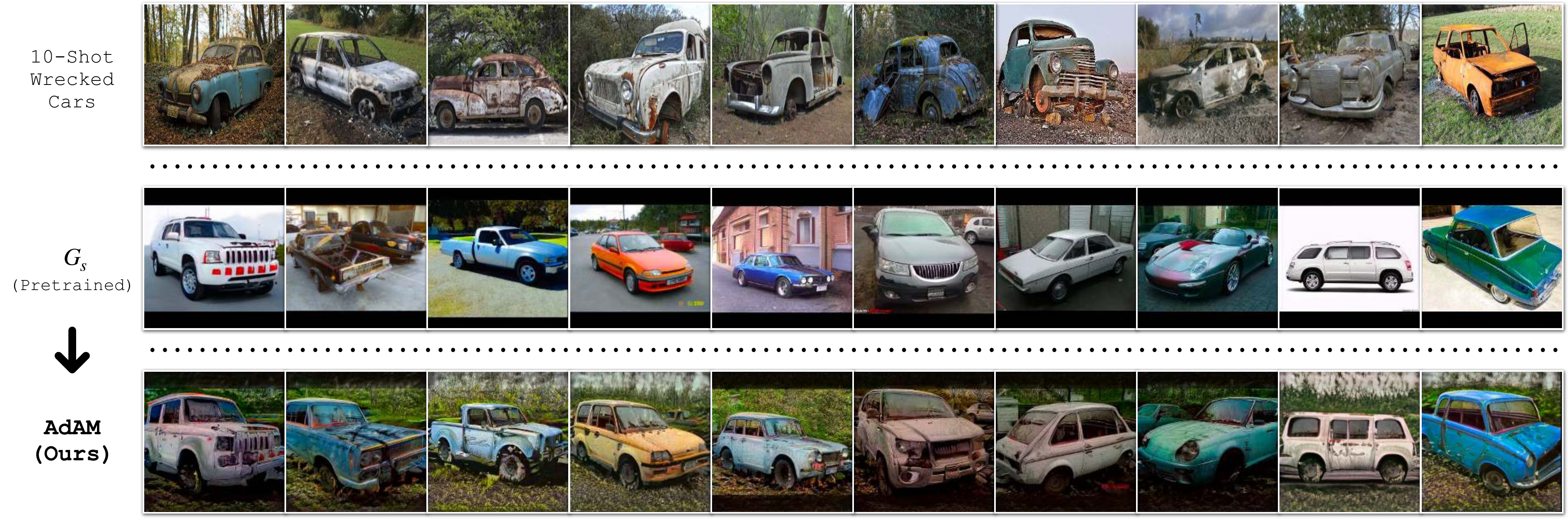}
    \vspace{-3mm}
    \caption{
    Few-shot adaptation results. We show that AdAM can adapt the source GAN to target domains with different proximity to the source given only few-shot samples.
    1. FFHQ $\rightarrow$ Sunglasses.
    2. FFHQ $\rightarrow$ AFHQ-Dog.
    3. LSUN Church $\rightarrow$ Palace.
    4. LSUN Cars $\rightarrow$ Wrecked Cars.
    Additional adaptation results with more source/target domains are in our Supplement.
    }
    \label{fig:add_domains}
\end{figure*}

In this section, we provide experimental results for the adaptation between two very distant domains: FFHQ $\rightarrow$ Cars using only 10 shots, aiming to answer two main questions: 
(1) Is there transferable knowledge from FFHQ to Cars for the FSIG task? (2) How does AdAM (our) compare with other methods in this setup? 
For this, in addition to transfer learning approaches discussed in the paper, we also add the results for training from scratch using only the same 10 Car samples. The results are in Table~\ref{table:ffhq_cars}. 

\begin{figure*}[t]
    \centering
    \includegraphics[width=\textwidth]{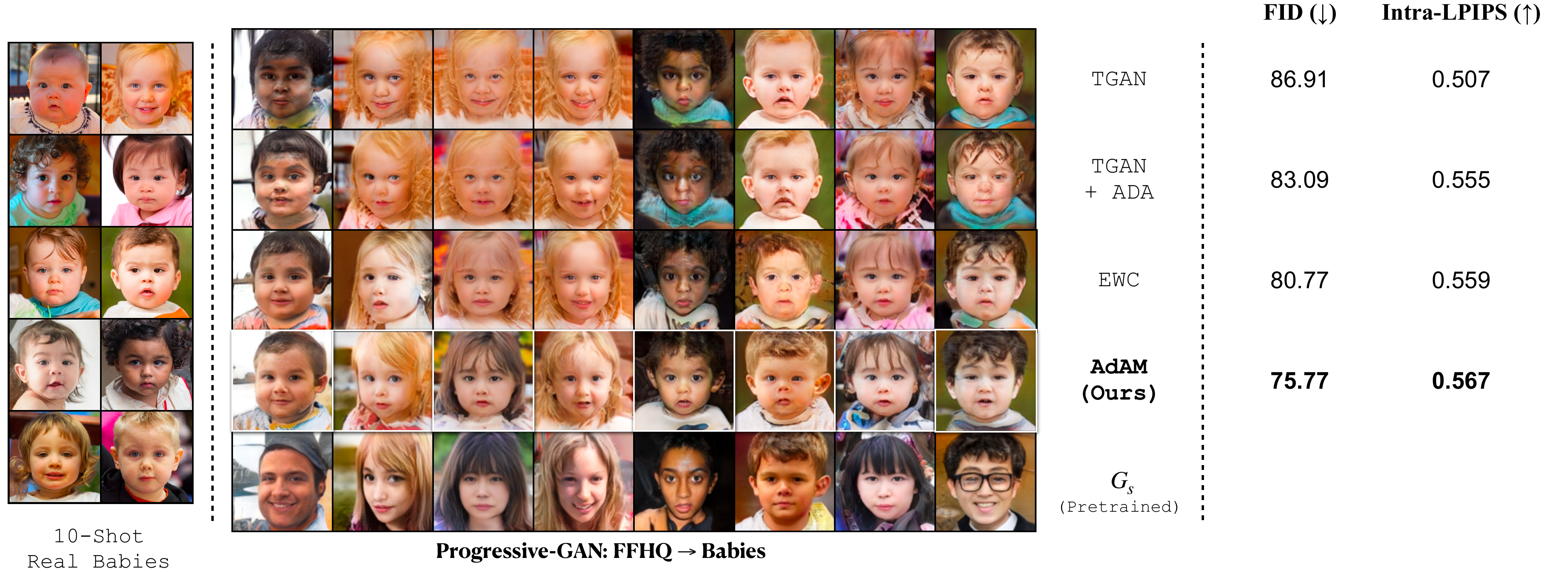}
    \includegraphics[width=\textwidth]{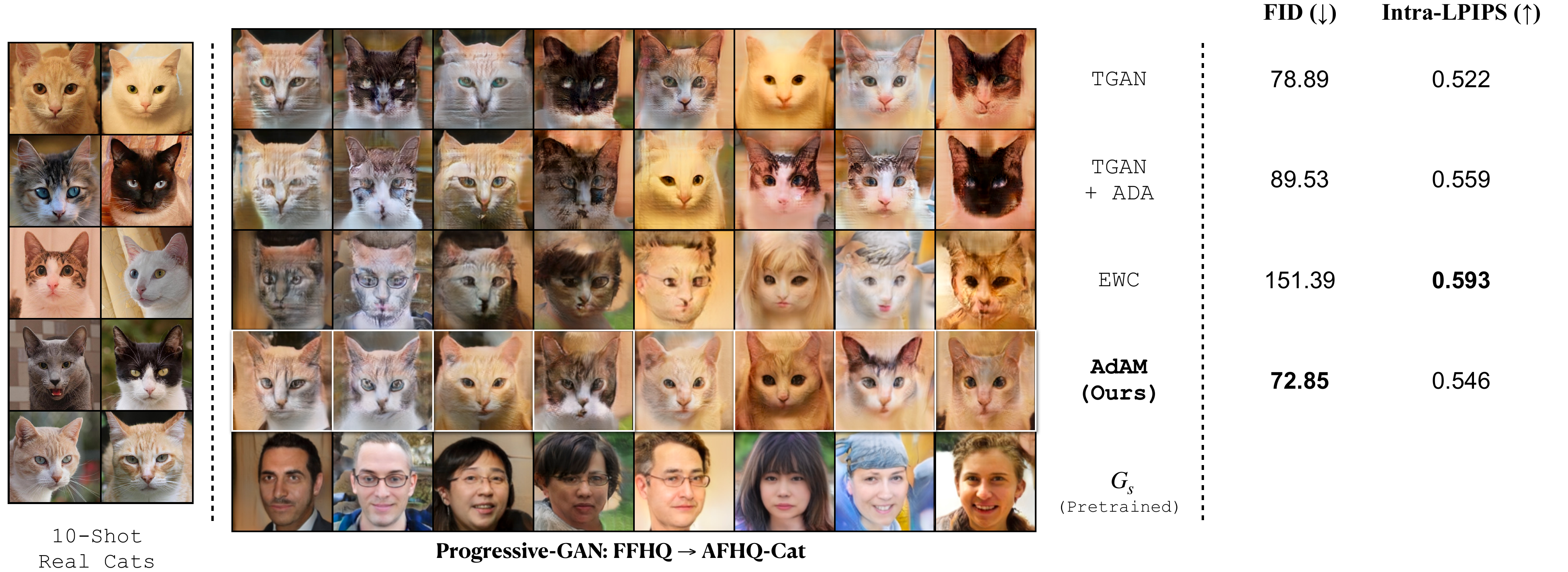}
    \caption{
    Qualitative and quantitative 10-shot adaptation results of FFHQ $\rightarrow$ Babies and FFHQ $\rightarrow$ Cat using pre-trained ProGAN \cite{karras2017progan}. 
    As one can observe, our proposed method outperforms existing FSIG methods. 
    More results are in the Supplement.
    }
    \label{fig:main_progan}
\end{figure*}

\subsection{Additional Source $\xrightarrow{}$ Target Results}
\label{subsec-supp:additional_source_target_domains}
In this section, we show additional 10-shot adaptation results for our proposed adaptation-aware FSIG method 
for target domains with different proximity. 
Visualization and comparison results are shown in Figure~\ref{fig:add_domains} and in our Supplement.
As one can observe, 
SOTA FSIG methods \cite{li2020fig_EWC, ojha2021fig_cdc, zhao2022dcl} are unable to adapt well to distant target domains due to only considering source task in knowledge preservation, while TGAN \cite{wang2018transferringGAN} suffers severe mode collapse. In contrast, AdAM (our) outperforms these methods and consistently produces high-quality images with good diversity.

\subsection{Additional GAN Architectures}
\label{subsec-supp:additional_gan_architectures}

We use an additional pre-trained GAN architecture, ProGAN \cite{karras2017progan}, to conduct FSIG experiments for FFHQ $\rightarrow$ Babies, FFHQ $\rightarrow$ Cat, Church $\rightarrow$ Haunted houses and Church $\rightarrow$ Palace setups. For fair comparison, we strictly follow the exact experiment setup discussed in Sec.~\ref{section5}.
We show qualitative and quantitative results for FFHQ $\rightarrow$ Babies, FFHQ $\rightarrow$ Cat adaptation in Figure~\ref{fig:main_progan} and Table~\ref{table:progan-results}. 
As one can observe, our proposed method consistently outperforms other baseline and SOTA FSIG methods with another pre-trained GAN model (ProGAN \cite{karras2017progan}), demonstrating the effectiveness and generalizability of our method. 
We also show qualitative results and analysis for Church $\rightarrow$ Haunted houses and Church $\rightarrow$ Palace adaptation in Supplement.



\begin{table}[t]  
    \caption{
    {
    Performance with additional GAN architecture.
    In this experiment, we use ProGAN \cite{karras2017progan} as the GAN model and compare it with other baseline and SOTA methods. We evaluate the performance under different source $\rightarrow$ target adaptation setups.}
    }
   \centering
        \begin{adjustbox}{width=0.48\textwidth}
        \begin{tabular}{l| c c | cc }
        \toprule
        \multirow{2}{*}{\textbf{Method}}
         & \multicolumn{2}{c|}{\textbf{FFHQ $\rightarrow$ Babies}}
         & \multicolumn{2}{c}{\textbf{FFHQ $\rightarrow$ Cat}}
         \\ 
         & {FID ($\downarrow$)} & {Intra-LPIPS ($\uparrow$)} & {FID ($\downarrow$)} & {Intra-LPIPS ($\uparrow$)}  \\
                \hline
        TGAN \cite{wang2018transferringGAN} & 86.91  & 0.507 & 78.89 & 0.522 \\
        TGAN + ADA \cite{karras2020ADA} & 83.09 & 0.555 & 89.53 & 0.559 \\
        EWC \cite{li2020fig_EWC} & 80.77 & 0.559 & 151.39 & \textbf{0.593} \\
        AdAM (Ours) & \textbf{75.77} & \textbf{0.567} & \textbf{72.85} & {0.546} \\
        \bottomrule
        \end{tabular}
        \end{adjustbox}
    \label{table:progan-results}
\end{table}

\tocless
\section{Explainability and Effectiveness of Knowledge Transfer in AdAM}
\label{section7}

In this section, we attempt to discover what form of visual information is encoded/generated by a specific high FI kernel identified by our importance probing algorithm in AdAM, and how it is preserved after target adaptation in FSIG.

\subsection{What form of visual information is encoded by high FI kernels in AdAM?}
This is a natural and complex problem and to our best knowledge, methods of visualizing generative models/GANs are still rather restrictive in terms of concepts or knowledge that can be visualized. 
Nevertheless, we leverage GAN Dissection \cite{bau2019gan}, a more established visualization method and explainable tool to visualize the internal representations that are highly correlated to important (high FI) kernels identified by our probing algorithm.

\textbf{Experiment setup:} We use LSUN-Church as the source domain, this is because the official GAN Dissection method \cite{GAN-dissection-web}
is more suitable for \textit{scene-based} image generators (and due to the limitation of the semantic segmentation used in the GAN Dissection pipeline \cite{bau2019gan}). We use the Palace from ImageNet-1K as the target domain. 
Following official GAN Dissection implementation \cite{bau2019gan}, we use the ProGAN \cite{karras2017progan} model. For a fair comparison, we strictly follow the exact experiment setup discussed in Sec.~\ref{section5} and Sec.~\ref{section6}.

\begin{figure*}[ht]
    \centering
    \includegraphics[width=\textwidth]{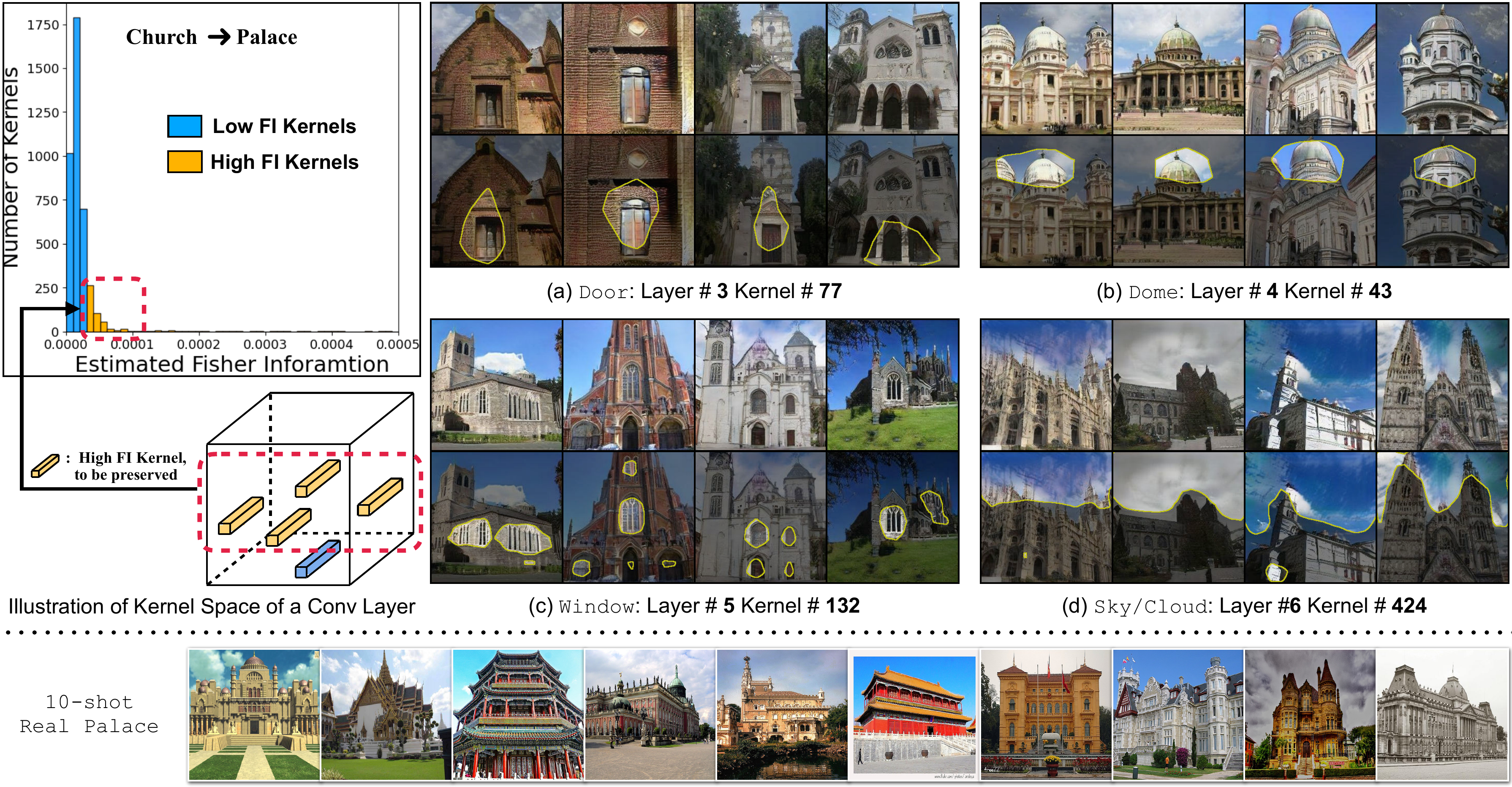}
    \vspace{-6mm}
    \caption{
    \textbf{Top:}
    Visualizing high FI kernels using GAN Dissection \cite{bau2019gan} for Church $\rightarrow$ Palace 10-shot adaptation.
    (Left)
    Illustration of kernel space of a Conv layer in the source generator and the estimated FI of source kernels for the target (Palace) FSIG adaptation.
    (Right)
    The visualization of different high FI kernels and the corresponding distinct semantics. 
    The first row shows different images generated by the source generator, and the second row highlights the concept encoded by the corresponding high FI kernel as determined by GAN Dissection.
    We observe that a notable amount of high FI kernels correspond to useful source domain concepts including (a) \texttt{building}, (b) \texttt{dome}, (c) \texttt{window},  and (d) \texttt{sky/cloud}, which are preserved when adapting to the target domain (Palace in this figure). 
    \textbf{Bottom:} 
    10-shot real Palace images as target samples.
    \textbf{Importantly}, we remark that the concepts encoded by high FI kernels determined by AdAM are useful and transferrable to the target domain adaptation.
    Best viewed in color.
    }
    \label{fig:gan_dissection}
\end{figure*}

\textbf{Results}:
Visualizing high FI kernels for Church $\rightarrow$ Palace adaptation: 
The results for importance estimation (via FI) for kernels and several distinct semantic concepts learned by high FI kernels are shown in Figure~\ref{fig:gan_dissection}.
We visualize four examples of high FI kernels that correspond to concepts of building, dome, window, and sky/cloud, respectively. 
Using GAN Dissection, we observe that a notable amount of high FI kernels correspond to useful source domain concepts which are preserved when adapting to the Palace domain. 
We remark that these preserved concepts encoded by high FI kernels (and determined by AdAM) are useful to the target domain adaptation (See the bottom side of Figure~\ref{fig:gan_dissection}). 

\textbf{Transferability of high FI kernels for FSIG:}
We remark that the distinct semantic concepts (examples in Figure~\ref{fig:gan_dissection}) encoded by the high FI kernels, and identified/preserved by AdAM, are effectively transferred to the target domain after few-shot adaptation, leading to the high quality and diversity of generated images of the resulting target generator.
We further note that our observation is consistent with different GAN architectures and various source $\rightarrow$ target adaptation setups, e.g., StyleGAN-V2 in Figure~\ref{fig:ablation_ip}, Figure~\ref{fig:add_domains}, and ProGAN results in Figure~\ref{fig:main_progan}. Additional results are included in our Supplement.

\subsection{Limitations and Future Work of GAN Dissection} 
In our experiments, 
although GAN Dissection can uncover useful semantic concepts and knowledge preserved by high FI kernels, GAN Dissection method \cite{bau2019gan} is limited by the datasets and models used for semantic segmentation. 
Hence this method is not able to uncover concepts that are not present in the semantic segmentation dataset (They use Broaden Dataset \cite{bau2017network}). Therefore, using GAN dissection we are currently unable to discover and visualize more fine-grained concepts preserved by our high FI kernels. 
On the other hand, the GAN dissection method \cite{bau2019gan} is built on top of ProGAN image generators, and it is still challenging to dissect image generators with more complex and advanced structures. 
Nevertheless, we remark that we have consistent observations on dissection/transferability results that demonstrate the effectiveness of our proposed method for FSIG tasks. 

\tocless
\section{Discussion}
\label{section8}

\begin{figure*}[t]
    \centering
    \includegraphics[width=\textwidth]{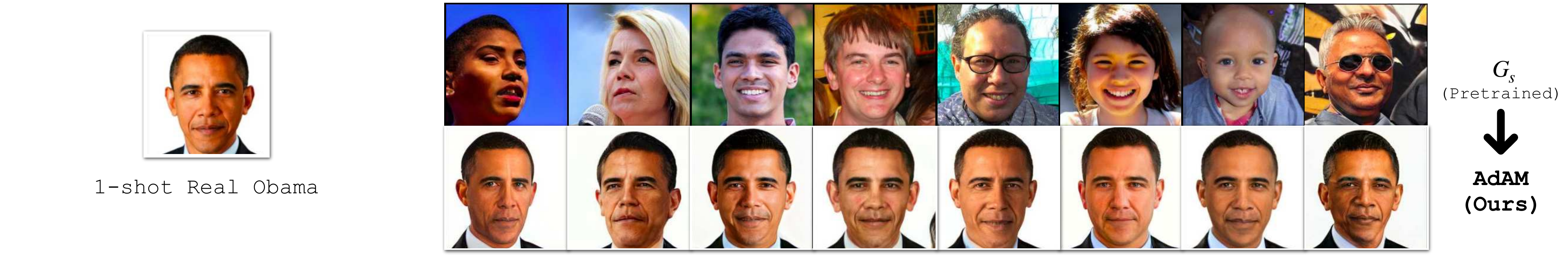}
    \caption{
    FFHQ $\rightarrow$ Obama photo (from Obama Dataset \cite{zhao2020differentiable}).
    We demonstrate that, given very few photo(s), our algorithm can generate diverse images of a particular person, similar to some recent popular models, e.g., DreamBooth \cite{ruiz2022dreambooth}. 
    Consequently, we urge researchers and practitioners to attend to safety and ethical concerns before using/deploying our method. 
    More details in Sec.~\ref{sec-supp:societal_impact}.
    }
    \label{fig:supp_obama}
\end{figure*}

\subsection{Conclusions}
Focusing on FSIG, we make two contributions.
First, we revisit current SOTA methods and their experiments. 
We discover that SOTA methods perform poorly in setups when source and target domains are more distant, as existing methods only consider the source domain/task for knowledge preservation. Second, we propose a new FSIG method which is target/adaptation-aware (AdAM). Our proposed method outperforms previous work across all setups of different source-target domain proximity.
We include rich extended experiments and analysis in the Supplement.

\subsection{Limitations and Future Works}
While our experiments are extensive compared to previous works, in practical applications, there are many possible target domains that cannot be included in our experiments. However, as our method is target/adaptation aware, we believe our method can generalize better than existing SOTA methods which are target-agnostic.

\subsection{Broader Impact} 
Our work makes contribution to the generation of synthetic data in applications where sample collection is challenging, e.g., photos of rare animal species. This is an important contribution to many data-centric applications.
Furthermore, transfer learning of generative models using a few data samples enables data and computation-efficient model development and has positive impact on environmental sustainability and the reduction of greenhouse gas emissions. 
While our work targets generative applications with limited data, it parallelly raises concerns regarding such methods being used for malicious purposes.
Given the recent success of forensic detectors \cite{Chandrasegaran_2022_ECCV,Wang_2020_CVPR,Chandrasegaran_2021_CVPR,frank2020leveraging, zhao2023recipe}, we conduct a simple study using Color-Robust forensic detector proposed in \cite{Chandrasegaran_2022_ECCV} on our Babies and Cat datasets. 
We observe that the model achieves 99.8\% and 99.9\% average precision (AP) respectively showing that AdAM samples can be successfully detected.
We also remark that our work presents opportunities for improving knowledge transfer methods \cite{hinton_kd, chandrasegaran22_icml, heo2019comprehensive, evci2022head2toe, zhao2023fs} in a broader context.

\subsection{Potential Risks and Ethical Concerns}
\label{sec-supp:societal_impact}
Given very limited target domain samples (e.g., 10-shot), our proposed method for FSIG is lightweight and achieves state-of-the-art results with different source/target domain proximity.
Though our work shows exciting results by pushing the limits of FSIG, we urge researchers, practitioners, and developers to use our work with privacy, ethical and moral concerns. 
In what next, we bring an example to our discussion.

\textbf{Adapting our algorithm to a particular person.}
The idea of FSIG aims to adapt a pretrained GAN to a target domain with limited samples. Consequently, it is possible and reasonable that a user of FSIG can take few-shot images of a particular person and generate diverse images of the person, which leads to potential safety and privacy concerns. We conduct an experiment to adapt a pretrained StyleGAN-V2 generator to a single Obama photo \cite{zhao2020differentiable}. The visualization results and analysis are shown in Figure~\ref{fig:supp_obama}. 
On the other hand, our method can be adapted to generate images of the same person by applying a more restrictive selection of the source model’s knowledge. However, this would degrade the diversity of the outputs and may not be suitable for general FSIG which our paper focuses on. 


\textbf{Our model is lightweight for deployment.} Compared to some recent multi-modal text-to-image generative models, e.g., Stable Diffusion \cite{ramesh2022hierarchical} for few-shot adaptation tasks, e.g., DreamBooth \cite{ruiz2022dreambooth} and Textual Inversion \cite{gal2022image}, our proposed method for GAN-based few-shot image generation can be easily applied on edge devices, due to the lightweight model size (Stable Diffusion has $\sim$890M parameters and StyleGAN-V2 generator has around $\sim$30M parameters, which is $\sim$30 times less scale). Therefore, it is vital for users of our algorithms to bear in mind these ethical concerns.




{
\tocless\bibliographystyle{IEEEtran}
\tocless\bibliography{reference}
}

\clearpage


{

\onecolumn
\section*{\Large Supplementary Material}



\section{Additional Implementation Details of AdAM}
\label{sec_supp:ip}

\subsection{Computational overhead}
\label{sec_supp:ip_overhead}

Our proposed Importance Probing (IP) in AdAM to measure the importance of each individual kernel in the source GAN for the target-domain is lightweight. 
\textit{i.e.}: proposed IP only requires 8 minutes compared to the adaptation step which requires $\approx$ 110 minutes (Averaged over 3 runs for FFHQ $\rightarrow$ Cat adaptation experiment).
This is achieved through two design choices:

\begin{itemize}
    \item During IP, only modulation parameters are updated. Given that our modulation design is low-rank KML, the number of trainable parameters is significantly small compared to the actual source GAN. \textit{e.g.,}: for the generator, the number of trainable parameters in our proposed IP is only 0.1M whereas the entire generator has 30.1M trainable parameters.
    
    \item Our proposed IP is performed for limited number of iterations to measure the importance for the target domain. \textit{i.e.}: IP stage requires only 500 iterations to achieve a good performance for adaptation, while the full adaptation for FSIG may require up to 6000 iterations.
\end{itemize}

Complete details on number of trainable parameters and compute time for our proposed method and existing FSIG works are provided in Table \ref{tab:ip_computation_comparison}.
As one can observe, our proposed AdAM (IP + adaptation) is better than existing FSIG works in terms of trainable parameters and compute time.

\begin{table}[h]
    \centering
    \caption{
    Comparison of training cost in terms of number of trainable parameters, iterations and compute time for different FSIG methods. 
    FFHQ is the source domain and we show results for Babies (top) and Cat (bottom) target domains.
    One can clearly observe that our proposed IP is extremely lightweight and our KML based adaptation contains much less trainable parameters in the source GAN. 
    All results are measured in containerized environments using a single Tesla V100-SXM2 (32 GB) GPU with batch size of 4. All reported results are averaged over 3 independent runs.
    }
        \begin{tabular}{l|cccc}
        \multicolumn{5}{c}{\textbf{FFHQ} $\rightarrow$ \textbf{Babies}}\\
         \toprule
         {Method} & {Stage} & { \# trainable params (M)} &  {\# iteration} & {\# time} \\\hline
         TGAN \cite{wang2018transferringGAN} & Adaptation & 30.1 & 3000 & 110 mins \\ 
         FreezeD \cite{mo2020freezeD} & Adaptation & 30.1 & 3000 & 110 mins \\ 
         EWC \cite{li2020fig_EWC} & Adaptation & 30.1 & 3000 & 110 mins \\ 
         CDC \cite{ojha2021fig_cdc} & Adaptation & 30.1 & 3000 & 120 mins\\ 
         DCL \cite{zhao2022dcl} & Adaptation & 30.1 & 3000 & 120 mins \\
         \cline{1-5}
         \multirow{2}{*}{\textbf{AdAM (Ours)}} & IP & \textbf{0.105} & \textbf{500} & 8 mins\\ 
         \cline{2-5}
         & Adaptation & \textbf{11.9} & \textbf{1500}
         & 45min \\
         \bottomrule
        \end{tabular}
        
        \vspace{1em}
        \begin{tabular}{l|cccc}
        \multicolumn{5}{c}{\textbf{FFHQ} $\rightarrow$ \textbf{AFHQ-Cat}}\\
         \toprule
         {Method} & {Stage} & { \# trainable params (M)} &  {\# iteration} & {\# time} \\\hline
         TGAN \cite{wang2018transferringGAN} & Adaptation & 30.1 & 6000 & 210 mins \\ 
         FreezeD \cite{mo2020freezeD} & Adaptation & 30.1 & 6000 & 200 mins \\ 
         EWC \cite{li2020fig_EWC} & Adaptation & 30.1 & 6000 & 220 mins \\ 
         CDC \cite{ojha2021fig_cdc} & Adaptation & 30.1 & 6000 & 300 mins\\ 
         DCL \cite{zhao2022dcl} & Adaptation & 30.1 & 6000 & 300 mins \\
         \cline{1-5}
         \multirow{2}{*}{\textbf{AdAM (Ours)}} & IP & \textbf{0.105} & \textbf{500} & 8 mins\\ 
         \cline{2-5}
         & Adaptation & \textbf{17.85} & \textbf{2500}
         & 105 mins \\
         \bottomrule
        \end{tabular}
    \label{tab:ip_computation_comparison}
\end{table}

\subsection{Fisher information approximation using proxy vectors}
\label{sec_supp:ip_fim_approximation}

Recall in Sec. {\color{red} 4} of main paper, we consider low-rank approximation of modulation matrix using outer product of proxy vectors:  $\mathbf{M}_i = reshape([\mathbf{m}_1^{i}\mathbf{m}_2^1, \dots, \mathbf{m}_1^i\mathbf{m}_2^{c_{in} \times k \times k}])$, where $|\mathbf{m}_2|={(c_{in} \times k \times k)}$.
In order to calculate the FI of the modulation matrix, we start with the FI of each element in this matrix. Considering $m_{ij}=\mathbf{m}_1^i\mathbf{m}_2^j$, following equation can be derived by simple application of 
chain rule of 
differentiation:
\begin{equation}
    \frac{\partial\mathcal{L}}{\partial m_{ij}} = \frac{1}{2\mathbf{m}_2^j}\frac{\partial\mathcal{L}}{\partial \mathbf{m}_1^i} 
    + \frac{1}{2\mathbf{m}_1^i}\frac{\partial\mathcal{L}}{\partial \mathbf{m}_2^j}
\end{equation}
We use the square of the gradients to estimate the FI ~\cite{achille2019task2vec}. Therefore, the following equation can be obtained between the FI of these variables:
\begin{equation}
    \mathcal{F}(m_{ij}) = \frac{1}{{4\mathbf{m}_2^j}^2}\mathcal{F}(\mathbf{m}_1^i)
    + \frac{1}{{4\mathbf{m}_1^i}^2}\mathcal{F}(\mathbf{m}_2^j)
    + \frac{1}{2\mathbf{m}_1^i\mathbf{m}_2^j}\frac{\partial\mathcal{L}}{\partial \mathbf{m}_1^i}\frac{\partial\mathcal{L}}{\partial \mathbf{m}_2^j}
\end{equation}
Then, the FI of the modulation matrix $\mathbf{M}_i = [m_{i1}, m_{i2}, \dots] $, can be calculated as:
\begin{equation}
    \begin{split}
    \label{eqn:FI_full}
    \mathcal{F}(\mathbf{M}_i) & = \mathlarger{\mathlarger{\sum}}_{j=1}^{|\mathbf{m}_2|}\mathcal{F}(m_{ij}) \\ 
    & = \mathlarger{\mathlarger{\sum}}_{j=1}^{|\mathbf{m}_2|} (\frac{1}{{4\mathbf{m}_2^j}^2}\mathcal{F}(\mathbf{m}_1^i)
    + \frac{1}{{4\mathbf{m}_1^i}^2}\mathcal{F}(\mathbf{m}_2^j) + \frac{1}{2\mathbf{m}_1^i\mathbf{m}_2^j}\frac{\partial\mathcal{L}}{\partial \mathbf{m}_1^i}\frac{\partial\mathcal{L}}{\partial \mathbf{m}_2^j}) \\
    & = \mathcal{F}(\mathbf{m}_1^i) \mathlarger{\mathlarger{\sum}}_{j=1}^{|\mathbf{m}_2|}\frac{1}{{4\mathbf{m}_2^j}^2}
    + \frac{1}{{4\mathbf{m}_1^i}^2} \mathlarger{\mathlarger{\sum}}_{j=1}^{|\mathbf{m}_2|} \mathcal{F}(\mathbf{m}_2^j) \\ & + \frac{1}{2\mathbf{m}_1^i}\frac{\partial\mathcal{L}}{\partial \mathbf{m}_1^i} \mathlarger{\mathlarger{\sum}}_{j=1}^{|\mathbf{m}_2|} \frac{1}{\mathbf{m}_2^j} \frac{\partial\mathcal{L}}{\partial \mathbf{m}_2^j}
    \end{split}
\end{equation}
We empirically 
observed
that discarding (i) the cross-term (ii) the coefficients 
($\frac{1}{{4\mathbf{m}_2^j}^2}$,
$\frac{1}{{4\mathbf{m}_1^i}^2}$) 
in the importance of each kernel in Eqn.~\ref{eqn:FI_full}
results in a similar FID for the final adapted model. 
Therefore, 
the estimation can be simpler and more lightweight.
In particular, the following  (simpler) estimated version of $\mathcal{F}(\mathbf{M}_i)$ is used in our experiments:  
\begin{equation}
\label{eq:FI_vectors_supp}
   \hat{\mathcal{F}}(\mathbf{M}_i) =
    \mathcal{F}(\mathbf{m}_1^i) + \frac{1}{|\mathbf{m}_2|} \sum_{j=1}^{|\mathbf{m}_2|}\mathcal{F}(\mathbf{m}_2^j)
\end{equation}
Note that $\hat{\mathcal{F}}(\mathbf{M}_i)$ intuitively estimates the FI of the modulation matrix by a weighted average of its constructing parameters corresponding to their occurrence frequency
in calculation of 
$\mathbf{M}_i$.
We remark that in our implementation, for reporting all of the results in the main paper, and also the additional results in this Supplement, we have used this lightweight estimation Eqn.~\ref{eq:FI_vectors_supp} to calculate the importance of each kernel during importance probing.

\section{Additional Experiment Results}
\label{sec-supp:additional_main_paper_results}

\subsection{Additional source / target domain adaptation}
\label{subsec-supp:additional_source_target_domains_supp}

\textbf{Proximity analysis.}
Following \cite{ojha2021fig_cdc}, we conduct extended experiments using Church as the source domain. 
\cite{ojha2021fig_cdc} uses Haunted houses and Van Gogh Houses as target domains.
Similar to Sec. {\color{red}3} in the main paper, our analysis confirms that these target domains are closer to the source domain (Church). 
We additionally include palace and yurt as target domains to relax the close proximity assumption.
The proximity visualization is shown in Figure \ref{fig-supp:proximity-visualization}.

\begin{figure}

    \includegraphics[width=0.99\linewidth]{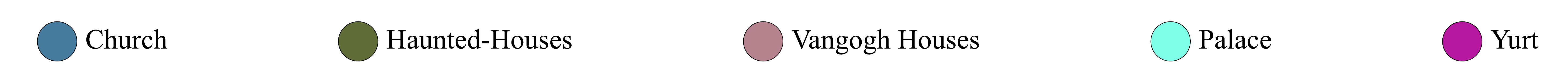} \\
    \vspace{-3 mm}
    \begin{tabular}{cc}
    
    \begin{minipage}{0.45\columnwidth}
     \includegraphics[width=1.0\linewidth]{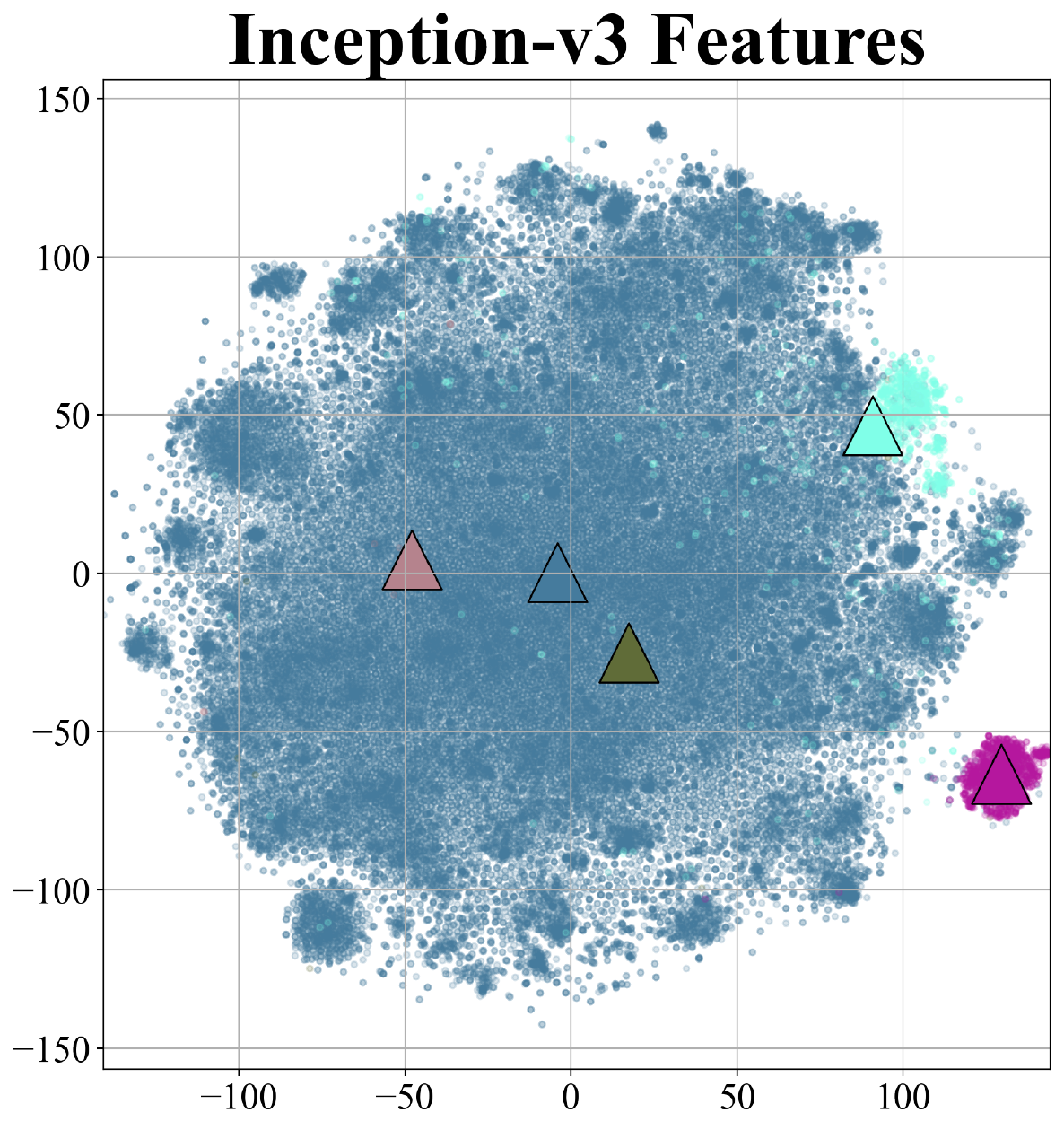}
     \end{minipage}
     
     & 

    \begin{minipage}{0.45\columnwidth}
     \includegraphics[width=1.0\linewidth]{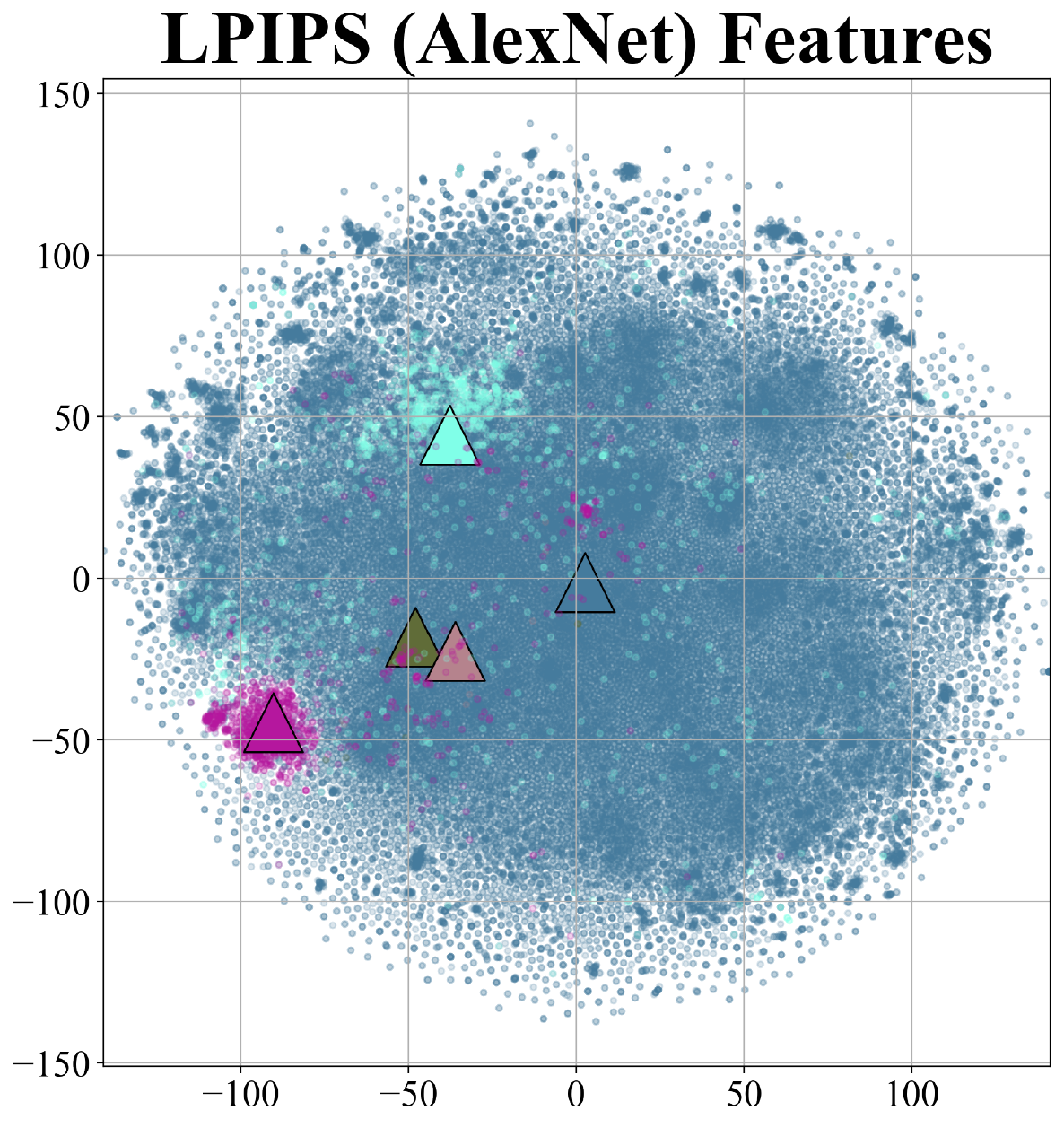}
     \end{minipage}
     
     \\
    \end{tabular}
  \caption{
\textit{Source-target domain proximity Visualization:}
We use Church as the source domain following \cite{ojha2021fig_cdc}.
We show 
source-target
domain proximity by
visualizing Inception-v3 (Left) \cite{szegedy2016rethinking} and LPIPS (Middle) \cite{zhang2018lpips} --using AlexNet \cite{krizhevsky2012alexnet} backbone-- features, 
and quantitatively using FID / LPIPS metrics (Right). 
For feature visualization, we use t-SNE \cite{JMLR:v9:vandermaaten08a_tsne} and show centroids ($\bigtriangleup$) for all domains. 
FID / LPIPS is measured with respect to FFHQ. 
There are 2 important observations: 
\textcircled{\raisebox{-0.9pt}{1}} Common target domains used in existing FSIG works (Haunted Houses, Van Gogh Houses) are notably proximal to the source domain (Church). This can be observed from the feature visualization and verified by FID / LPIPS measurements.
\textcircled{\raisebox{-0.9pt}{2}} 
We clearly show using feature visualizations and FID / LPIPS measurements that additional setups -- Palace \cite{deng2009imagenet} and Yurt \cite{deng2009imagenet} -- represent target domains that are distant from the source domain (Church).
We remark that due to availability of only 10-shot samples in the target domain, FID / LPIPS are not measured in these setups.
}
\label{fig-supp:proximity-visualization}
\end{figure}

\textbf{Adaptation results.}
\label{sup-sec:extended_experiments_10_shot_results}
Besides the results in the main paper, we show complete 10-shot adaptation results for our proposed AdAM for additional source / target domains:
Church $\rightarrow$ Haunted House (Figure \ref{fig:supp_haunted}),
Church $\rightarrow$ Van Gogh's House (Figure \ref{fig:supp_van}),
Church $\rightarrow$ Palace (Figure \ref{fig:supp_palace}),
Church $\rightarrow$ Yurt (Figure \ref{fig:supp_yurt}),
FFHQ $\rightarrow$ AFHQ-Dog (Figure \ref{fig:supp_dog}), 
FFHQ $\rightarrow$ AFHQ-Wild (Figure \ref{fig:supp_wild}), 
FFHQ $\rightarrow$ Sunglasses (Figure \ref{fig:supp_sunglasses}), 
FFHQ $\rightarrow$ MetFaces \cite{karras2020ADA} (Figure \ref{fig:supp_metface}), 
FFHQ $\rightarrow$ Sketches (Figure \ref{fig:supp_sketches}), 
FFHQ $\rightarrow$ Amedeo Modigliani’s Paintings (Figure \ref{fig:supp_amedeo}),
FFHQ $\rightarrow$ Otto Dix’s Paintings (\ref{fig:supp_otto}) and 
FFHQ $\rightarrow$ Raphael’s Paintings (Figure \ref{fig:supp_raphael}),
Cars $\rightarrow$ Wrecked Cars (Figure \ref{fig:supp_cars}). 
We also include the comparison to SOTA methods with distant target domains, see FFHQ $\rightarrow$ AFHQ-Dog in Figure \ref{fig:compare_dog}, FFHQ $\rightarrow$ AFHQ-Wild in Figure \ref{fig:compare_wild} and Church $\rightarrow$ Palace in Figure \ref{fig:supp_failure_palace}.
As one can observe, SOTA FSIG methods \cite{li2020fig_EWC, ojha2021fig_cdc, zhao2022dcl} are unable to adapt well to distant target domains due to \textit{only considering source domain / task in knowledge preservation.}
We remark that TGAN \cite{wang2018transferringGAN} suffers severe mode collapse.
We clearly show that AdAM (ours) outperforms SOTA FSIG methods \cite{li2020fig_EWC, ojha2021fig_cdc, zhao2022dcl} and produces high quality images with good diversity.

\begin{figure}[h]
    \centering
    \includegraphics[width=\textwidth]{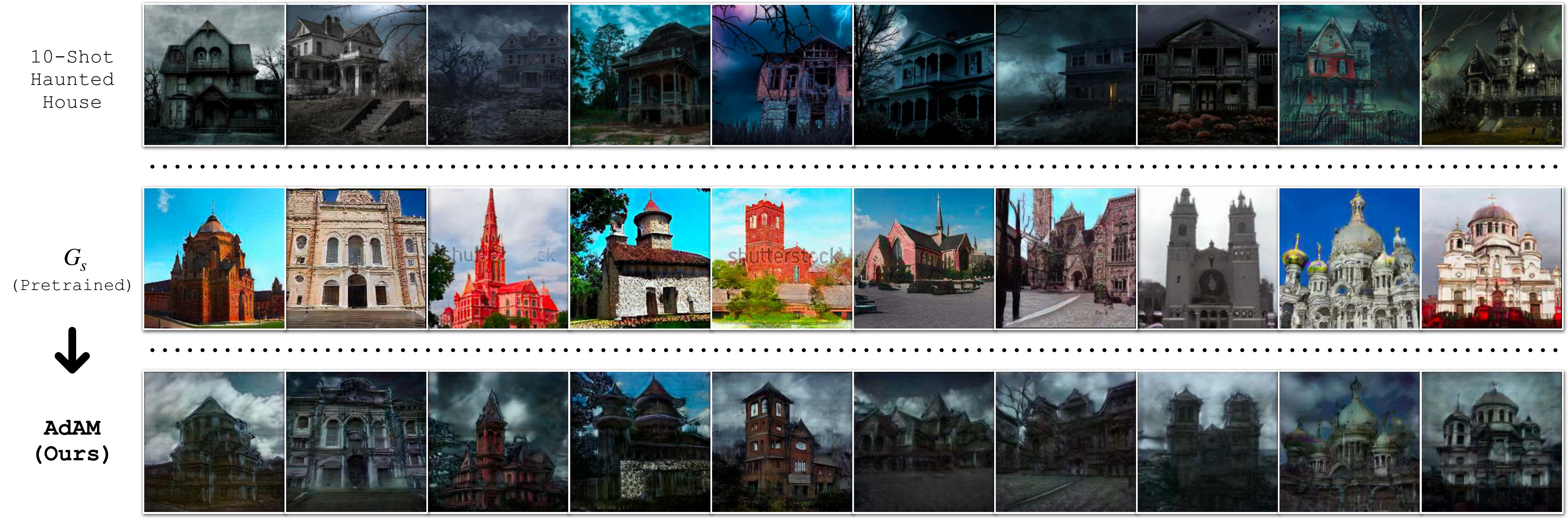}
    \caption{Church $\rightarrow$ Haunted House}
    \label{fig:supp_haunted}
\end{figure}

\begin{figure}[h]
    \centering
    \includegraphics[width=\textwidth]{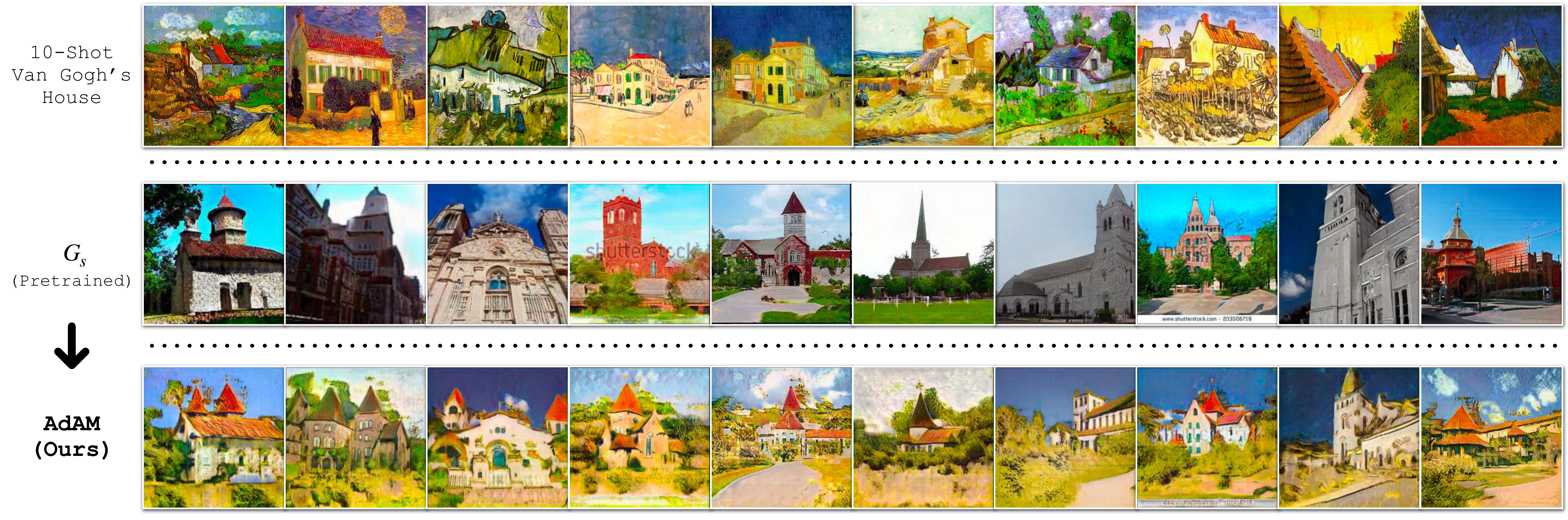}
    \caption{Church $\rightarrow$ Van Gogh's House}
    \label{fig:supp_van}
\end{figure}

\begin{figure}[h]
    \centering
    \includegraphics[width=\textwidth]{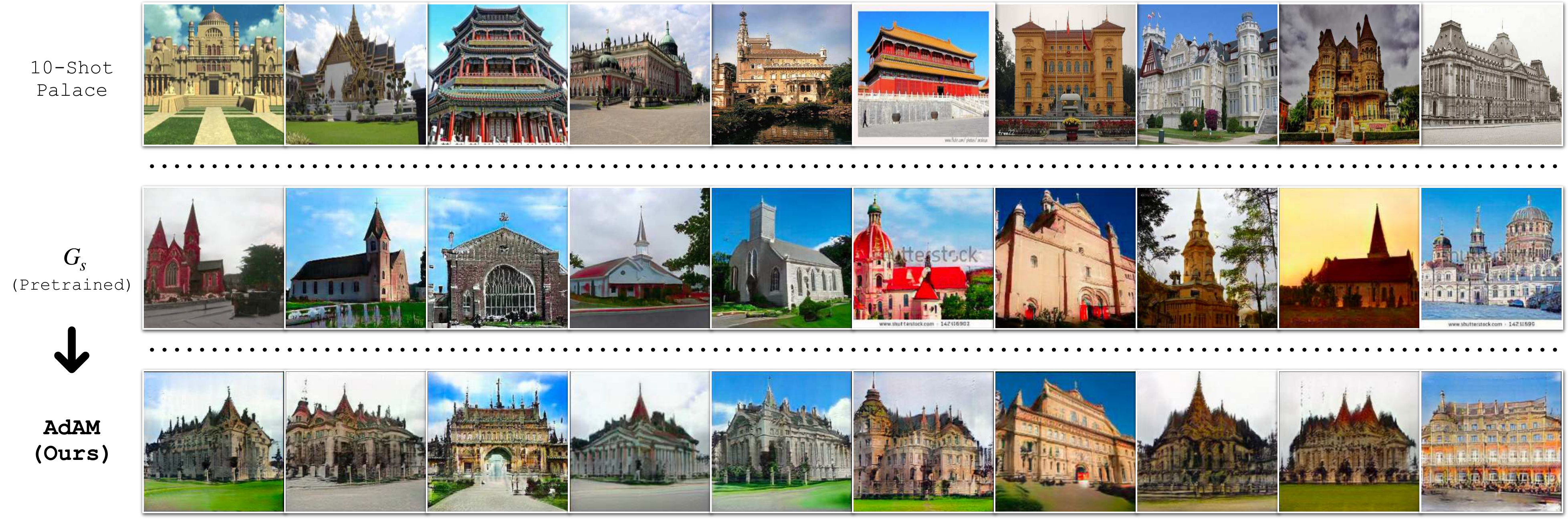}
    \caption{Church $\rightarrow$ Palace (distant domain)}
    \label{fig:supp_palace}
\end{figure}

\begin{figure}[h]
    \centering
    \includegraphics[width=\textwidth]{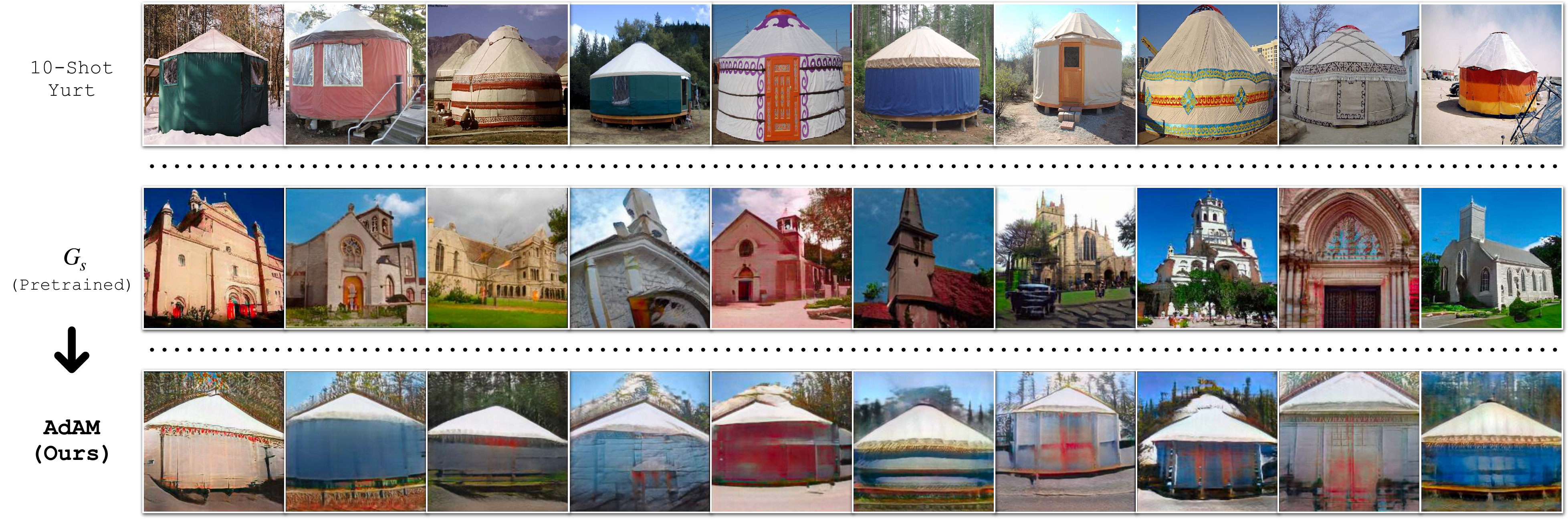}
    \caption{Church $\rightarrow$ Yurt (distant domain)}
    \label{fig:supp_yurt}
\end{figure}

\begin{figure}[h]
    \centering
    \includegraphics[width=\textwidth]{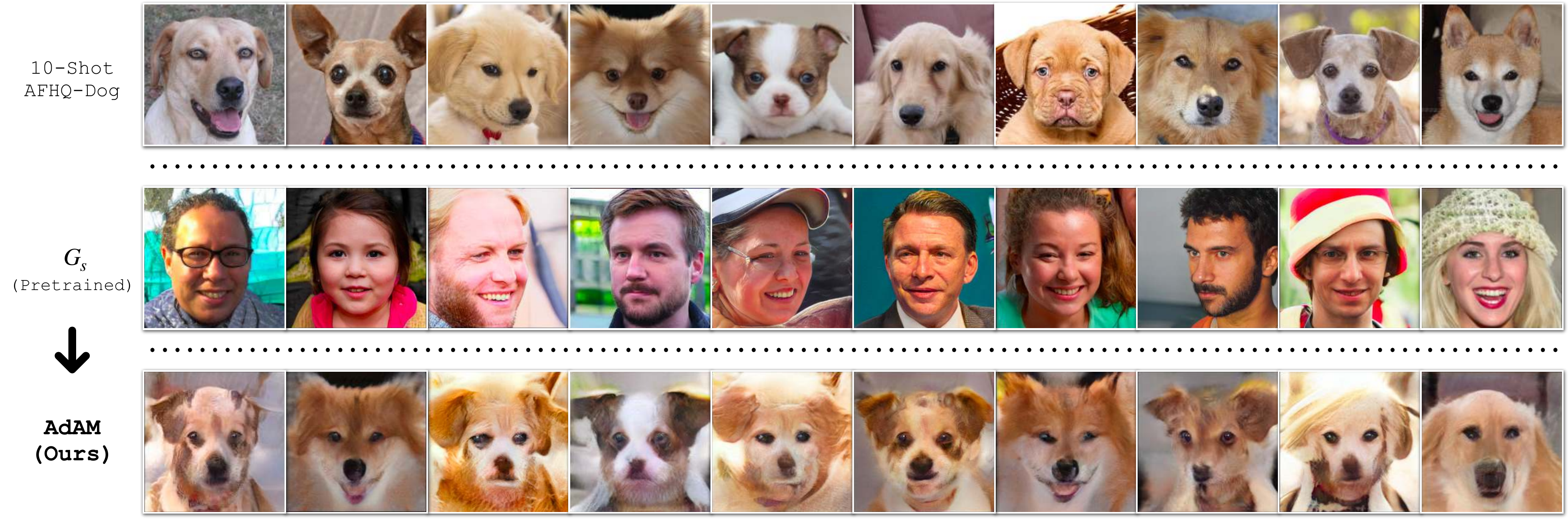}
    \caption{FFHQ $\rightarrow$ AFHQ-Dog (distant domain)}
    \label{fig:supp_dog}
\end{figure}

\begin{figure}[h]
    \centering
    \includegraphics[width=\textwidth]{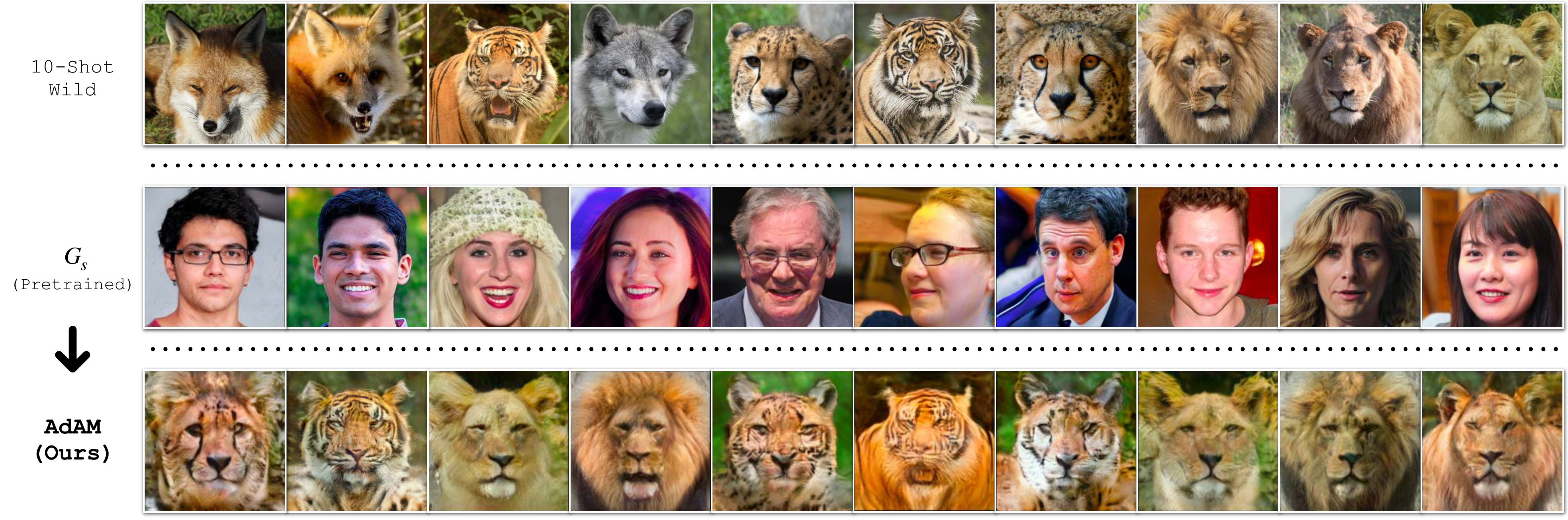}
    \caption{FFHQ $\rightarrow$ AFHQ-Wild (distant domain)}
    \label{fig:supp_wild}
\end{figure}

\begin{figure}[h]
    \centering
    \includegraphics[width=\textwidth]{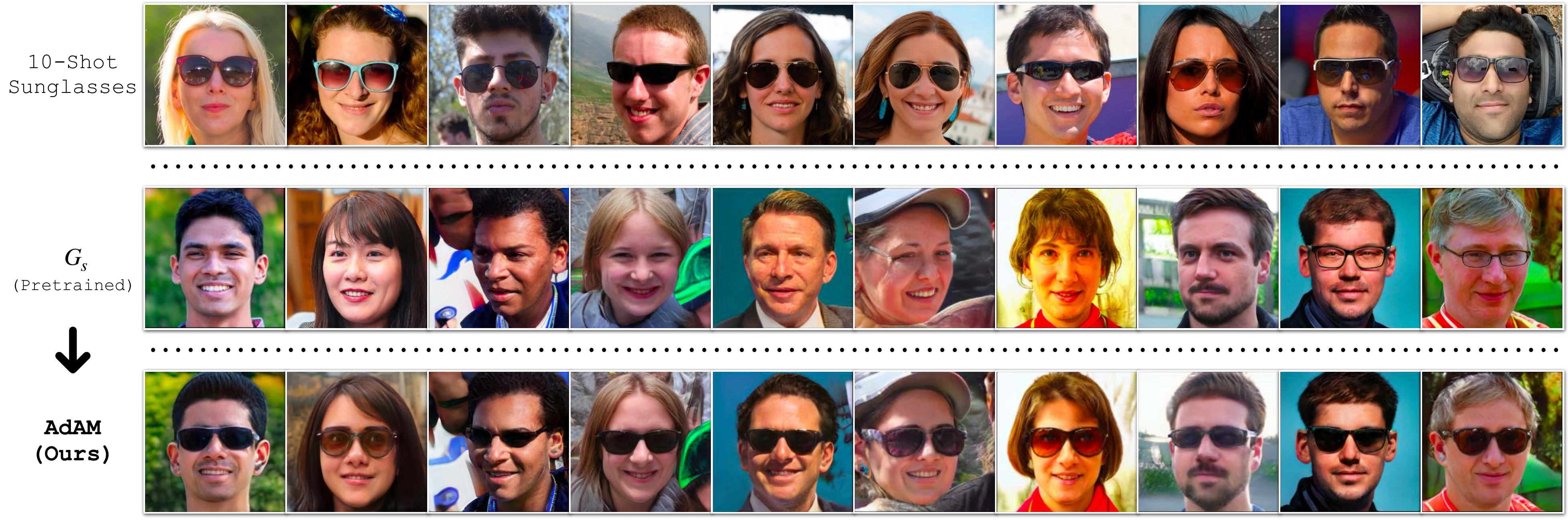}
    \caption{FFHQ $\rightarrow$ Sunglasses}
    \label{fig:supp_sunglasses}
\end{figure}

\begin{figure}[h]
    \centering
    \includegraphics[width=\textwidth]{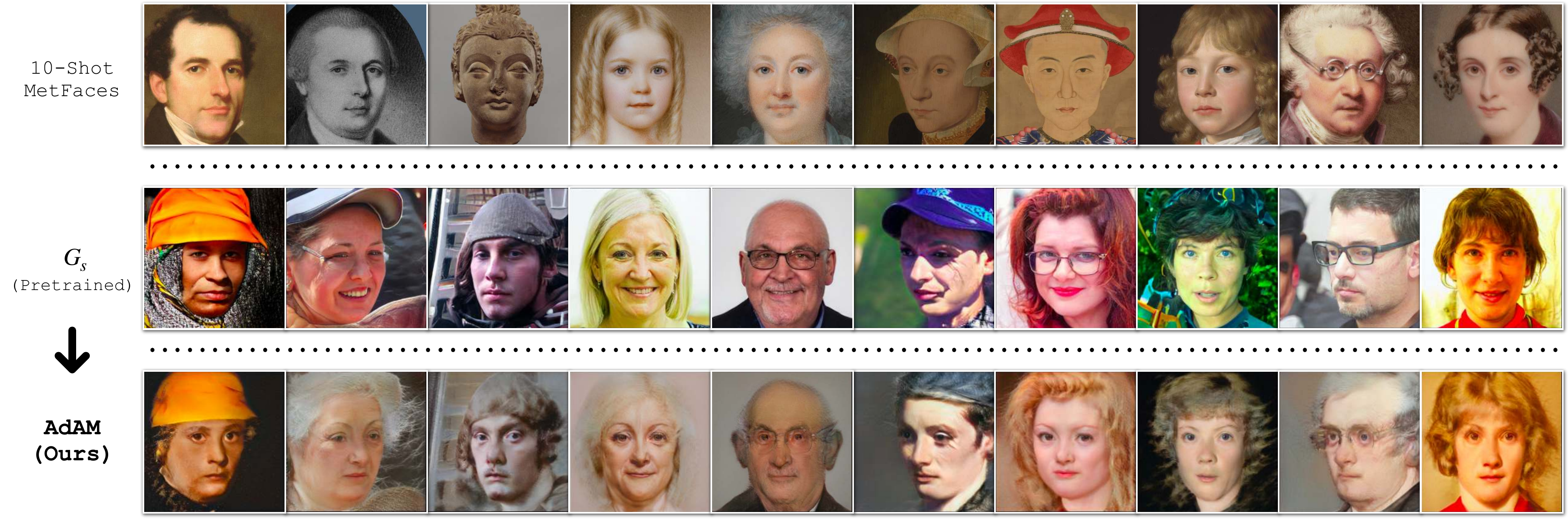}
    \caption{FFHQ $\rightarrow$ MetFaces}
    \label{fig:supp_metface}
\end{figure}

\begin{figure}[h]
    \centering
    \includegraphics[width=\textwidth]{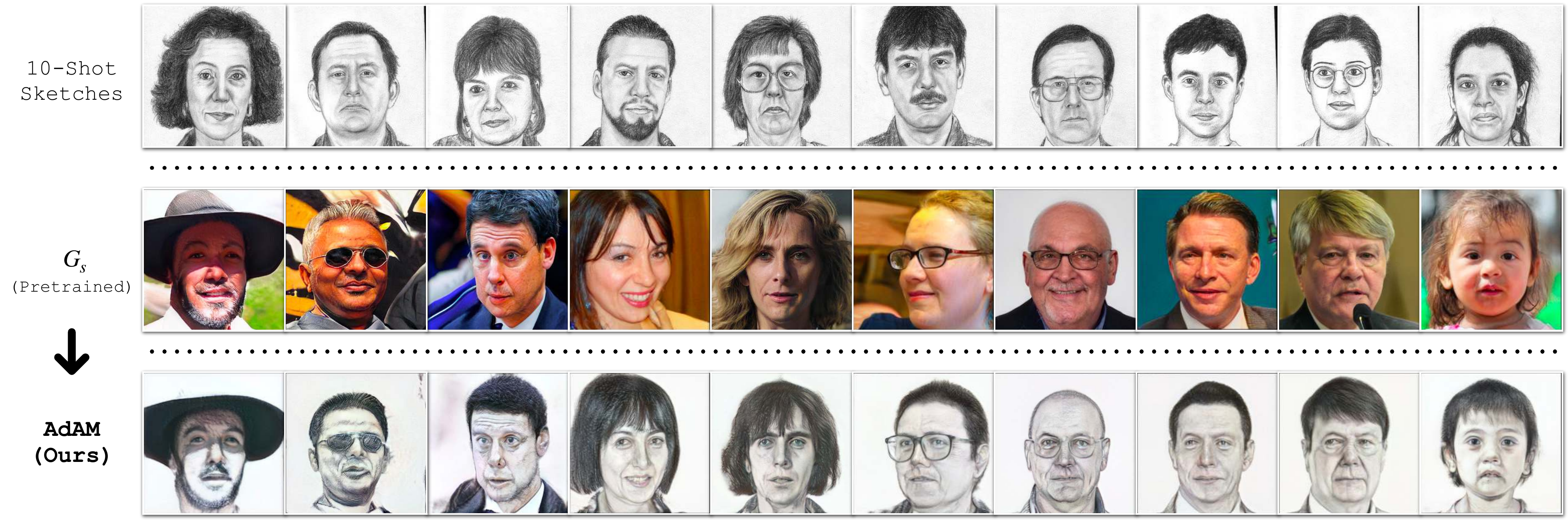}
    \caption{FFHQ $\rightarrow$ Sketches}
    \label{fig:supp_sketches}
\end{figure}

\begin{figure}[h]
    \centering
    \includegraphics[width=\textwidth]{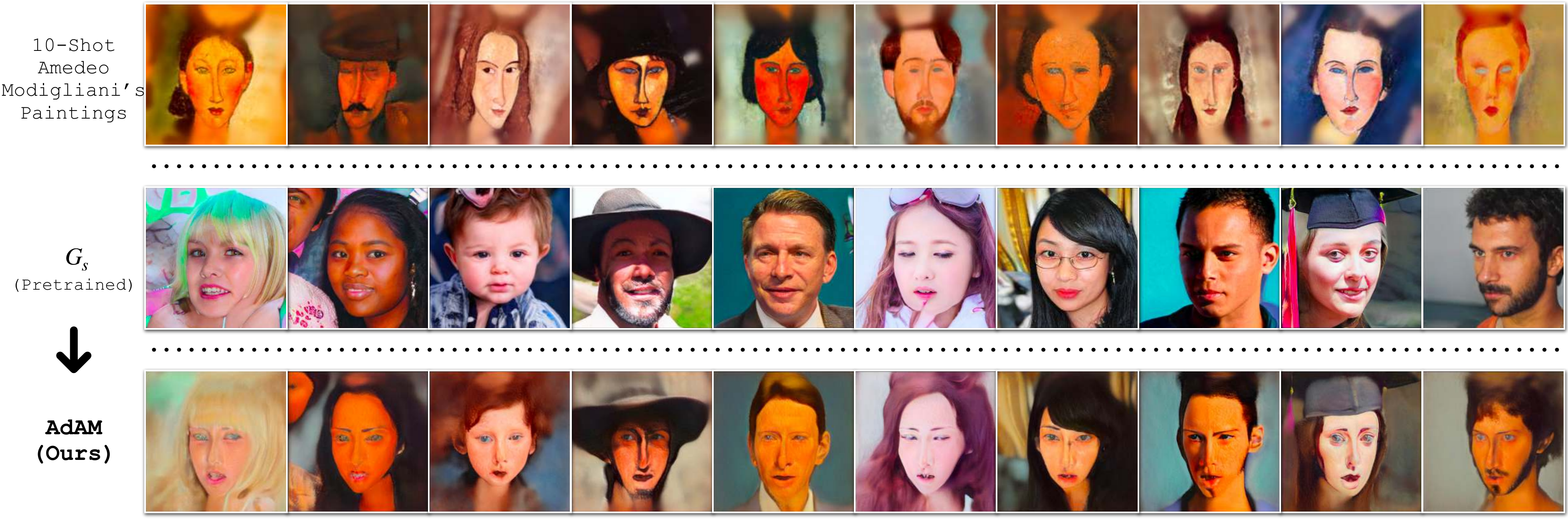}
    \caption{FFHQ $\rightarrow$ Amedeo Modigliani’s Paintings}
    \label{fig:supp_amedeo}
\end{figure}

\begin{figure}[h]
    \centering
    \includegraphics[width=\textwidth]{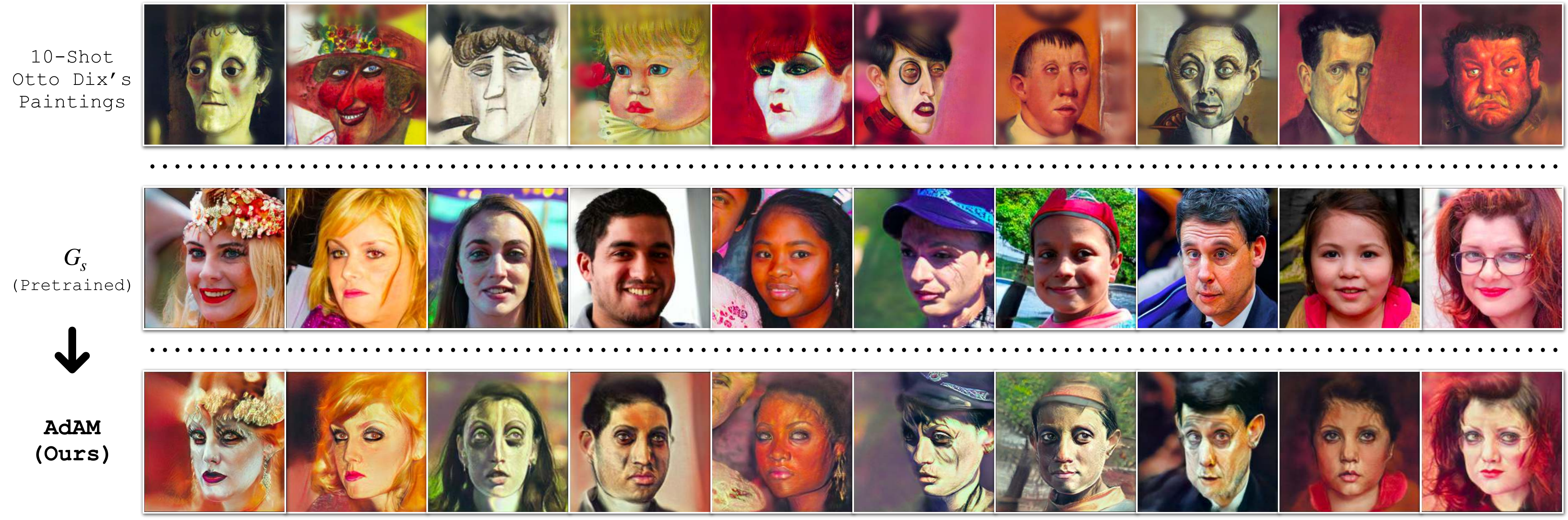}
    \caption{FFHQ $\rightarrow$ Otto Dix’s Paintings}
    \label{fig:supp_otto}
\end{figure}

\begin{figure}[h]
    \centering
    \includegraphics[width=\textwidth]{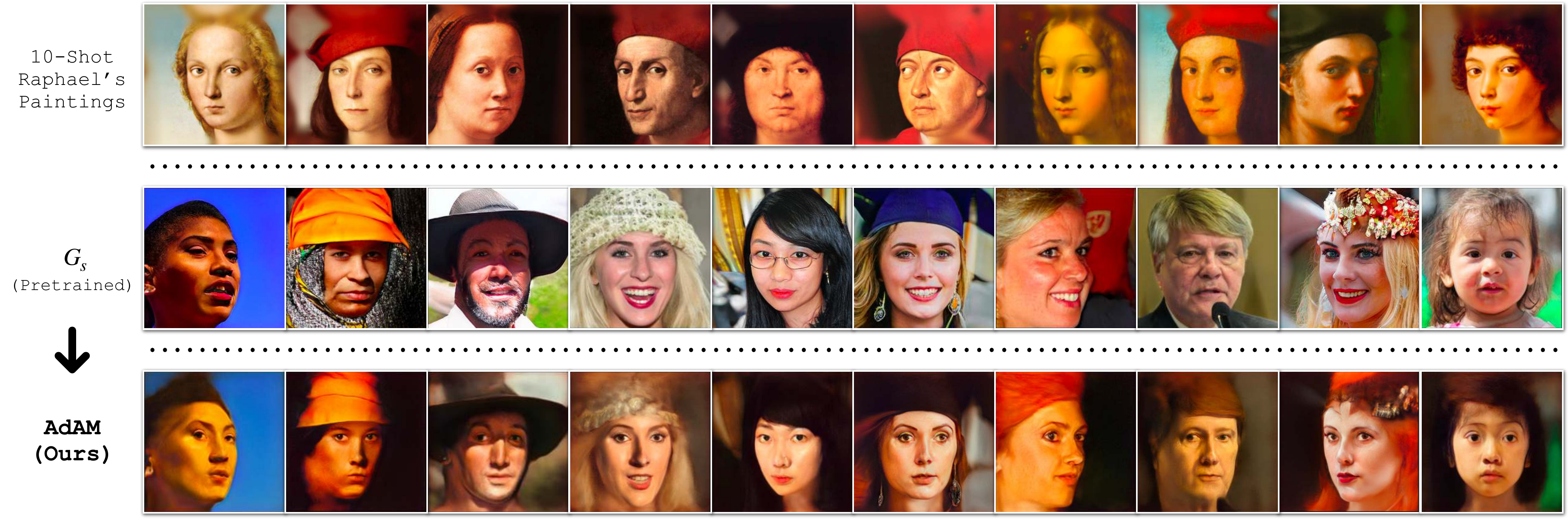}
    \caption{FFHQ $\rightarrow$ Raphael’s Paintings}
    \label{fig:supp_raphael}
\end{figure}

\begin{figure}[h]
    \centering
    \includegraphics[width=\textwidth]{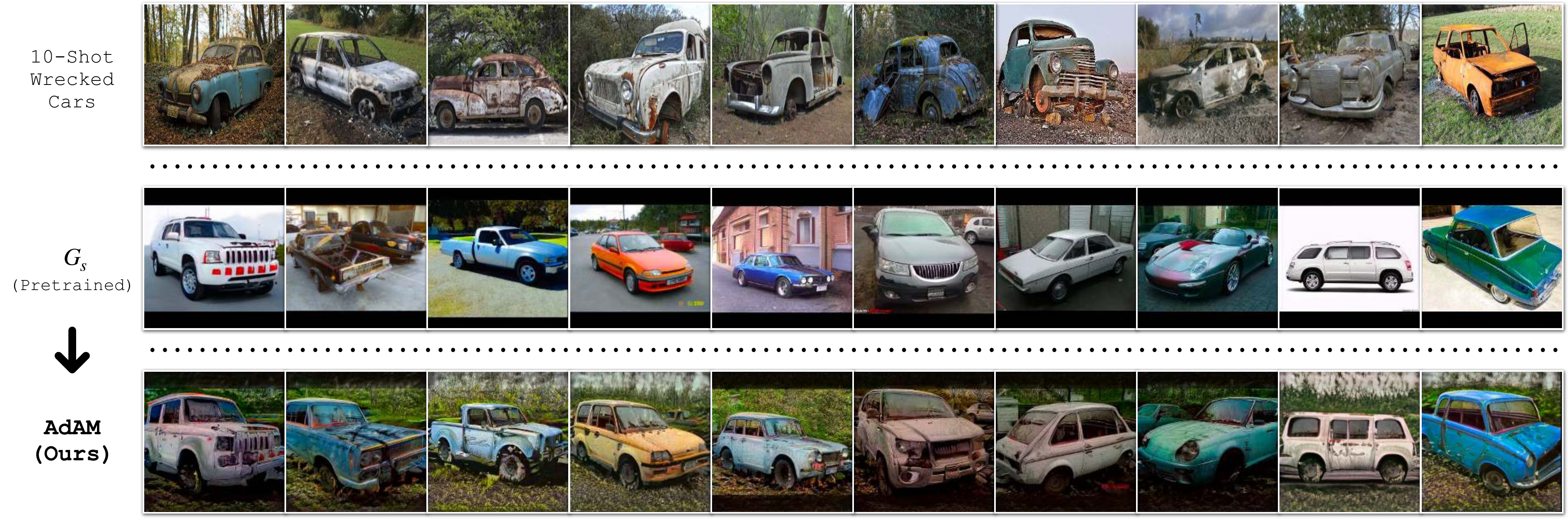}
    \caption{Cars $\rightarrow$ Wrecked Cars}
    \label{fig:supp_cars}
\end{figure}

\begin{figure}[h]
    \centering
    \includegraphics[width=\textwidth]{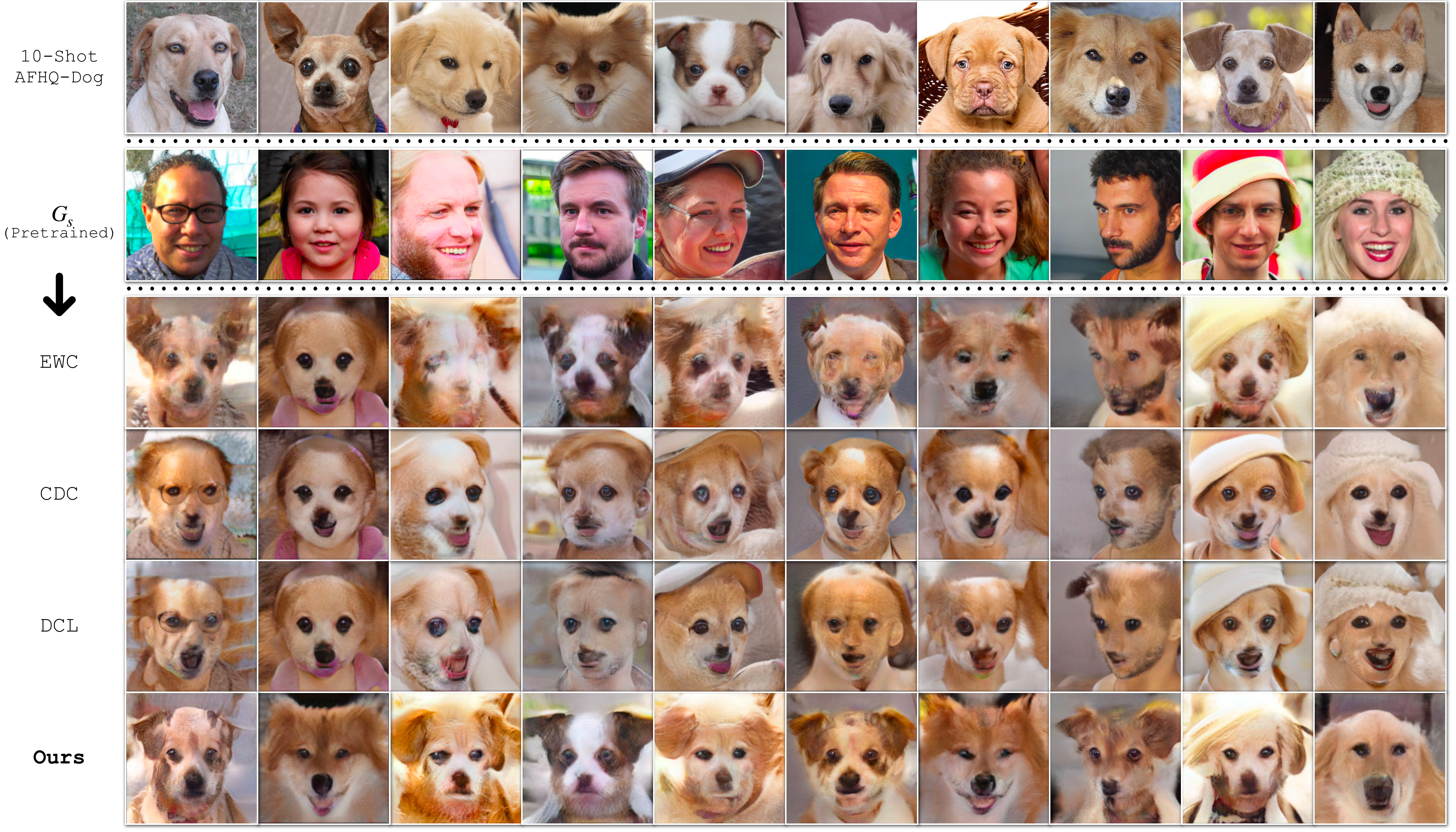}
    \caption{
    Comparison to SOTA methods on
    FFHQ $\rightarrow$ AFHQ-Dog.
    }
    \label{fig:compare_dog}
\end{figure}

\begin{figure}[h]
    \centering
    \includegraphics[width=\textwidth]{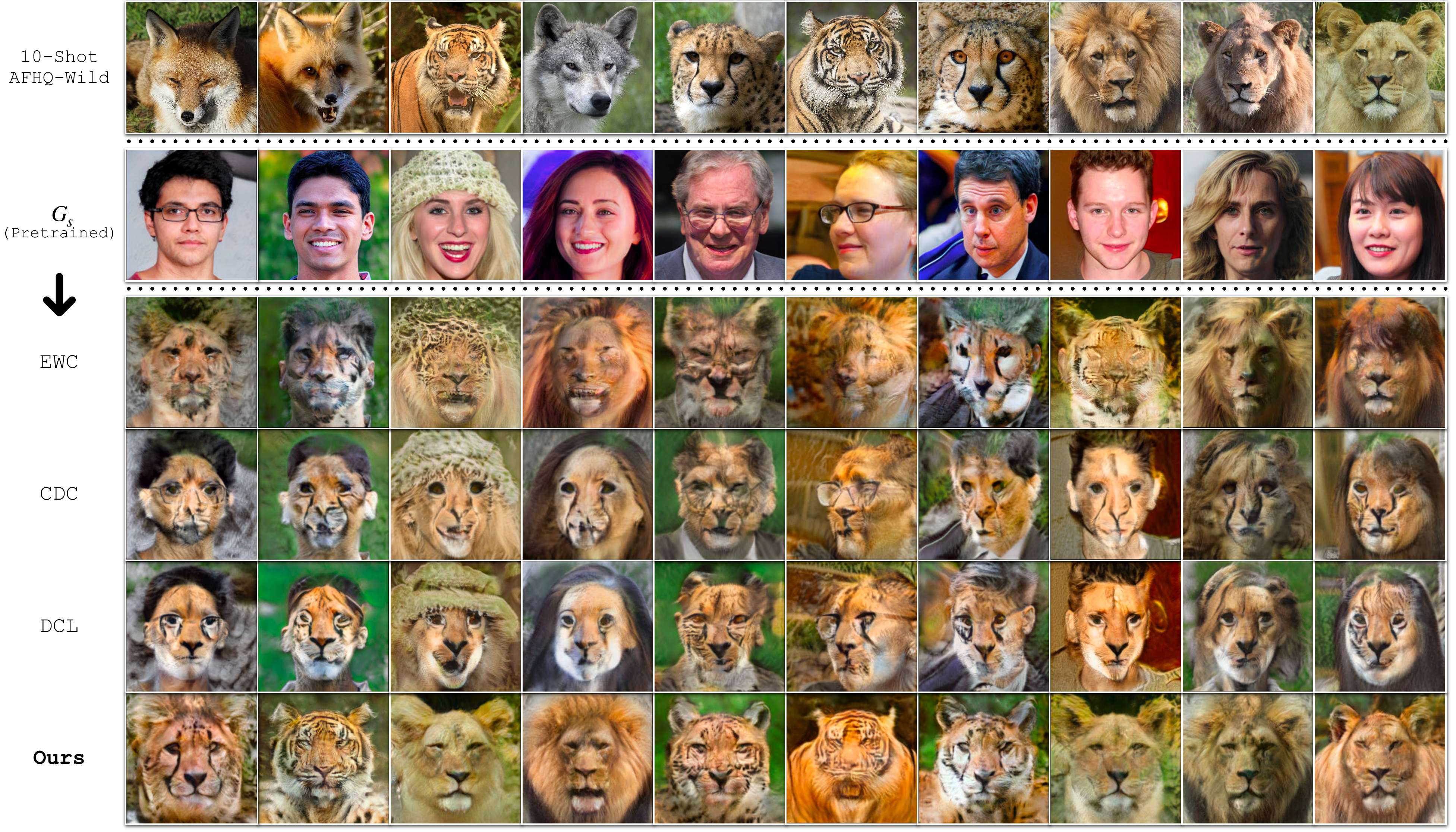}
    \caption{
    Comparison to SOTA methods on
    FFHQ $\rightarrow$ AFHQ-Wild.
    }
    \label{fig:compare_wild}
\end{figure}

\begin{figure}[h]
    \centering
    \includegraphics[width=0.6\textwidth]{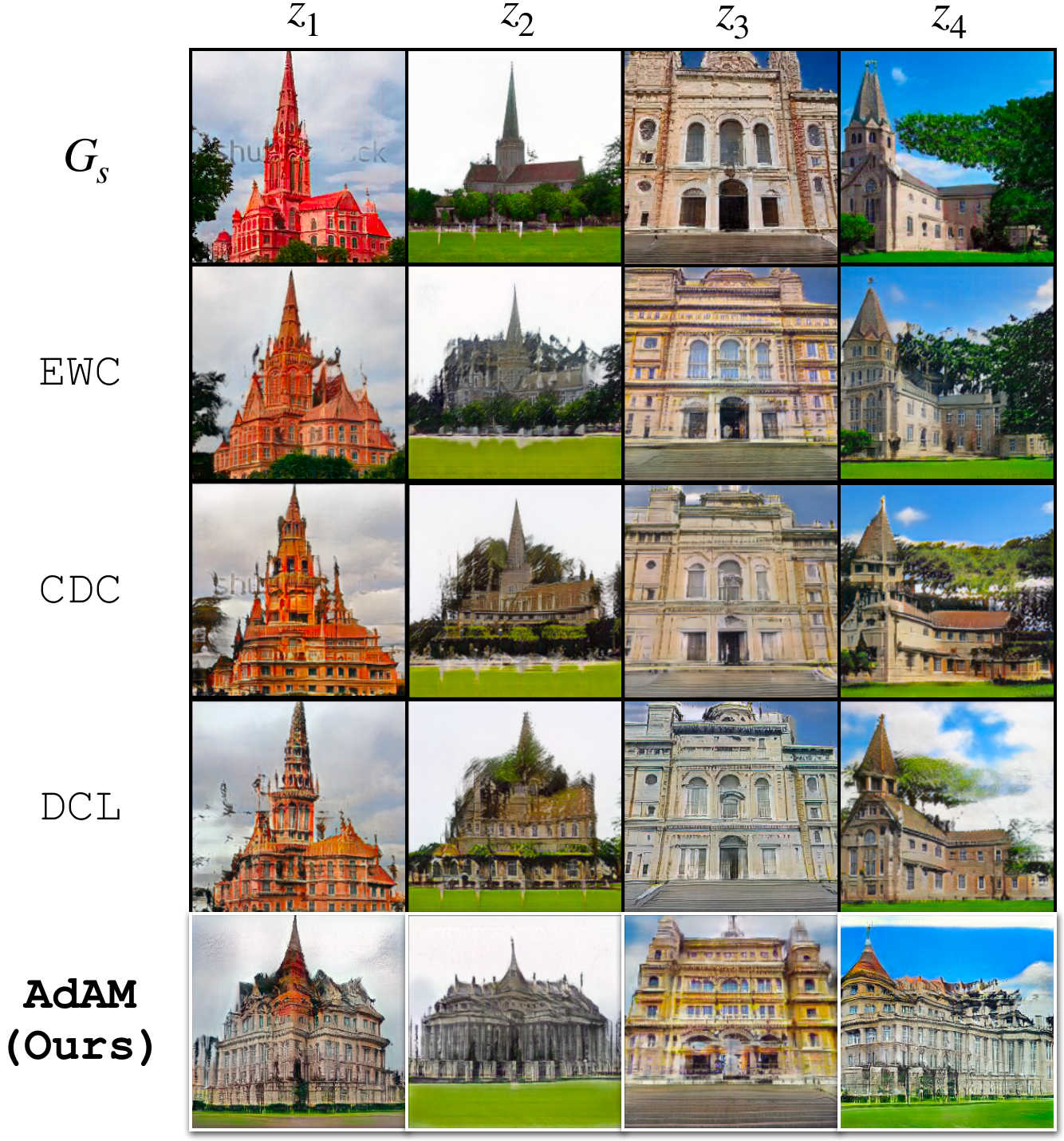}
    \caption{Church $\rightarrow$ Palace (distant domain)}
    \label{fig:supp_failure_palace}
\end{figure}

\subsection{Visualization of adaptation results with more shots}
In addition to the analysis of increasing the number of shots for target adaptation in Figure {\color{red} 6} and Table {\color{red} 5} in the main paper, here we additionally show the generated images with 100-shot training data, on Babies and AFHQ-Cat as the target domains. FFHQ is the source domain. 
The results are in Figure \ref{fig:100-shot} where each column represents a fixed noise input. 
Compared to baseline and SOTA methods, our generated images can still produce the best quality and  diversity.

\subsection{Additional GAN architectures}
\label{subsec-supp:additional_gan_architectures_supp}

We use an additional pre-trained GAN architecture, ProGAN \cite{karras2017progan}, to conduct FSIG experiments for FFHQ $\rightarrow$ Babies, FFHQ $\rightarrow$ Cat, Church $\rightarrow$ Haunted houses and Church $\rightarrow$ Palace setups. For fair comparison, we strictly follow the exact experiment setup discussed in Section \ref{subsec-supp:additional_source_target_domains_supp}. 

\textbf{Results.} The qualitative and quantitative results are in Figures \ref{fig:rebuttal_progan_haunted_houses} and \ref{fig:rebuttal_progan_palace}. 
We remark that the results for FFHQ $\rightarrow$ Babies, FFHQ $\rightarrow$ Cat are included in the main paper, see Sec. {\color{red} 6.7}.
As one can observe, our proposed method consistently outperforms other baseline and SOTA FSIG methods with another pre-trained GAN model (ProGAN \cite{karras2017progan}), demonstrating the effectiveness and generalizability of our method over various GAN architectures.

\begin{figure}[]
    \centering
    \includegraphics[width=0.8\textwidth]{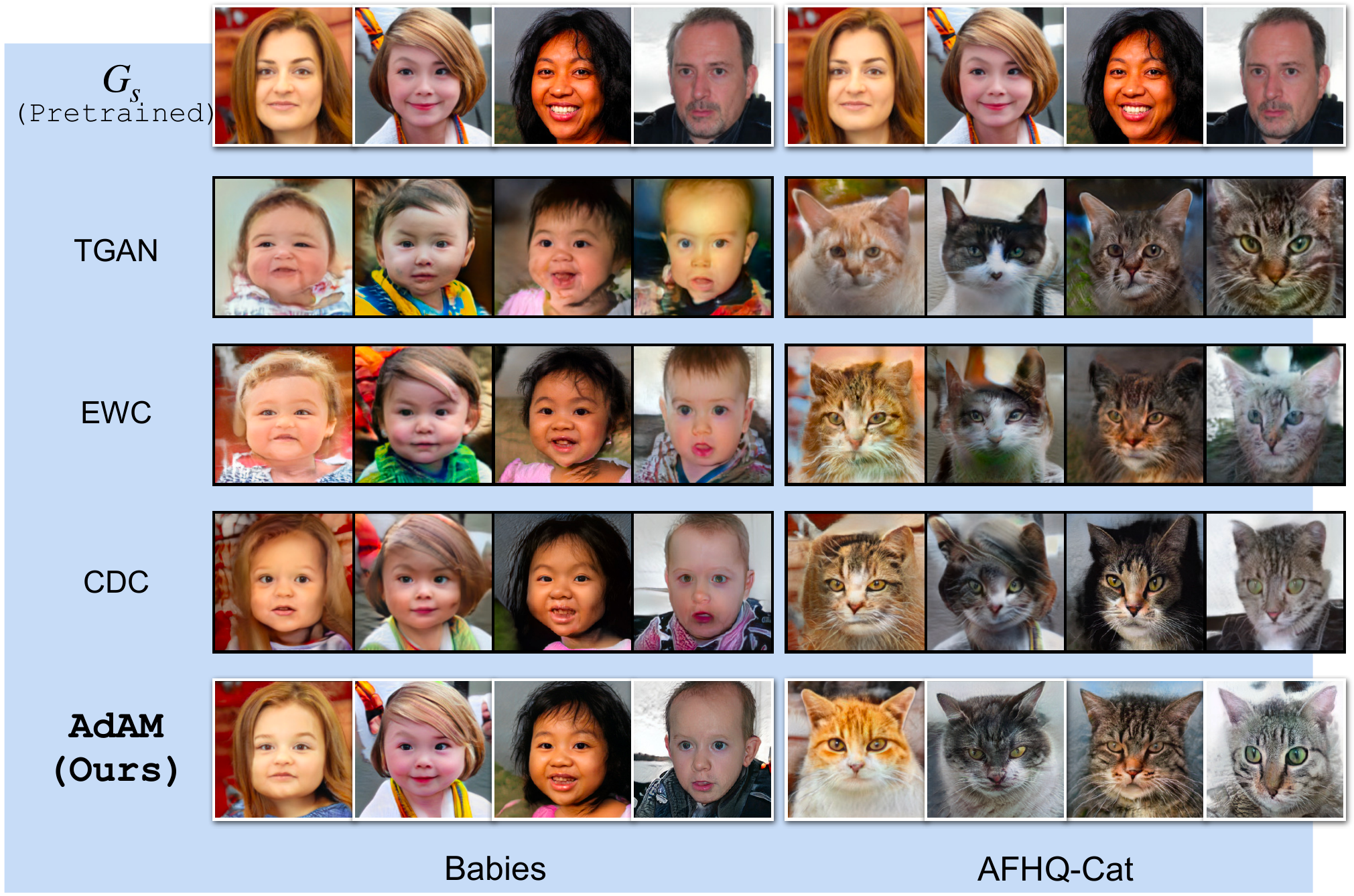}
    \caption{FFHQ $\rightarrow$ Babies (Left) and AFHQ-Cat (Right) with 100 samples for adaptation.}
    \label{fig:100-shot}
\end{figure}

\subsection{KID / Intra-LPIPS / standard deviation of experiments}
\label{sec-supp:kid_intr_lpips}

\textbf{KID / Intra-LPIPS.} 
In addition to FID scores reported in the main paper, we evaluate KID \cite{li2018mmd-aae} and Intra-LPIPS \cite{zhang2018lpips}.
We remark the KID ($\downarrow$) is another metric in addition to FID ($\downarrow$) to measure the quality of generated samples, and Intra-LPIPS ($\uparrow$) measures the diversity of generated samples. 
In literature, the original LPIPS \cite{zhang2018lpips} evaluates the perceptual distance between images. We follow CDC \cite{ojha2021fig_cdc} and DCL \cite{zhao2022dcl} to measure the Intra-LPIPS, a variant of LPIPs, to evaluate the degree of diversity. Firstly, we generate 5,000 images and assign them to one of 10-shot target samples, based on the closet LPIPS distance. Then, we calculate the LPIPS of 10 clusters and take average.
KID and Intra-LPIPS results are reported in Tables \ref{table-supp:kid} and \ref{table-supp:intra_lpips} respectively.
As one can observe, our proposed adaptation-aware FSIG method outperforms SOTA FSIG methods \cite{li2020fig_EWC, ojha2021fig_cdc, zhao2022dcl} and produces high quality images with good diversity.

\begin{table}[!h]
    \centering
    \caption{
   KID ($\downarrow$) score of different methods with the same checkpoint of Table {\color{red}2}
    in the main paper. The values are in $10^3$ units, following \cite{karras2020ADA, chai2021ensembling}.
    \label{tab:supp_kid}
    }
    \begin{adjustbox}{width=0.8\textwidth}
    \begin{tabular}{l|cccccccccc}
    \toprule
         Method & TGAN & FreezeD & EWC & CDC & DCL & RSSA & LLN & AdAM \\ \hline
         Babies & 81.92 & 65.14 & 51.81 & 51.74 & 43.46 & 73.99 & 39.42 & \textbf{28.38}\\
         AFHQ-Cat & 41.912 & 38.834 & 58.65 & 196.60 & 117.82 & 247.6 & 267.62 & \textbf{32.78} \\
    \bottomrule
    \end{tabular}
    \end{adjustbox}
    \label{table-supp:kid}
\end{table}

\begin{table}[!h]
    \centering
    \caption{
    Intra-LPIPS ($\uparrow$) of different methods, the standard deviation is calculated over 10 clusters. Compared to the baseline models (TGAN/FreezeD) or state-of-the-art FSIG methods (EWC/CDC/DCL/RSSA/LLN), our proposed method can achieve a good trade-off between diversity and quality of the generated images, see Table {\color{red} 2} in main paper for FID score.
    }
    \begin{adjustbox}{width=\textwidth}
    \begin{tabular}{l|cccccccccc}
    \toprule
         Method & TGAN & FreezeD & EWC & CDC & DCL & RSSA & LLN & AdAM \\ \hline
         Babies  & 0.517 $\pm$ 0.03 & 0.518 $\pm$ 0.06 & 0.523 $\pm$ 0.03 & 0.573 $\pm$ 0.04 & 0.582 $\pm$ 0.03 & 0.577 $\pm$ 0.03 & 0.585 $\pm$ 0.02 & 0.590 $\pm$ 0.03 \\
         AFHQ-Cat & 0.490 $\pm$ 0.02 & 0.492 $\pm$ 0.04 & 0.587 $\pm$ 0.04 & 0.629 $\pm$ 0.03 & 0.616 $\pm$ 0.05 & 0.618 $\pm$ 0.03 & 0.630 $\pm$ 0.04 & 0.557 $\pm$ 0.02 \\
    \bottomrule
    \end{tabular}
    \end{adjustbox}
    \label{table-supp:intra_lpips}
\end{table}

\begin{figure}[h]
    \centering
    \includegraphics[width=\textwidth]{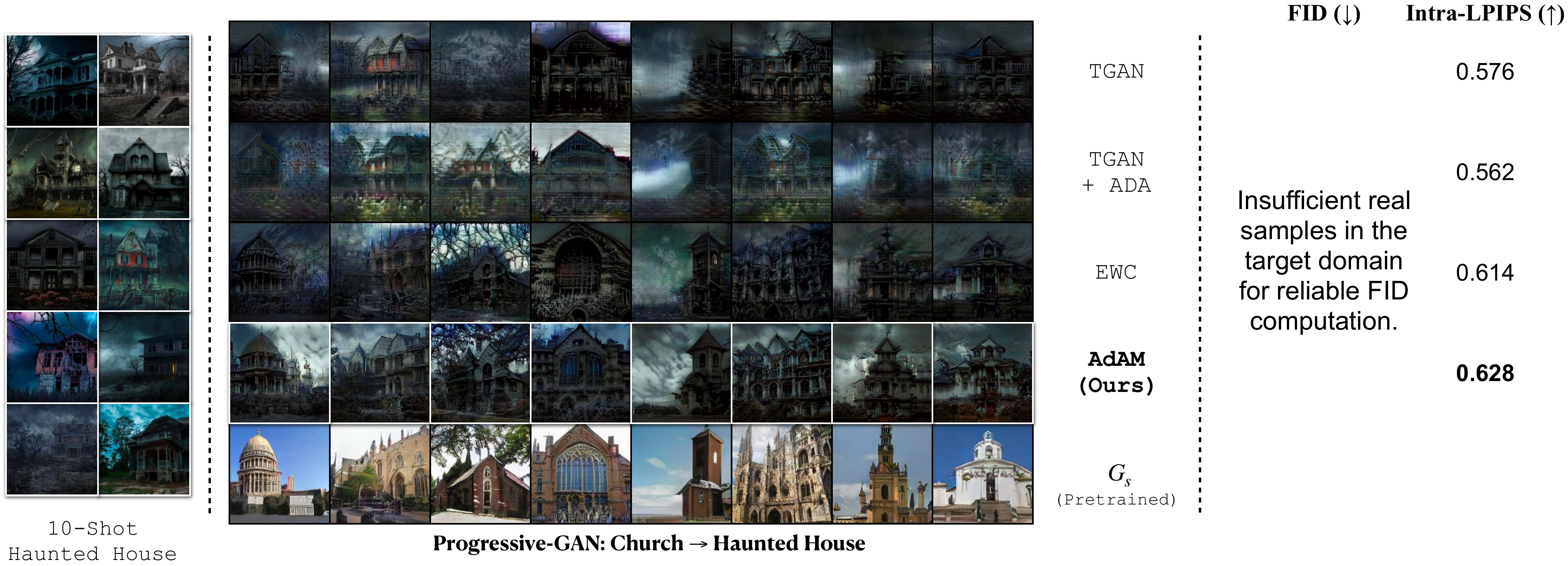}
    \caption{
    Church $\rightarrow$ Haunted Houses adaptation results using pre-trained ProGAN \cite{karras2017progan} generator.}
    \label{fig:rebuttal_progan_haunted_houses}
\end{figure}

\begin{figure}[h]
    \centering
    \includegraphics[width=\textwidth]{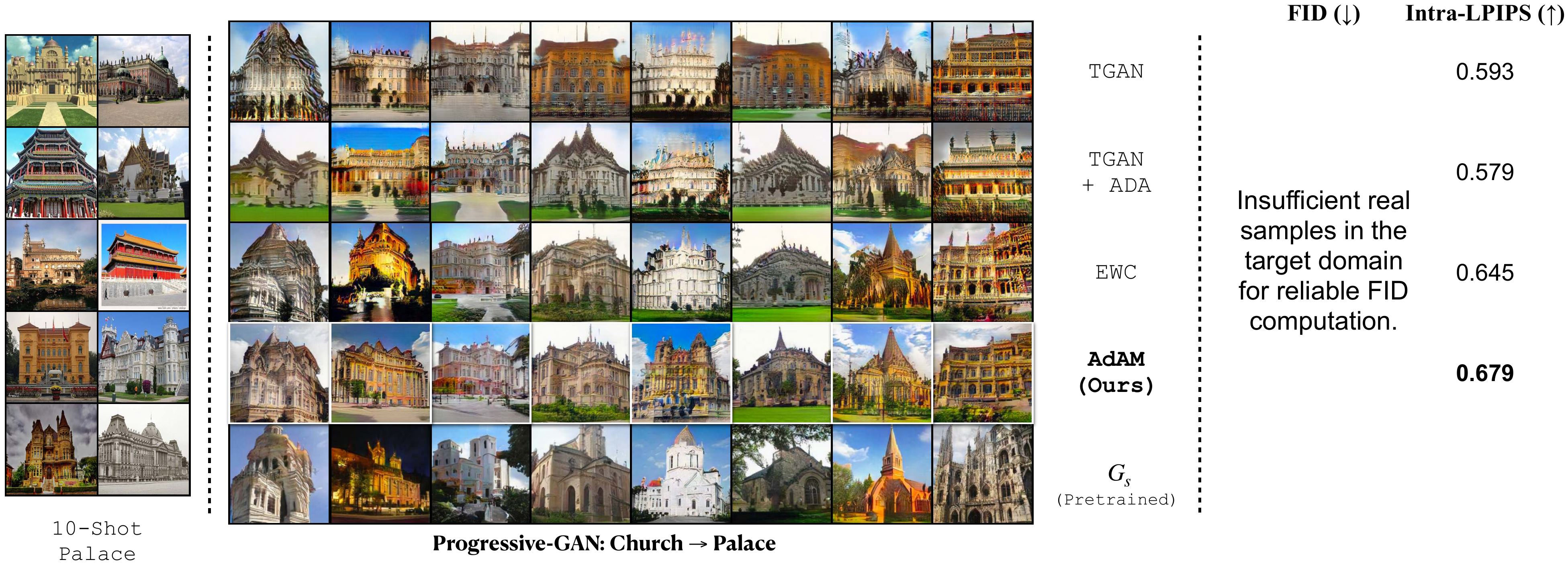}
    \caption{
    Church $\rightarrow$ Palace adaptation results using pre-trained ProGAN \cite{karras2017progan} generator.}
    \label{fig:rebuttal_progan_palace}
\end{figure}

\subsection{FID measurements with limited target domain samples}
\label{supp-sec:fid-analysis}
To characterize source $\rightarrow$ target domain proximity, we used FID and LPIPS measurements in Sec. {\color{red} 3} in the main paper. 
FID involves distribution estimation using first-order (mean) and second-order (trace) moments, i.e.: $\text{FID} = \text{mean}_{component}+ \text{trace}_{component}$ \cite{heusel2017FID}.
Generally, 50K real and generated samples are used for FID calculation \cite{chong2020effectively}.
Given that some of our target domain datasets contain limited samples, e.g.:AFHQ Cat \cite{choi2020starganv2}, Dog \cite{choi2020starganv2}, Wild \cite{choi2020starganv2} datasets contain $\approx$ 5K samples, we conduct extensive experiments to show that FID measurements with limited samples give reliable estimates, thereby reliably characterizing source $\rightarrow$ target domain proximity.
Specifically, we decompose FID into mean and trace components and study the effect of target domain sample size to show that our proximity measurements using FID are reliable.

\textbf{Experiment setup.}
We use 3 large datasets namely FFHQ \cite{karras2020styleganv2} (70K samples), LSUN-Bedroom \cite{yu15lsun} (70K samples) and LSUN-Cat \cite{yu15lsun} (70K samples). 
We use FFHQ (70K samples) as the source domain and study the effect of sample size on FID measure.
Specifically, we decompose FID into mean and trace components in this study.
We consider FFHQ (self-measurement), LSUN-Bedroom and LSUN-cat as target domains. 
We sample 13, 130, 1300, 2600, 5200, 13000, 52000 samples from the target domain and measure the FID with FFHQ (70K samples), and compare it against the FID obtained by using the entire 70K samples from the target domain.

\textbf{Results / Analysis.}
The results are shown in Table \ref{table-supp:fid-analysis}.
As one can observe, with $\approx$ 2600 samples, we can reliably estimate FID as it becomes closer to the FID measured using the entire 70K target domain samples.
Hence, we show that our source $\rightarrow$ target proximity measurements using FID are reliable.

\begin{table}[!h]
\caption{
\textit{FID measurements with limited target domain samples give reliable estimates to characterize source $\rightarrow$ target domain proximity:}
FFHQ (70K) is the source domain.
We use FFHQ (self-measurement), LSUN-Bedroom and LSUN-Cat as target domains.
We use different number of samples from target domain to measure FID. We also decompose FID into mean and trace components in this study.
We sample 13, 130, 1300, 2600, 5200, 13000, 52000 images from the target domain and measure the FID with source domain (FFHQ / 70K), and compare it against the FID obtained by using the entire 70K samples from the target domain.
Each experiment is repeated \textit{100 times} and we report the results with standard deviation.
We also report mean and trace components separately.
As one can observe, with $\approx$ 2600 samples, we can reliably estimate FID as it becomes closer to the FID measured using the entire 70K target domain samples.
Therefore, this study shows that our source $\rightarrow$ target proximity measurements using FID are reliable.
}
  \begin{adjustbox}{width=\textwidth}
  \begin{tabular}{l|c|cccccccc}\toprule
\textbf{FID} &\textbf{} &\textbf{13} &\textbf{130} &\textbf{1300} &\textbf{2600} &\textbf{5200} &\textbf{13, 000} &\textbf{52, 000} &\textbf{70, 000} \\ \toprule
\multirow{3}{*}{FFHQ}  &FID &196.3 $\pm$ 11.8 &83.4 $\pm$ 2.2 &15.3 $\pm$ 0.2 &\textbf{7 $\pm$ 0.1} &3.3 $\pm$ 0 &1.2 $\pm$ 0 &0.1 $\pm$ 0 &0 $\pm$ 0 \\
&mean &12 $\pm$ 2.7 &1.3 $\pm$ 0.3 &0.1 $\pm$ 0 &\textbf{0.1 $\pm$ 0} &0 $\pm$ 0 &0 $\pm$ 0 &0 $\pm$ 0 &0 $\pm$ 0 \\
&trace &184.3 $\pm$ 10.3 &82.2 $\pm$ 2 &15.2 $\pm$ 0.2 &\textbf{6.9 $\pm$ 0.1} &3.3 $\pm$ 0 &1.1 $\pm$ 0 &0.1 $\pm$ 0 &0 $\pm$ 0 \\ \midrule
\multirow{3}{*}{Bedroom} &FID &358.5 $\pm$ 9.3 &301.9 $\pm$ 2.4 &251 $\pm$ 1.2 &\textbf{243.6 $\pm$ 0.8} &240.1 $\pm$ 0.5 &238.2 $\pm$ 0.4 &237.2 $\pm$ 0.2 &237.2 $\pm$ 0.1 \\
&mean &139.3 $\pm$ 8 &131.8 $\pm$ 2.5 &131.4 $\pm$ 0.9 &\textbf{131.1 $\pm$ 0.6} &131.1 $\pm$ 0.4 &131.1 $\pm$ 0.3 &131.1 $\pm$ 0.1 &131.1 $\pm$ 0.1 \\
&trace &219.1 $\pm$ 9.9 &170.1 $\pm$ 1.9 &119.6 $\pm$ 0.6 &\textbf{112.5 $\pm$ 0.4} &109.1 $\pm$ 0.3 &107.1 $\pm$ 0.2 &106.1 $\pm$ 0.1 &106 $\pm$ 0.1 \\ \midrule
\multirow{3}{*}{Cat} &FID &370.2 $\pm$ 18.7 &283.7 $\pm$ 4.4 &209.7 $\pm$ 1.2 &\textbf{199.9 $\pm$ 0.8} &195.3 $\pm$ 0.6 &192.8 $\pm$ 0.4 &191.4 $\pm$ 0.2 &191.3 $\pm$ 0.1 \\
&mean &105.7 $\pm$ 8.4 &93 $\pm$ 2.2 &91.7 $\pm$ 0.8 &\textbf{91.7 $\pm$ 0.5} &91.6 $\pm$ 0.4 &91.6 $\pm$ 0.2 &91.6 $\pm$ 0.1 &91.6 $\pm$ 0.1 \\
&trace &264.5 $\pm$ 15.7 &190.7 $\pm$ 3.6 &118 $\pm$ 0.9 &\textbf{108.2 $\pm$ 0.6} &103.7 $\pm$ 0.4 &101.2 $\pm$ 0.3 &99.9 $\pm$ 0.1 &99.7 $\pm$ 0.1 \\
\bottomrule
\end{tabular}
\end{adjustbox}
\label{table-supp:fid-analysis}
\end{table}

\subsection{KML induces restrained update of the kernels}
\label{restrain-kml}
Recall that for a convolutional kernel that we applied KML, we use Eqn~{\color{red} 1} in the main paper to modulate its parameters for the knowledge preservation, where we only update the modulation parameters and keep the pretrained weights fixed.
In our experiments, we have demonstrated that KML can preserve source knowledge that is useful for target domain adaptation. 
In this section, we demonstrate that, our proposed KML in AdAM indeed achieves restrained update of the important kernels, therefore it leads to the effective knowledge preservation (more implementation details of KML are in Sec. {\color{red} 4} of the main paper).

\textbf{Experiment setups.}
We compute the percentage of weights $\mathbf{W}$ update (normalized), take average over all kernels and layers, and compare with three types of training scheme: simple fine-tuning, KML to all kernels in the GAN, and our proposed AdAM with a threshold to determine if a kernel should be modulated. The output is a single scalar $\boldsymbol{q}$ (i.e., percentage) as below:
\begin{equation}
    \boldsymbol{q}\% = \Delta \mathbf{W} / \|\mathbf{W}\| \times 100\%,
\end{equation}
where $\Delta \mathbf{W} = |\hat{\mathbf{W}} - \mathbf{W}|$ and $\hat{\mathbf{W}}$ is obtained in Eqn~{\color{red} 1} in the main paper (note that in AdAM only part of the kernels are updated).
For the source/target domains, we apply FFHQ $\rightarrow$ Babies and FFHQ $\rightarrow$ AFHQ-Cat, similar to the experiments in Table {\color{red} 2} in the main paper.

\textbf{Results and analysis.} 
The results are shown in Table \ref{tab:parameter_update_percent}. 
We demonstrate that, indeed the proposed KML can achieve restrained update of kernels that is updated, therefore the source knowledge important for target domain adaptation is effectiveness preserved.

\begin{table}[h]
    \centering
    \caption{
    Percentage change of weights before and after the target domain adaptation. It is clearly observed that, our proposed KML indeed achieves restrained updated to individual kernels, and consequently the source knowledge identified important for target domain is effectively preserved. This is also validated through various visualization results in our experiments.
    }
        \begin{tabular}{l|cc}
        \multicolumn{3}{c}{\textbf{FFHQ} $\rightarrow$ \textbf{Babies}}\\
         \toprule
         {Method} & {Generator} & {Discriminator} \\\hline
         TGAN \cite{wang2018transferringGAN} & 35.83\% & 46.76\% \\ 
         AdAM (apply KML to all kenrels) & 2.62 \% & 2.28\% \\
         AdAM (Ours, apply KML to partial kernels) & 19.03\% & 20.32\% \\
         \bottomrule
        \end{tabular}
    
    \vspace{2mm}
        \begin{tabular}{l|cc}
        \multicolumn{3}{c}{\textbf{FFHQ} $\rightarrow$ \textbf{AFHQ-Cat}}\\
         \toprule
         {Method} & {Generator} & {Discriminator} \\\hline
         TGAN \cite{wang2018transferringGAN} & 50.48\% & 73.34\% \\ 
         AdAM (apply KML to all kenrels) & 7.28\% & 4.80\% \\
         AdAM (Ours, apply KML to partial kernels) & 35.64 \% & 32.99\% \\
         \bottomrule
        \end{tabular}
    \label{tab:parameter_update_percent}
\end{table}

\subsection{Additional Results/Analysis for GAN Dissection}
In this section, we provide additional GAN dissection results as a way to interpret and 
visualize the knowledge / information encoded by high FI kernels (identified by our proposed AdAM for FSIG) that are important for target domain adaptation: 

\begin{itemize}
    \item Visualizing high FI kernels for Church $\rightarrow$ Haunted Houses adaptation : The results for FI estimation for kernels and several distinct semantic concepts learnt by high FI kernels are shown in Figure \ref{fig:rebuttal_gan_dissection_haunted_houses}.
    In Figure \ref{fig:rebuttal_gan_dissection_haunted_houses}, we visualize four examples of high FI kernels: (a), (b), (c), (d) corresponding to concepts building, building, tree and wood respectively. Using GAN Dissection, we observe that a notable amount of high FI kernels correspond to useful source domain concepts including building, tree and wood (texture) which are preserved when adapting to Haunted Houses target domain. We remark that these preserved concepts are useful to the target domain for adaptation. 
    
    \item Visualizing high FI kernels for Church $\rightarrow$ Palace adaptation : The results for FI estimation for kernels and several distinct semantic concepts learnt by high FI kernels are shown in Figure \ref{fig:rebuttal_gan_dissection_palace}.
    In Figure \ref{fig:rebuttal_gan_dissection_palace}, we visualize four examples of high FI kernels: (a), (b), (c), (d) corresponding to concepts grass, grass, building and building respectively. 
    Using GAN Dissection, we observe that a notable amount of high FI kernels correspond to useful source domain concepts including grass and building which are preserved when adapting to Palace target domain. We remark that these preserved concepts are useful to the target domain (Palace) for adaptation. 
\end{itemize}

\begin{figure}[h]
    \centering
    \includegraphics[width=\textwidth]{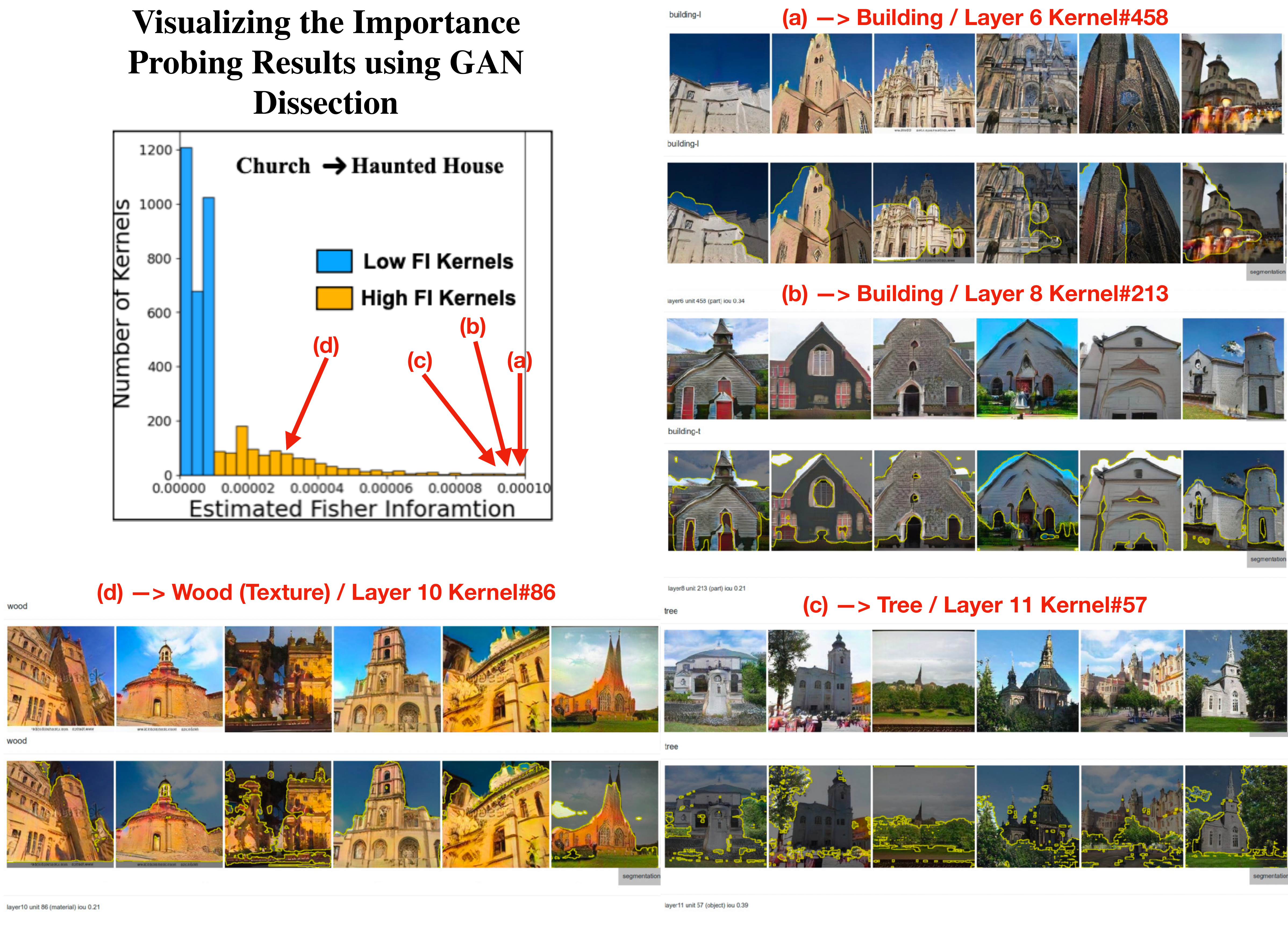}
    \caption{
    Visualizing high FI kernels using GAN Dissection \cite{bau2019gan} for Church $\rightarrow$ Haunted Houses 10-shot adaptation.
    In visualization of each high FI kernel, the first row shows different images generated by the source generator, and the second row highlights the concept encoded by the corresponding high FI kernel as determined by GAN Dissection.
    We observe that a notable amount of high FI kernels correspond to useful source domain concepts including building (a, b), tree (c) and wood (d) which are preserved when adapting to Haunted Houses target domain. We remark that these preserved concepts are useful to the target domain (Haunted House) for adaptation.}
    \label{fig:rebuttal_gan_dissection_haunted_houses}
\end{figure}

\begin{figure}[h]
    \centering
    \includegraphics[width=\textwidth]{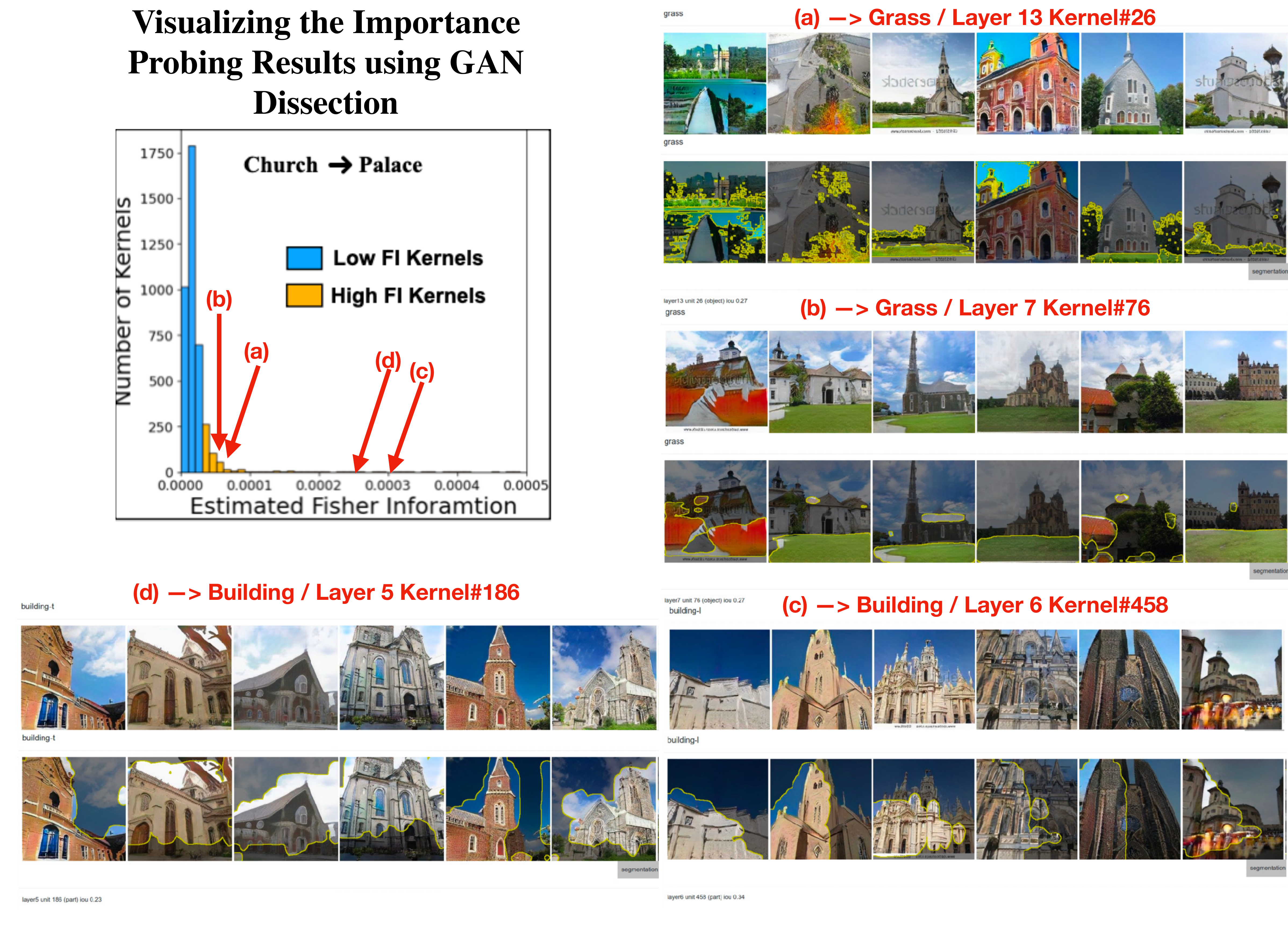}
    \caption{
    Visualizing high FI kernels using GAN Dissection \cite{bau2019gan} for Church $\rightarrow$ Palace 10-shot adaptation.
    In visualization of each high FI kernel, the first row shows different images generated by the source generator, and the second row highlights the concept encoded by the corresponding high FI kernel as determined by GAN Dissection.
    We observe that a notable amount of high FI kernels correspond to useful source domain concepts including grass (a, b) and building (c, d) which are preserved when adapting to Palace target domain. We remark that these preserved concepts are useful to the target domain (palace) for adaptation.}
    \label{fig:rebuttal_gan_dissection_palace}
\end{figure}


\section{Additional Discussion}
\label{sec-supp:checklist}

\subsection{Potential societal impact}
\label{sec-supp:societal_impact_supp}
Given very limited target domain samples (i.e.: 10-shot or less), our proposed method achieves SOTA results in FSIG with different source / target domain proximity.
In the main paper, we discussed the ethical concern  by adapting our FSIG method to a particular person (we used Obama Dataset). Here, we show additional adaptation results (1-shot, 5-shot and 10-shot), as shown in Figure \ref{fig:supp_full_obama}.
Though our work shows exciting results by pushing the limits of FSIG, we urge researchers, practitioners and developers to use our work with privacy, ethical and moral concerns. 
\begin{figure}[!t]
    \centering
    \includegraphics[width=\textwidth]{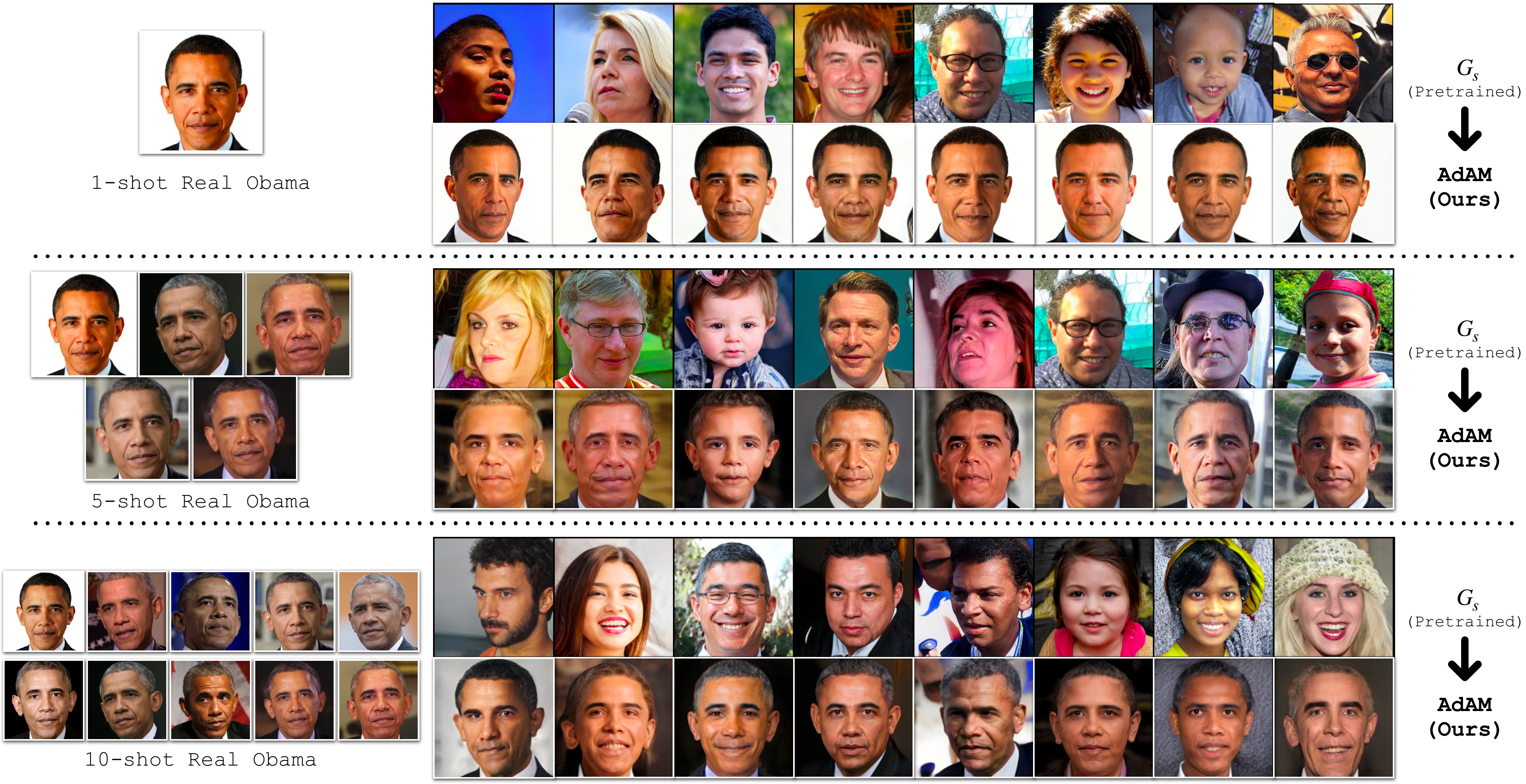}
    \caption{Complete results for FFHQ $\rightarrow$ Obama Dataset adaptation.}
    \label{fig:supp_full_obama}
\end{figure}

\subsection{Amount of compute}
\label{sec-supp:amount_compute}
The amount of computation consumed in this work is reported in Table~\ref{table-supp:compute}. We include the compute amount for each experiment as well as the CO$_2$ emission (in kg). In practice, our experiments can be run on a single GPU and each experiment can be finished in 2 hours, therefore, the computational demand of our work is not high. 

\setlength{\tabcolsep}{1.5 mm}
\renewcommand{\arraystretch}{1.25}
\begin{table}[!ht]
    \centering
    \caption{
The GPU hours consumed for the experiments conducted to obtain the reported values. CO$_2$ emission values are computed using \url{https://mlco2.github.io/impact}~\cite{lacoste2019quantifying_co2}. 
We remark since we mostly use the pretrained models in our experiments, therefore, our algorithm is friendly to individual practitioners.
}
  \begin{adjustbox}{width=0.8\textwidth}
  \begin{tabular}{l|c|c|c}\toprule[1.5pt]
\textbf{Experiment} &\textbf{Hardware Platform} &\textbf{GPU hours} &\textbf{Carbon emitted in kg} \\
\hline
Main paper : Table 2 & \multirow{15}{*}{Tesla V100-SXM2 (32 GB)} & 306 & 52.33 \\ 
Main paper : Table 3 &  & 48 & 8.21    \\ 
Main paper : Table 4 &  & 15 & 2.56    \\ 
Main paper : Table 5 &  & 183 & 31.29   \\ 
Main paper : Table 6 &  & 48 &  8.21   \\ 
Main paper : Table 7 &  & 31 &  5.30   \\ 
Main paper : Figure 2 & & 6 & 1.03  \\ 
Main paper : Figure 3 / Figure 5 & & 120 & 20.52  \\ 
Main paper : Figure 6 & & 40 & 6.84 \\ 
Main paper : Figure 7 & & 22 & 3.76 \\
Main paper : Figure 8 &  & 210 & 35.91 \\ 
Main paper : Figure 9 &  & 120 & 20.52 \\
Main paper : Figure 10 & & 12 & 2.05  \\ 
Main paper : Figure 11 & & 34 & 5.81 \\ 
Main paper : Figure 12 &  & 2 & 0.34 \\ 
\hline
Supplement : Extended Experiments & \multirow{2}{*}{Tesla V100-SXM2 (32 GB)} & 86 & 14.71 \\ 
Supplement : Ablation Study & & 32 & 5.47 \\ 
\hline
Hyper-parameter tuning & Tesla V100-SXM2 (32 GB) & 30 & 5.13 \\ 
\hline
\textbf{Total} & - & \textbf{1345} & \textbf{229.99} \\
\bottomrule[1.5pt]
\end{tabular}
\end{adjustbox}
\label{table-supp:compute}
\end{table}

\end{document}